\documentclass[twoside,11pt]{article}

\usepackage[preprint]{jmlr2e}

\usepackage{graphicx}
\usepackage[shortlabels]{enumitem}
\usepackage{amsmath}
\usepackage{caption}
\usepackage{subcaption}
\usepackage{wrapfig}
\usepackage{array}
\usepackage{multirow}
\usepackage{adjustbox}
\usepackage{comment}
\usepackage[leftcaption]{sidecap}
\usepackage{booktabs}      % professional-quality tables
\usepackage{wrapfig}       % for wrapping text around figure/table
\newcommand{\lrschedule}{\textit{Knee schedule}}
\newcommand{\lrscheduleshort}{\textit{Knee}}

\jmlrheading{22}{2021}{1-35}{}{}{}{Nikhil Iyer, V.Thejas, Nipun Kwatra, Ramachandran Ramjee and Muthian Sivathanu}

\ShortHeadings{Wide-minima Density Hypothesis and the Explore-Exploit Learning Rate Schedule}{Iyer, Venkatesh, Kwatra, Ramjee and Sivathanu}

\firstpageno{1}

\begin{document}

\title{Wide-minima Density Hypothesis and the \\Explore-Exploit Learning Rate Schedule}

\author{\name Nikhil Iyer  
     \thanks{Work done during an internship at Microsoft Research India} \\
       \addr Microsoft Research India
       \email iyernikhil007@gmail.com
       \AND
       \name V Thejas \footnotemark[1] \\
       \addr Atlassian India
       \email thejasvenkatesh97@gmail.com 
       \AND
       \name Nipun Kwatra \\ 
       \addr Microsoft Research India
       \email Nipun.Kwatra@microsoft.com
       \AND 
       \name Ramachandran Ramjee \\ 
       \addr Microsoft Research India
       \email ramjee@microsoft.com
       \AND 
       \name Muthian Sivathanu \\
       \addr Microsoft Research India
       \email muthian@microsoft.com}

\editor{}

\maketitle

\begin{abstract}

Several papers argue that wide minima generalize better than narrow minima. In this paper, through detailed experiments that not only corroborate the generalization properties of wide minima, we also
provide empirical evidence for a new hypothesis that the density of wide minima is likely lower than the density of narrow minima.
Further, motivated by this hypothesis, we design a novel explore-exploit learning rate schedule. On a variety of image and natural language datasets, compared to their original hand-tuned learning rate baselines, we show that our explore-exploit schedule can result in either up to 0.84\% higher absolute accuracy using the original training budget or up to 57\% reduced training time while achieving the original reported accuracy. For example, we achieve state-of-the-art (SOTA) accuracy for IWSLT'14 (DE-EN) dataset by just modifying the learning rate schedule of a high performing model.

%While the generalization properties of neural networks are not yet well understood, several papers argue that wide minima generalize better than narrow minima. In this paper, through detailed experiments that not only corroborate the generalization properties of wide minima, we also provide empirical evidence for a new hypothesis that the density of wide minima is likely lower than the density of narrow minima.

%Further, motivated by this hypothesis, we design a novel explore-exploit learning rate schedule. On a variety of image and natural language datasets, compared to their original hand-tuned learning rate baselines, we show that our explore-exploit schedule can result in either up to 0.84\% higher absolute accuracy using the original training budget or up to 57\% reduced training time while achieving the original reported accuracy. For example, we achieve SOTA accuracy for IWSLT'14 (DE-EN) and WMT'14 (DE-EN) datasets by just modifying the learning rate schedule of a high performing model.
    
\end{abstract}
\section{Introduction}
\label{sec:introduction}

One of the fascinating properties of deep neural networks (DNNs) is their ability to generalize well, i.e., deliver high accuracy on the unseen test dataset. It is well-known that the learning rate (learning rate) schedules  play an important role in the generalization performance~\citep{keskar2016large,wu2018sgd,goyal-imagenet-in-an-hour-2017}. In this paper, we study the question, {\it what are the key properties of a learning rate schedule that help DNNs generalize well during training?}

We start with a series of experiments training Resnet18 on Cifar-10 over 200 epochs. We vary the number of epochs trained at a high learning rate of $0.1$, called the {\it explore} epochs, from 0 to 100  and divide up the remaining epochs equally for training with learning rates of $0.01$ and $0.001$. Note that the training loss typically stagnates around 50 epochs with $0.1$ learning rate. Despite that, we find that as the number of explore epochs increase to 100, the average test accuracy also increases. We also find that the minima found in higher test accuracy runs are wider than the minima from lower test accuracy runs, corroborating past work on wide-minima and generalization~\citep{keskar2016large,hochreiter1997flat,jastrzkebski2017three,wang2018identifying}. Moreover, what was particularly surprising was that, even when using fewer explore epochs, a few runs out of many trials still resulted in high test accuracies!

Thus, we not only find that an initial exploration phase with a high learning rate is essential to the good generalization of DNNs, but that {\it this exploration phase needs to be run for sufficient time, even if the training loss stagnates much earlier. Further, we find that, even when the exploration phase is not given sufficient time, a few runs still see high test accuracy values.}

%Consider the following series of experiments of Cifar-10 training over 200 epochs using a standard step schedule with learning rates of $0.1, 0.01$ and $0.001$. We vary the number of epochs trained using $0.1$ learning rate, called the explore epochs, from 40 to 100 epochs, and divide up the rest of the training equally between $0.01$ and $0.001$ learning rates. For each experimental setting, we conduct 50 random trials and plot the distributions of final test accuracy and the largest Eigenvalue of the Hessian of the loss at the final minima (a measure of the minima sharpness~\cite{keskar2016large,wu2018sgd}) in Figures~\ref{fig:accuracy_hist_explores} and~\ref{fig:eigenvalues_hist_explores}, respectively. It is clear from the figures that, as the number of explore epochs increase, the distribution gets skewed towards higher test accuracy and wider minima.

%To explain this observation, we look into how the choice of learning rate can impact the chances of the optimization landing in a wide vs narrow minima. To generalize well, we want the optimizer to land in wide minima. 

To explain these observations, we hypothesize that, {\it in the DNN loss landscape, the density of narrow minima is significantly higher than that of wide minima.} Intuitively, a large learning rate can escape narrow minima easily (as the optimizer can jump out of them with large steps). However, once it reaches a wide minima, it is likely to get stuck in it (if the "width" of the wide minima is large compared to the step size). With fewer explore epochs, a large learning rate might still get lucky occasionally in finding a wide minima but invariably finds only a narrower minima due to their higher density. As the explore duration increases, the probability of eventually landing in a wide minima also increases. Thus, {\it a minimum duration of explore} is necessary to land in a wide minimum with high probability. 

An observation on the rarity of wide minima has been hinted at by prior work~\citep{wu2018sgd,baldassi2020shaping} based on theoretical analysis of simple neural networks (see Section~\ref{sec:related_work}). In this paper, we add significant empirical evidence to these theoretical observations. We believe that all these results together constitute sufficient evidence for this observation to now be classified as a hypothesis, that we term the wide-minima density hypothesis.

The hypothesis helps {\it explain} not only our experiments but also the generalization out-performance of prior heuristic-based learning rate decay schemes such as cosine decay~\citep{loshchilov2016sgdr}. Cosine decay {\it implicitly} maintains a higher learning rate during the first half of training compared to schemes like linear decay. Based on the hypothesis, the higher learning rate allows cosine decay to find wider minima with higher probability, resulting in cosine decay's better generalization compared to linear decay.

Apart from helping explain empirical observations, the hypothesis also enables a principled learning rate schedule design that {\it explicitly} accounts for the requisite explore duration. Motivated by the hypothesis, we design a novel {\it Explore-Exploit} learning rate schedule, where the initial \textit{explore} phase optimizes at a high learning rate in order to arrive in the vicinity of a wide minimum. This is followed by an \textit{exploit} phase which descends to the bottom of this wide minimum. We give {\it explore} phase enough time so that the probability of landing in a wide minima is high. For the \textit{exploit} phase, we experimented with multiple schemes, and found a simple, parameter-less, linear decay to zero to be effective. \textit{Thus, our proposed learning rate schedule optimizes at a constant high learning rate for a given duration, followed by a linear decay to zero. We call this learning rate schedule the \lrschedule{}.}

%For example, on BERT\textsubscript{BASE} fine-training~\cite{devlin2018bert}, \lrschedule{} is able achieve an EM score of 81.38, compared to 80.9. 

We extensively evaluate the \lrschedule{} across a wide range of models and datasets, ranging from NLP (BERT pre-training, Transformer on WMT'14(EN-DE) and IWSLT'14 (DE-EN)) to CNNs (ImageNet on ResNet-50, Cifar-10 on ResNet18), and spanning multiple optimizers: SGD Momentum, Adam, RAdam,  and LAMB.  In all cases, \lrschedule{} improves the test accuracy of state-of-the-art hand-tuned learning rate schedules, when trained using the original training budget. The explore duration is a hyper-parameter in \lrschedule{} but even if we set the explore duration to a fixed 50\% fraction of total training budget, we find that it still outperforms prior schemes.

We also experimented with reducing the training budget, and found that \lrschedule{} can achieve the same accuracy as the baseline under significantly reduced training budgets. For the BERT\textsubscript{LARGE} pretraining, WMT'14(EN-DE) and ImageNet experiments, we are able to train in 33\%, 57\% and 44\% less training budget, respectively, for the same test accuracy. This corresponds to significant savings in GPU compute, e.g. \textit{savings of over 1000 V100 GPU-hours} for BERT\textsubscript{LARGE} pretraining.

% Savings: 27hrs for imagenet, 20hrs for wmt (4gpus) and 24hrs for bert(64gpus)

The main contributions of our work \footnote{Our work is available at:  https://github.com/nikhil-iyer-97/wide-minima-density-hypothesis} are:
\begin{enumerate}[nosep]
 \item A hypothesis of lower density of wide minima in the DNN loss landscape, backed by extensive experiments, that explains why a high learning rate needs to be {\it maintained for sufficient duration} to achieve good generalization.
  \item The hypothesis {\it explains} the good performance of heuristic-based schemes such as cosine decay,  and promotes a {\it principled design} of learning rate decay schemes.
 \item Motivated by the hypothesis, we design an {\it Explore-Exploit} learning rate schedule called \lrschedule{} that outperforms prior heuristic-based learning rate schedules, including achieving state-of-the-art results on the  IWSLT’14 (DE-EN) dataset.
\end{enumerate}
\vspace{-4pt}
\section{Related Work}
\label{sec:related_work}

\noindent
{\bf Generalization.}
There has been a lot of work on understanding the generalization characteristics of DNNs. \citet{kawaguchi2016deep} found that DNNs have many local minima, but all local minima were also the global minima. It has been observed by several authors that wide minima generalize better than narrow minima~\citep{arora2018stronger,hochreiter1997flat,keskar2016large,jastrzkebski2017three,wang2018identifying} but there have been other works questioning this hypothesis as well~\citep{dinh2017sharp,golatkar2019time,guiroy2019towards,jastrzebski_iclr_2019,yoshida2017spectral}.

\citet{keskar2016large} found that small batch SGD generalizes better and lands in wider minima than large batch SGD. However, recent work has been able to generalize quite well even with very large batch sizes~\citep{goyal-imagenet-in-an-hour-2017,large-batch-training-openai-2018,large-batch-training-google-2018}, by scaling the learning rate linearly as a function of the batch size.~\citet{jastrzebski_iclr_2019}
analyze how batch size and learning rate influence the curvature of not only the SGD endpoint but also the whole trajectory. They found that small batch or large step SGD have similar characteristics, and yield smaller and earlier peak of spectral norm as well as smaller largest eigenvalue.~\citet{chaudhari2019entropy,shapinglandscape2019baldassi} propose methods to drive the optimizer to wide minima. ~\citet{wang2018identifying} analytically show that generalization of a model is related to the Hessian and propose a new metric for the generalization capability of a model that is unaffected by model reparameterization of~\citet{dinh2017sharp}. ~\citet{yoshida2017spectral} argue that regularizing the spectral norm of the weights of the neural network help them generalize better. On the other hand,~\citet{arora2018stronger} derive generalization bounds by showing that networks with low stable rank (high spectral norm) generalize better.~\citet{guiroy2019towards} looks at generalization in gradient-based meta-learning and they show experimentally that generalization and wide minima are not always correlated. 
Finally,~\citet{golatkar2019time} show that regularization results in higher test accuracy specifically when it is applied during initial phase of training, similar to the importance of \lrschedule{}'s explore phase during initial phase of training. In a similar vein,~\citet{li2019towards} explain the regularization benefits of the initial higher learning rate by showing that higher learning rate helps networks learn easier-to-fit general patterns. 

%~\citet{dinh2017sharp} show analytically using model reparameterization that wide minima can be converted to sharp minima without hurting generalization.

\noindent
{\bf Neural network loss landscapes.}
The landscape of loss in neural networks have been extensively studied~\citep{draxler2018essentially, freeman2016topology,garipov2018loss,sagun2017empirical}. These papers point out that the loss landscape contains both wide and narrow minima, and there may even exist a path from one minima to another without barriers. However, there are multiple paths between these minima and some paths indeed face barriers 
(e.g., see Figure 1 in \citet{draxler2018essentially}). Since we don't know which path SGD and other optimizers might follow, even if wide and narrow minima are part of a single basin, SGD and other optimizers might still require higher learning rates to navigate from narrow to wide minima.

\noindent
{\bf Lower density of wide minima.} \citet{wu2018sgd} compares the sharpness of minima obtained by full-batch gradient descent (GD) with different learning rates for small neural networks on FashionMNIST and Cifar10 datasets. They find that GD with a given learning rate finds the theoretically sharpest feasible minima for that learning rate. Thus, in the presence of several flatter minimas, GD with lower learning rates does not find them, leading to the conjecture that density of sharper minima is perhaps larger than density of wider minima. \citet{baldassi2020shaping} show analytically for simple, two-layer non-convex networks that wide minima exists and are rare, compared to narrow minima, local minima and saddle points. In this paper, we add significant evidence to these theoretical observations based on empirical results obtained on large-scale, state-of-the-art neural networks through carefully designed experiments.
\section{Wide-Minima Density Hypothesis}
\label{sec:wide_minima_hypothesis}

Many popular learning rate schedules, such as the step decay schedules for image datasets, start the training with high learning rate, and then reduce the learning rate periodically. For example, consider the case of Cifar-10 on Resnet-18, trained using a typical step learning rate schedule of $0.1, 0.01,$ and  $0.001$ for 100, 50, 50 epochs each. In many such schedules, even though training loss stagnates after several epochs of high learning rate, one still needs to continue training at high learning rate in order to get good generalization. 

\begin{figure*}[ht]
\begin{minipage}{0.45\textwidth}
  \centering
  %\vspace{12pt}
  \includegraphics[width=1\columnwidth]{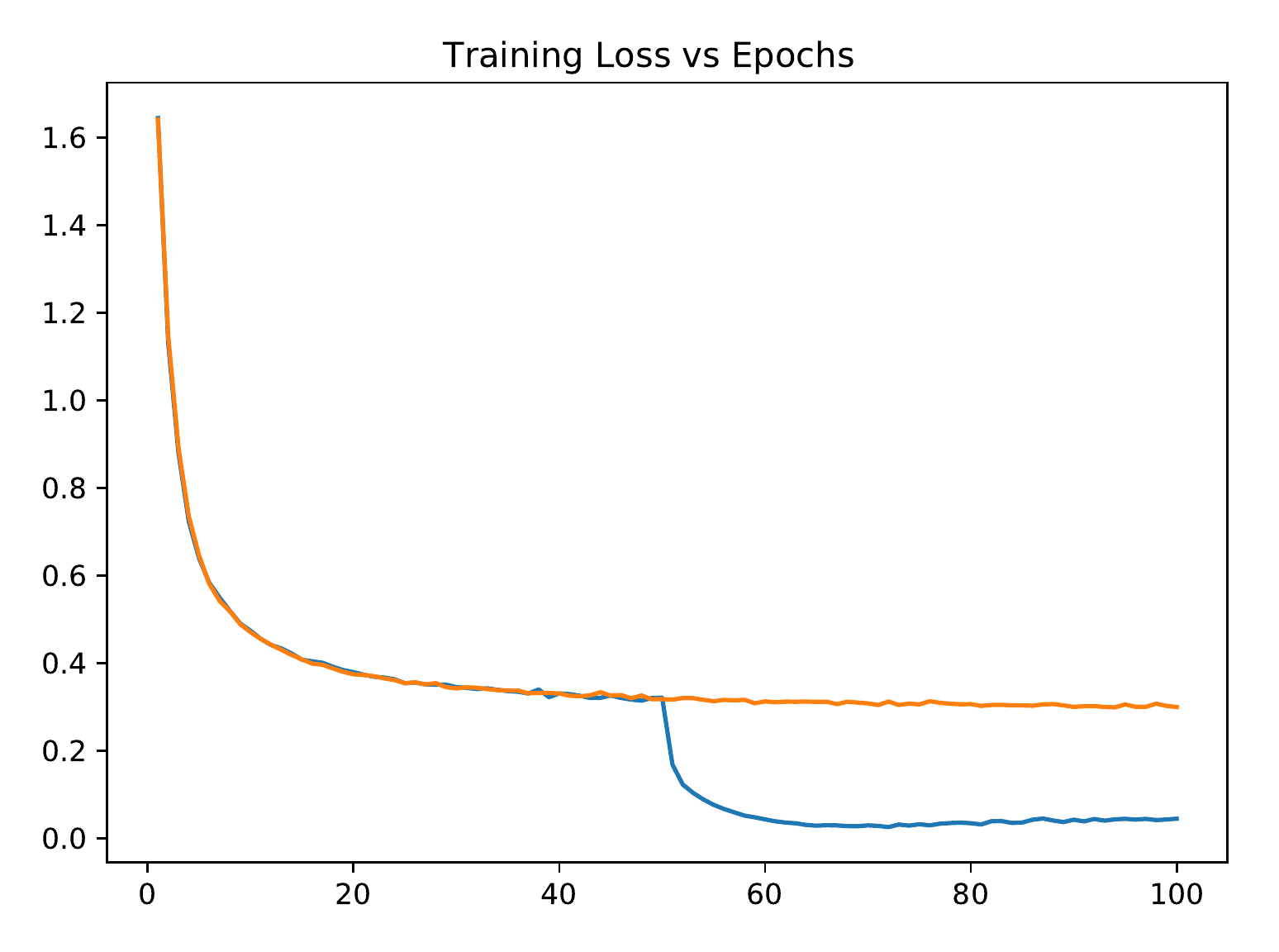}
  \caption{Training loss for Cifar-10 on Resnet-18. Orange plot uses a fixed learning rate of 0.1, while in blue plot, the learning rate is reduced from 0.1 to 0.01 at epoch 50.
  %Locally, it makes sense to reduce the learning rate at epoch 50, as it significantly increases the rate of loss descent.
  }
  \label{fig:cifar-trloss-warmup50vs100}
\end{minipage}
%\end{figure}
\hfil
\begin{minipage}{0.5\textwidth}
  \centering
  \small
  \captionsetup{type=table} %% tell latex to change to table
  %\vspace{12pt}
  \caption{Cifar-10 on Resnet-18 trained for 200 epochs with Momentum. A learning rate of 0.1 is used for the explore epochs. Half the remaining epochs are trained at 0.01 and the other half at 0.001. Reported results are average over 4 runs.}
\begin{tabular}{ccc}
\toprule
Epochs at & Test Accuracy & Train Loss \\ 
0.1 LR &  Avg. (Std. Dev) &  Avg. (Std. Dev.)\\
\midrule
    0 & 94.34 (0.13) & 0.0017 (8e-5) \\
    30 & 94.81 (0.15) & 0.0017 (8e-5) \\
    40 & 94.91 (0.14) & 0.0018 (9e-5) \\
    60 & 95.01 (0.14) & 0.0018 (1e-4) \\ 
    80 & 95.05 (0.15)  & 0.0019 (1e-4) \\
    100 & 95.10 (0.14)  & 0.0021 (1e-4) \\
\bottomrule
\end{tabular}
  \label{tab:warmup_accuracy_baseline}
\end{minipage}
\end{figure*}

For example, Figure~\ref{fig:cifar-trloss-warmup50vs100} shows the training loss for Cifar-10 on Resnet-18, trained with a fixed learning rate of 0.1 (orange curve), compared to a model trained via a step schedule with learning rate reduced at epoch 50 (blue curve). As can be seen from the figure, the training loss stagnates after $\approx$ 50 epochs for the orange curve, and locally it makes sense to reduce the learning rate to decrease the loss. However, as shown in Table~\ref{tab:warmup_accuracy_baseline}, generalization is directly correlated with duration of training at high learning rate, with the highest test accuracy achieved when the high learning rate is used for 100 epochs, well past the point where training loss stagnates. Note that the final training loss remains similar for all runs.

To understand the above phenomena, we perform another experiment. We train Cifar-10 on Resnet-18 for 200 epochs, using a high learning rate of $0.1$ for only 30 epochs and then use learning rate of $0.01$ and $0.001$ for 85 epochs each. We repeat this training 50 times with different random weight initializations. On an average, as expected, this training yields a low test accuracy of $94.81$. However, {\it in 1 of the 50 runs, we find that the test accuracy reaches $95.24$, even higher than the average accuracy of $95.1$ obtained while training at high learning rate for 100 epochs!}

%{\bf Minima Width definition.}
%We would now like to understand the correlation between these test accuracy values and the shape of the minima. In this paper, we characterize the minima width by the curvature of loss surface around the minimum~\cite{keskar2016large, chaudhari2019entropy}. Specifically, we use the highest eigenvalue\footnote{We used the opensource implementation at \url{https://github.com/noahgolmant/pytorch-hessian-eigenthings}} of the Hessian of the loss surface at the end of training as a measure of the minima width. Thus, we define one minima as wider than another minima if it has a lower eigenvalue in the direction of the sharpest curvature. This metric of using the highest eigenvalue to measure minima width has been used in several previous papers (\cite{wu2018sgd}, \cite{keskar2016large}).

%To identify the relation between minima width and test accuracy for the above (40-epoch at 0.1 learning rate) runs, we compute their highest eigenvalues. We find that the high test accuracy runs consistently have smaller eigenvalues (wider minima) compared to the low accuracy runs, corroborating the observations of several previous papers~\cite{hochreiter1997flat,keskar2016large,jastrzkebski2017three,wang2018identifying}. For example the run with highest test accuracy of $94.98$ (training loss: $0.00152$) had an eigenvalue of $0.03$, while the run with median accuracy of $94.58$ (training loss: $0.00155$) had an eigenvalue of $0.23$, and the run with minimum accuracy of $94.11$ (training loss: $0.00157$) had an eigenvalue of $0.76$.

\subsection{Hypothesis}

To explain the above observations, i.e., using a high learning rate for {\it short duration} results in {\it low average test accuracy with rare occurrences of high test accuracy}, while using the same high learning rate for {\it long duration} achieves {\it high average test accuracy and frequent occurrences of high test accuracy}, we introduce a new hypothesis. We hypothesize that, \textit{in the DNN loss landscape, the density of narrow minima is significantly  higher than that of wide minima}. 

Intuitively, a large learning rate can escape narrow minima ``valleys'' easily (as the optimizer can jump out of them with large steps). However, once it reaches a wide minima ``valley'', it is likely to get stuck in it (if the ``width'' of the wide valley is large compared to the step size). This intuition is backed by theoretical results from~\citet{xie2020diffusion} that show that the {\it time to escape a minimum using SGD is exponential in the inverse of learning rate as well as inverse of the sharpness} (measured by eigenvalue of the Hessian at the minima). Thus, large learning rates escape narrow minima exponentially faster than wide minima. 

If wide and narrow minima were uniformly distributed, SGD with a large LR would be able to quickly escape the narrow minima, land on a wide minima and get stuck there. Yet, we see that we need to maintain large LR for significant duration for landing in a wide minima with high probability. On the other hand, if our hypothesis is true, i.e., wide minima are much fewer than narrow minima, the probability of landing in a wide minima after escaping a narrow minima is low, and the optimizer needs to take a lot of steps to have a high probability of eventually landing in a wide minimum. Thus, the hypothesis is a better explanation for the observation in Table~\ref{tab:warmup_accuracy_baseline}, where the average accuracy continues to improve as we increase the number of high learning rate training steps. The hypothesis also explains why very few (just 1) of the 50 runs trained at $0.1$ learning rate for just 30-epochs also manages to attain high accuracy---these runs just got lucky in a  probabilistic sense and landed in a wide minimum even with a shorter duration of explore.

\begin{figure*}[t]
  \centering
  \resizebox{\textwidth}{!}{%
  \begin{tabular}{cccc}
    \includegraphics[width=.25\textwidth]{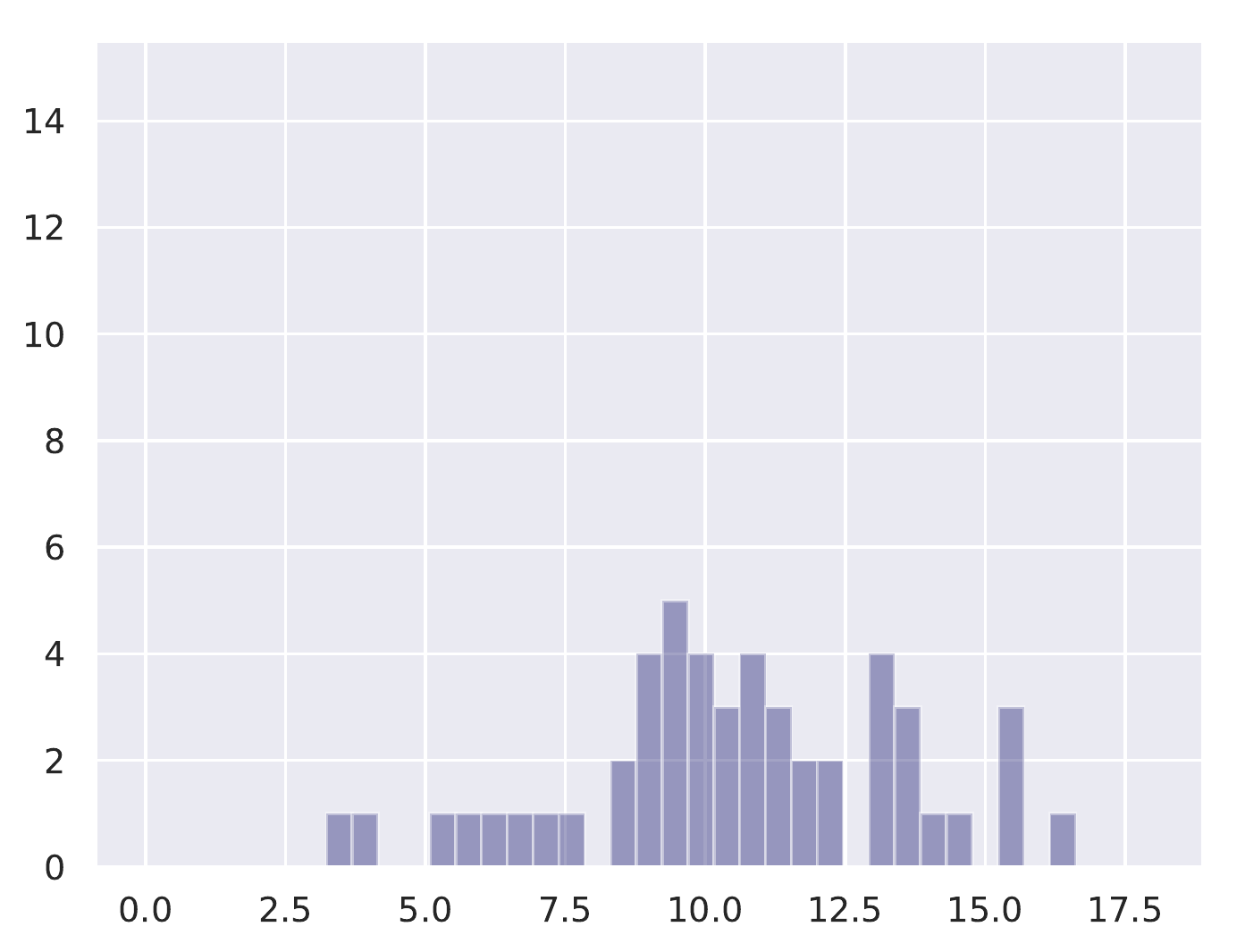}&   
    \includegraphics[width=.25\textwidth]{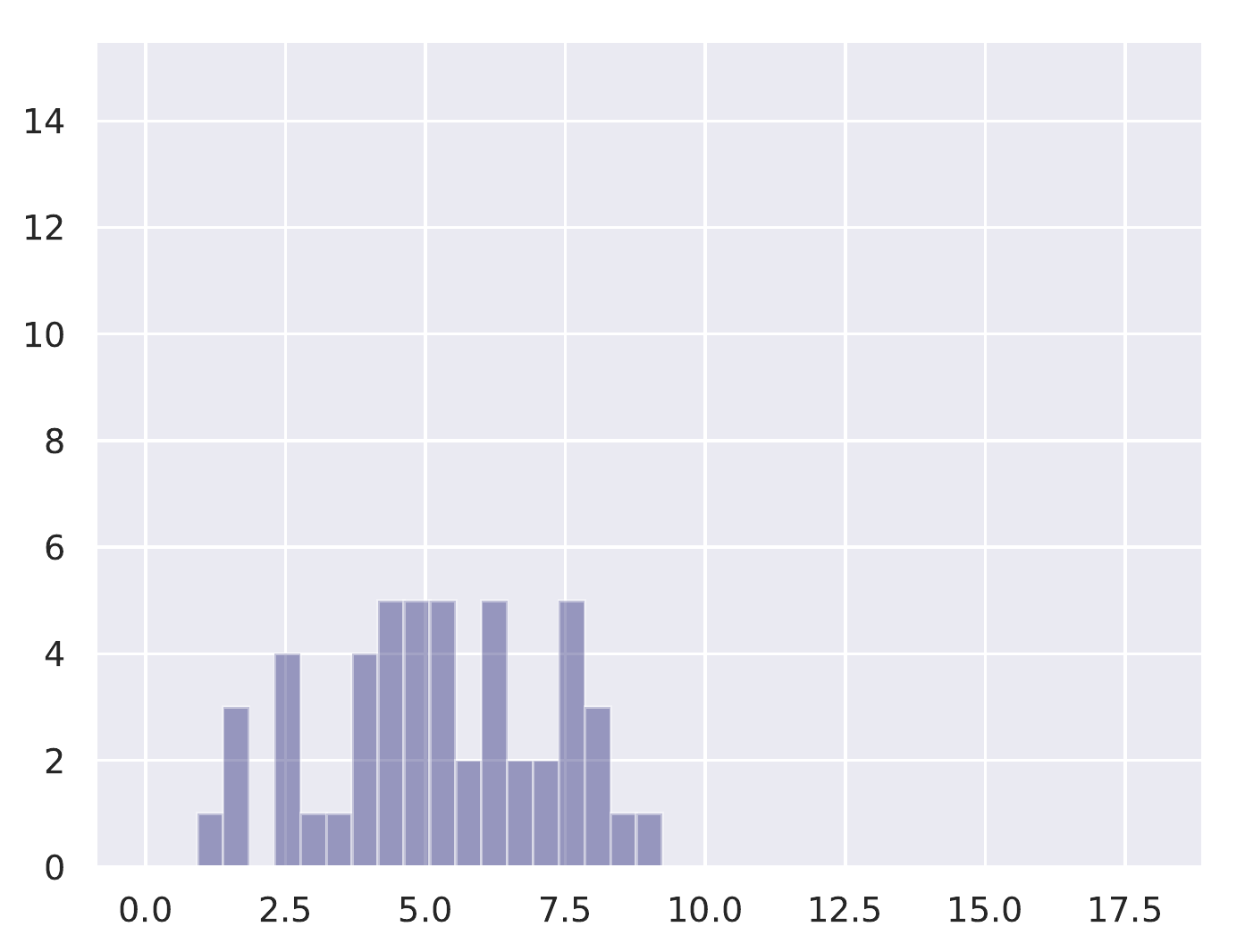}&
    \includegraphics[width=.25\textwidth]{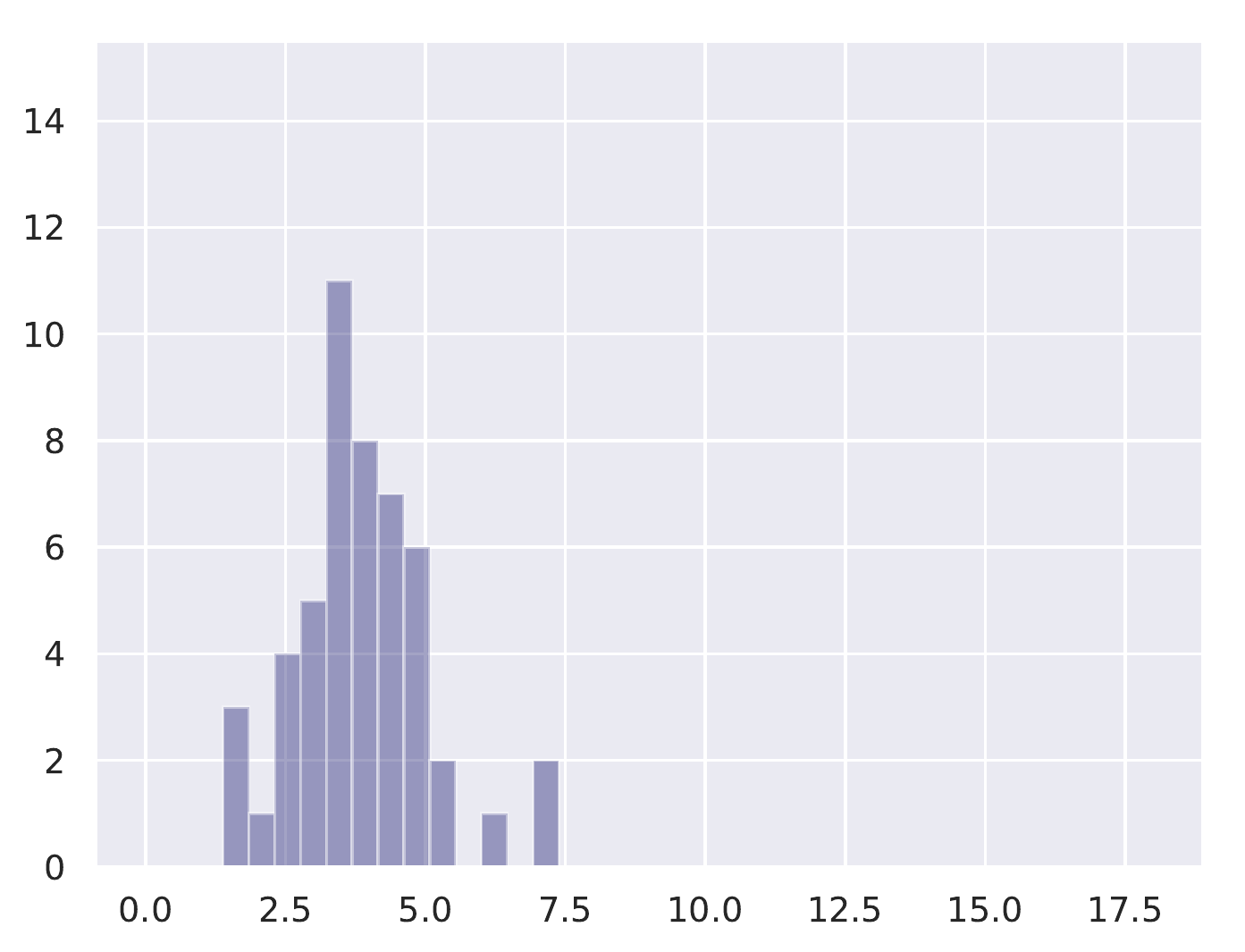}&
    \includegraphics[width=.25\textwidth]{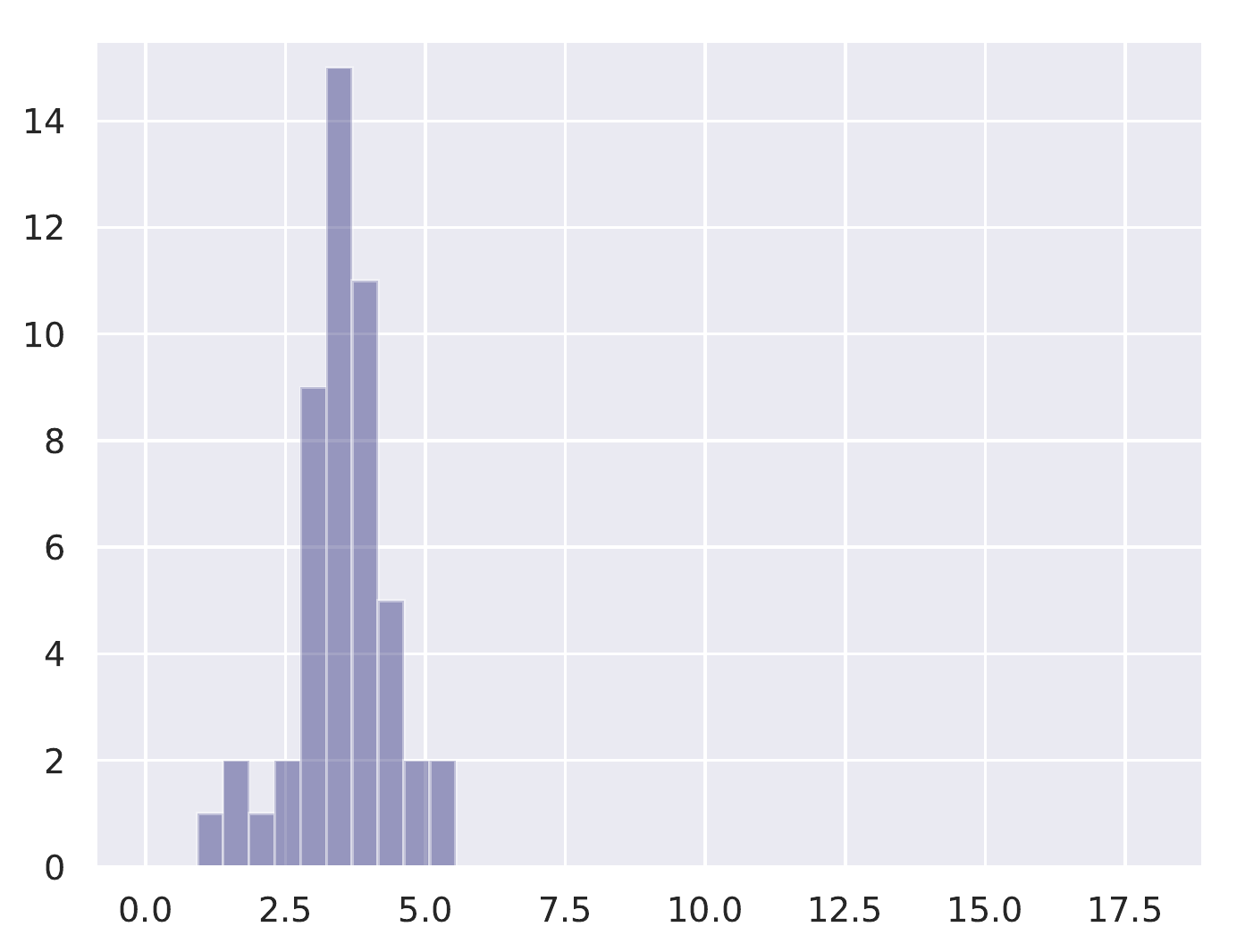}\\
    {\footnotesize (a) Explore 0} & {\footnotesize (b) Explore 30} & {\footnotesize (c) Explore 60} & {\footnotesize (d) Explore 100}
  \end{tabular}
  }
\caption{Histogram of minima sharpness~\citep{keskar2016large} for 50 random trials of Cifar-10 on Resnet-18. Each figure shows histograms for runs with different number of explore epochs. The distribution moves toward lower sharpness and tightens as the number of explore epochs increase.}
  \label{fig:keskar_hist_explores}
 \vspace{6pt} 
 \centering
  \resizebox{\textwidth}{!}{
  \begin{tabular}{cccc}
    \includegraphics[width=0.25\textwidth]{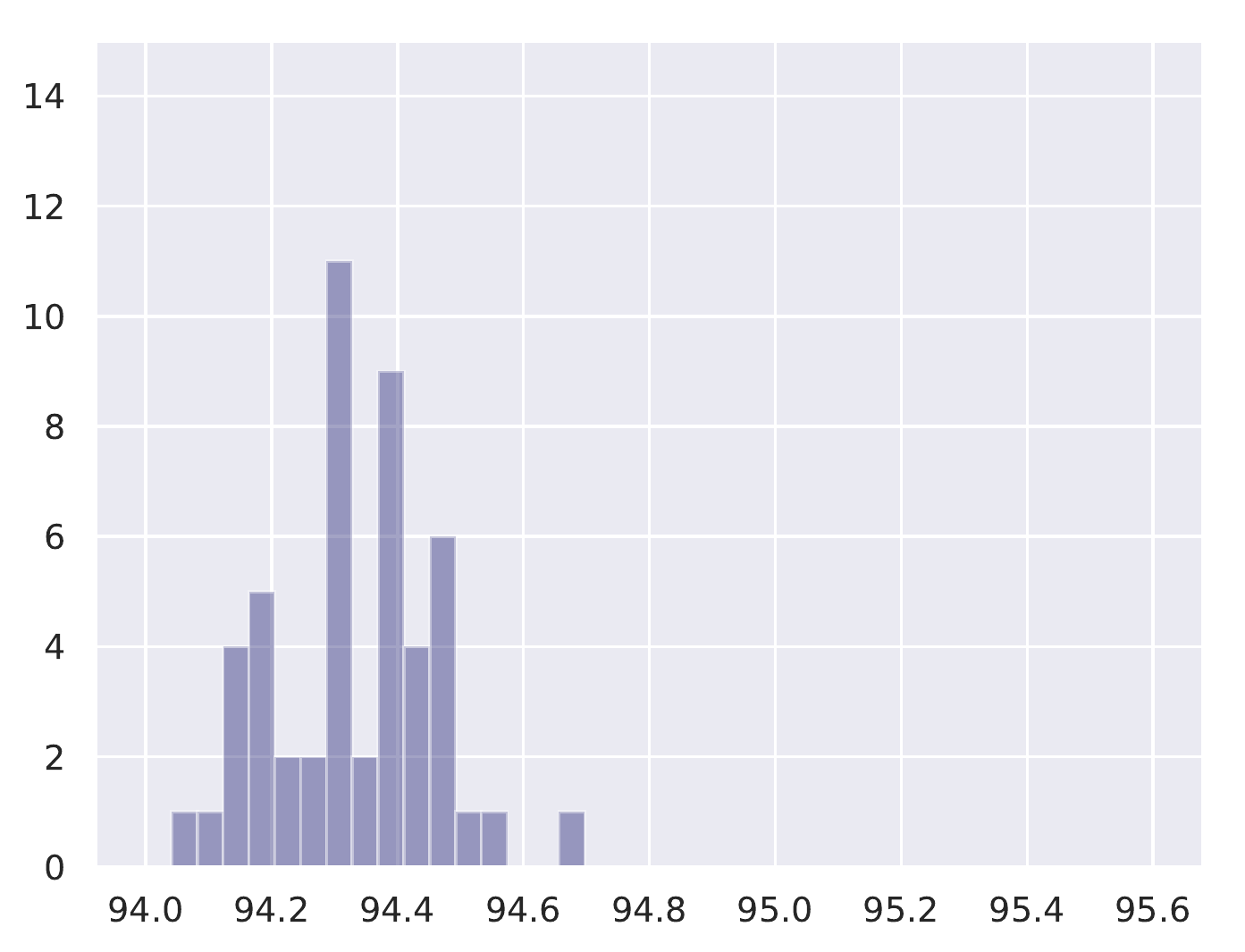}&   
    \includegraphics[width=0.25\textwidth]{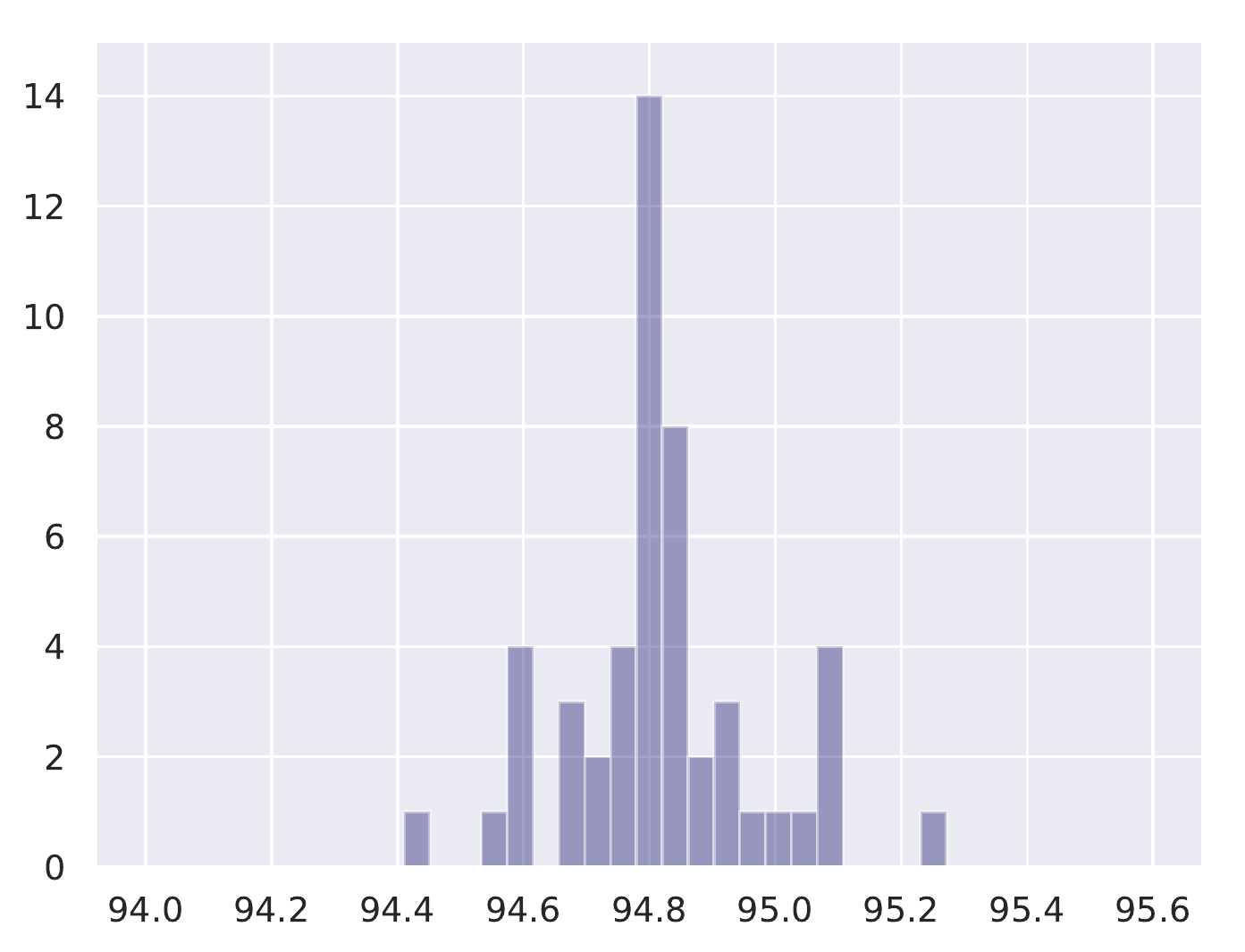}&
    \includegraphics[width=0.25\textwidth]{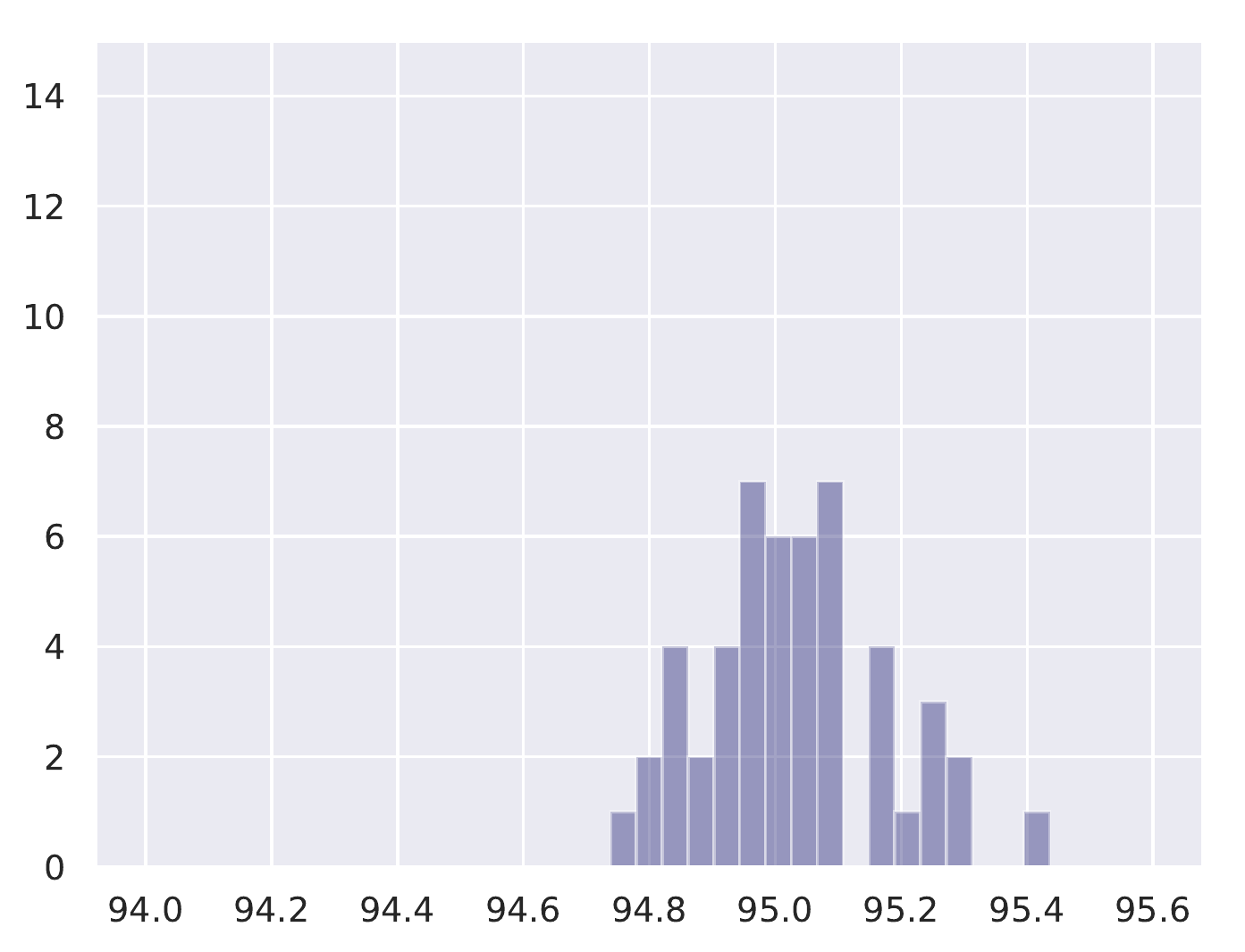}&
    \includegraphics[width=0.25\textwidth]{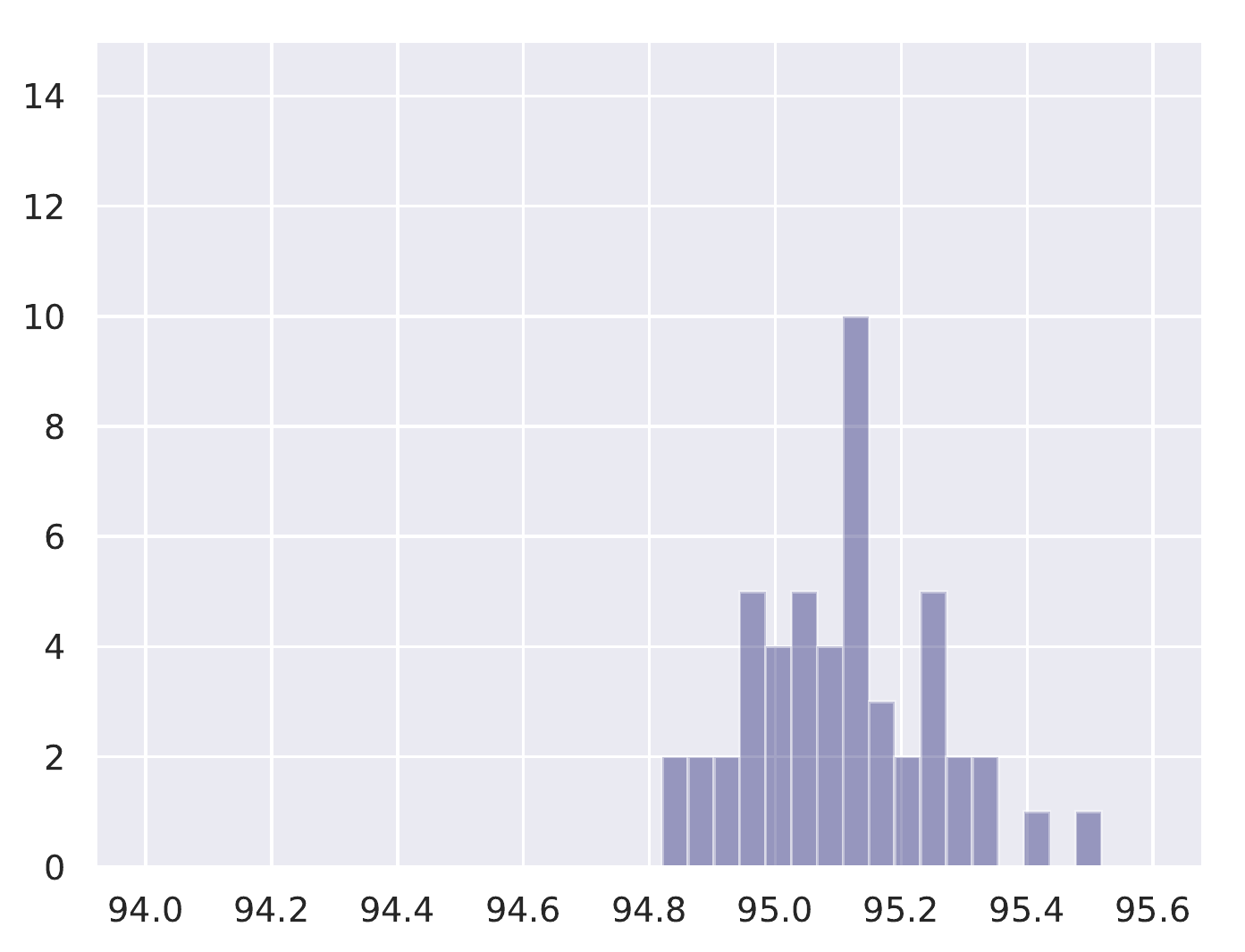}\\
    {\footnotesize (a) Explore 0} & {\footnotesize (b) Explore 30} & {\footnotesize (c) Explore 60} & {\footnotesize (d) Explore 100}
  \end{tabular}
  }
\caption{Histogram of test accuracy for 50 random trials of Cifar-10 on Resnet-18. Each figure shows histograms for runs with different number of explore epochs. The distribution moves toward higher test accuracy and sharpens as the number of explore epochs increase.}
  \label{fig:accuracy_hist_explores}
\end{figure*}

To validate this hypothesis further, we run experiments similar to the one in Table~\ref{tab:warmup_accuracy_baseline}. Specifically, we train Cifar-10 on Resnet-18 model for 200 epochs using a standard step schedule with learning rate of $0.1, 0.01, 0.001$. We vary the number of epochs trained using the high learning rate of 0.1, called the {\it explore epochs}, from 0 to 100 epochs, and divide up the rest of the training equally between 0.01 and 0.001. For each experimental setting, we conduct 50 random trials and plot the distributions of final test accuracy and the minima sharpness as defined by the metric in \cite{keskar2016large} (see section~\ref{sec:metrics_for_wide_minima}). If our hypothesis is true, then the more you explore, the higher the probability of landing (and getting stuck) in a wide minima region, which should cause the distribution to tighten and move towards wider minima (lower sharpness), as the number of explore steps increase. This is exactly what is observed in Figure~\ref{fig:keskar_hist_explores}. Also since wide minima correlate with higher test accuracy, we should see the test accuracy distribution move towards higher accuracy and sharpen, as the number of explore steps increase. This is confirmed as well in Figure~\ref{fig:accuracy_hist_explores}.

\noindent
{\bf Longer training with low learning rate is not sufficient.}
Finally, to verify whether explore at high learning rate is essential, we train Cifar-10 for 10,000 epochs at a fixed lower learning rate of 0.001. The training loss converged but the final test accuracy was only \textbf{93.9}, compared to an accuracy of over 95\% in 200 epochs in Table~\ref{tab:warmup_accuracy_baseline}. Thus, even training $50\times$ longer at low learning rate is not sufficient to achieve good generalization. Again, this observation ties in well with the theoretical results from~\citet{xie2020diffusion} where the authors show that the time to escape a minimum using SGD is exponential in the inverse of learning rate. Thus, {\it this result adds further evidence to our density hypothesis, since even training $50\times$ longer at a low
learning rate is not sufficient to land in a wide minima.}

\vspace{6pt}

%%%% Multi-Scale  %%%%
\noindent
{\bf Multi-scale.} Given the importance of explore at high learning rate, a natural question that may arise is whether explore is necessary at smaller learning rate as well. To answer this, we train the same network for a total of 200 epochs with an initial high learning rate of $0.1$ for 100 epochs, but now we vary the number of epochs trained with the learning rate of $0.01$ (we call this finer-scale explore), and train with learning rate of $0.001$ for the remaining epochs. As can be seen from Table~\ref{tab:finewarmup_accuracy_baseline}, although the final training loss remains similar, we find that finer-scale explore also plays a role similar to the initial explore in determining the final test accuracy. {\it This indicates that our hypothesis about density of wide/narrow regions indeed holds at multiple scales}.

\begin{table}[h]
\small
\centering
\caption{Cifar-10 on Resnet-18 trained for 200 epochs. A learning rate of 0.1 is used for the first 100 epochs. We then vary the number of epochs trained with learning rate of $0.01$ (called finer-scale explore), and train the remaining epochs with a learning rate of $0.001$. We report averages values over 3 runs.}
\label{tab:finewarmup_accuracy_baseline}
%\vspace{3pt}
\begin{tabular}{cccc}
\toprule
Explore Epochs (Finer-scale) & Test Accuracy & Training Loss & Sharpness \\
\midrule
  10 & 94.78 & 0.0031 & 5.48\\ 
  20 & 94.91 & 0.0026 & 4.47\\ 
  30 & 95.00 & 0.0023 & 4.02\\ 
  40 & 95.02 & 0.0021 & 3.91\\ 
  50 & 95.10 & 0.0021 & 3.54\\ 
\bottomrule
\end{tabular}
\end{table}
\vspace{-6pt}

\subsection{Minima Sharpness }
\label{sec:metrics_for_wide_minima}
Our hypothesis predicts that higher explore helps the optimizer land in a wider minimum, which in turn helps generalization. We demonstrated this empirically in Figure~\ref{fig:keskar_hist_explores}, where we plotted the distribution of the minima sharpness, as measured by the sharpness metric introduced by \citep{keskar2016large}. In this section, we describe Keskar's sharpness metric in detail.
%The algorithm of \cite{keskar2016large} to compute this metric, however, did not scale well to large networks.
We also introduce a simple projected gradient ascent scheme to compute this metric efficiently, which scales well to large networks. Finally, we also evaluate our hypothesis with a different metric for minima sharpness, the \textit{Fisher Score}, which is based on the Fisher information matrix.

\subsubsection{Keskar's Sharpness Metric}

Keskar's sharpness metric is based on measuring the maximum jump in the network's output function $F$ in a small neighborhood around the minimum. After a few simplifications, Keskar's metric for sharpness around a point $x$ can be written as:
\begin{equation} \label{eq:keskar_metric}
     S_{x,F}(\epsilon) := \frac{(max_{y \in  C_{\epsilon}(x)} F(x + y)) - F(x)}{1+ F(x)} \times 100,
\end{equation}
where $C_{\epsilon}(x)$ is an $\epsilon$ neighborhood around $x$. \cite{keskar2016large} mentions that under certain conditions and for small values of $\epsilon$, $S_{x,F}$ is proportional to the largest eigenvalue of the Hessian. Please see \cite{keskar2016large} for more details. For our measurements we choose an $\epsilon$ of $1e^{-4}$.

For solving the maximization problem in Equation~\ref{eq:keskar_metric}, \cite{keskar2016large} uses a second-order L-BFGS-B \citep{byrd-lbgfs} optimization scheme. However, in our experiments we found the method to be very slow. To combat this, \cite{keskar2016large} limited their runs to 10 iterations but we found that results were suboptimal using few iterations. Instead, we employed a projected gradient ascent scheme to solve Equation~\ref{eq:keskar_metric}. In each optimization step, we took a small step with a learning rate of 0.001 in the gradient direction and projected the updated point to lie inside $C_{\epsilon}(x)$. Because of the first order nature, this method is much faster. We found that even 1000 iterations were fast to compute and the results were much better than the second order method in all cases we evaluated.

Using Keskar's sharpness metric, we had shown in Figure~\ref{fig:keskar_hist_explores} that the distribution of minima sharpness moves towards lower values as the number of explore epochs increase. In Table~\ref{tab:keskar_avg}, we also report the average sharpness of the minima for varying explores. As predicted by our hypothesis, average sharpness decreases as number of explore epochs increase.

\begin{table}[hbt!]
\begin{minipage}{.6\textwidth}
    \caption{Keskar's sharpness metric for Cifar-10 on Resnet-18 trained for 200 epochs with Momentum. A learning  rate of 0.1 is used for the explore epochs. Half the remaining epochs are trained at 0.01 and the other half at 0.001. We report the average sharpness over 50 different trials.}
\label{tab:keskar_avg}
\end{minipage}\hfill
\begin{minipage}{.35\textwidth}\vspace*{-14pt}%
\begin{tabular}{cc}
  \toprule
  Explore Epochs       &  Sharpness \\
%      &  &  \\ \hline
\midrule
  0 & 10.56\\
  30 & 5.43\\
  60 & 3.86\\
  100 & 3.54\\
 \bottomrule
\end{tabular}
\end{minipage}
\end{table}
\vspace{-15pt}

\subsubsection{\textbf{Fisher Score}}
The maximum Eigen value of the Fisher Information Matrix (FIM) estimates the highest curvature at a point, and is used as another metric to measure minima sharpness~\citep{fim2018information}. We used an unbiased estimate of the true Fisher matrix (see~\cite{empiricalfisher2019limitations}) using 10 unbiased samples per training data. Table~\ref{tab:fim_scores} shows the average Fisher scores for the Cifar-10 experiments at varying explores. Again, the sharpness measured by the Fisher score decreases as the number of explore epochs increase.

\begin{table}[hbt!]
\begin{minipage}{.6\textwidth}
    \caption{Fisher Score for Cifar-10 on Resnet-18 trained for 200
epochs with Momentum. A learning rate of 0.1 is used for the explore epochs. Half the remaining epochs are trained at 0.01 and the other half at 0.001. We report the average Fisher score over 10 different trials.}
\label{tab:fim_scores}
\end{minipage}\hfill
\begin{minipage}{.35\textwidth}\vspace*{0pt}%
\begin{tabular}{cc}
  \toprule
  Explore Epochs      &  FIM score   \\
%      &  &  \\ \hline
\midrule
  0 & 0.051\\
  30 & 0.046\\
  60 & 0.043\\
  100 & 0.042\\
 \bottomrule
\end{tabular}
\end{minipage}
\end{table}

\section{Explore-Exploit Learning Rate Schedule}
\label{explore-exploit schedule}
Given that we need to explore at multiple scales for good generalization, how do we go about designing a good learning rate schedule? The search space of the varying learning rate steps and their respective explore duration is enormous. 

Fortunately, since the explore at the initial scale is searching over the entire loss surface while explore at finer-scales is confined to exploring only the wide-minima region identified by the initial explore, the former is more crucial. In our experiments as well, we found that the initial portion of training is much more sensitive to exploration and needs a substantial number of \textit{explore} steps, while after this initial phase, several decay schemes worked equally well. This is similar to the observations in~\citep{golatkar2019time} where the authors found that regularization such as weight-decay and data augmentation mattered significantly only during the initial phase of training.

The above observations motivate our {\it Explore-Exploit} learning rate schedule, where the \textit{explore} phase first optimizes at a high learning rate for some minimum time in order to land in the vicinity of a wide minima. We should give the \textit{explore} phase enough time (a hyper-parameter), so that the probability of landing in a wide minima is high.
%Since the ratio of number of narrow vs wide minimas can depend on the architecture, dataset, etc., predicting the right explore phase duration is hard, and is currently a hyperparameter for \system{}.
After the \textit{explore} phase, we know with a high probability, that the optimizer is in the vicinity of a wide region. We now start the {\it exploit} phase to descend to the bottom of this wide region while progressively decreasing the learning rate. Any smoothly decaying learning rate schedule can be thought of as doing micro \textit{explore-exploit} at progressively reduced scales. A steady descent would allow more \textit{explore} duration at all scales, while a fast descent would explore less at higher learning rates. We experimented with multiple schedules for the exploit phase, and found a simple linear decay to zero, that does not require any hyper-parameter, to be effective in all the models/datasets we tried. We call our proposed learning rate schedule which starts at a constant high learning rate for some minimum time, followed by a linear decay to zero, the \lrschedule{}.

Note that any learning rate decay scheme incorporates an implicit explore during the initial part, where the learning rate stays high enough. To evaluate the benefit of an explicit explore phase, we compare \lrschedule{} against several decay schemes such as linear and cosine. Interestingly, the results depend on the length of training. For long budget experiments, simple decay schemes perform comparable to \lrschedule{} in some experiments, since the implicit explore duration is also large, helping these schemes achieve good generalization. However for short budget experiments, these schemes perform significantly worse than \lrschedule{}, since the implicit explore duration is much shorter. See Table~\ref{tab:all_results_combined} ,~\ref{tab:all_results_combined_short} and~\ref{tab:target_accuracy_epochs} for the comparison.

\noindent
{\bf Warmup.} Some optimizers such as Adam use an initial warmup phase to slowly increase the learning rate. However, as shown in~\citet{liu2019variance_radam}, learning rate warmup is needed mainly to reduce variance during initial training stages and can be eliminated with an optimizer such as RAdam. Learning rate warmup is also used for large-batch training~\citep{goyal-imagenet-in-an-hour-2017}. Here, warmup is necessary since the learning rate is scaled to a very large value to compensate for the large batch size. This warmup is complementary and can be incorporated into \lrschedule{}. For example, we do this for BERT\textsubscript{LARGE} pretraining experiment where a large 16k batch size was used.

\vspace{-6pt}
\section{Evaluation}
\label{sec:experiments}

In this section we present extensive empirical evaluation of \lrschedule{} on multiple models and datasets across various optimizers, and compare \lrschedule{} against the original hand-tuned learning rate baselines. We first provide an overview of our main results followed by detailed experimental results. We then run further experiments to validate our wide-minima density hypothesis, as well as run sensitivity analysis of seed learning rate on the \lrschedule{}. 

Note that, for completeness, we present a detailed comparison of \lrschedule{} with many other learning rate schedules in literature such as linear decay, cosine decay~\citep{loshchilov2016sgdr}, one-cycle~\citep{smith2018disciplined_onecycle} in Appendix~\ref{sec:extra_baselines}.

\subsection{Experiments}

We evaluate \lrschedule{} on multiple models and datasets spanning both vision and NLP problems. The training of these models spanned various optimizers including SGD Momentum, Adam~\citep{adam_kingma2014adam}, RAdam~\citep{liu2019variance_radam} and LAMB~\citep{bert76lamb}. For all experiments, we used an out of the box policy, where we only change the learning rate schedule, without modifying anything else. We evaluate on multiple image datasets -- Imagenet on Resnet-50, Cifar-10 on Resnet-18; as well as various NLP datasets -- pretraining BERT\textsubscript{LARGE} on Wikipidea+BooksCorpus and fine-tuning it on SQuADv1.1; and WMT'14 (EN-DE), IWSLT'14 (DE-EN) on Transformers.

\subsection{Results Overview}

In all our experiments, we find that \lrschedule{} {\it shows an improvement in test accuracy over the original hand-tuned learning rate baseline as well as various other learning rate schedules in the literature. Further, we also find that \lrschedule{} can achieve the same accuracy as the baseline with a much reduced training budget.}

\begin{table*}[ht]
\small
\centering
{\setlength{\extrarowheight}{1pt}%

\caption{%Test accuracy of the different experiments on all learning rate schedules tried in this paper. 
We report the top-1 accuracy for ImageNet and Cifar-10, BLEU score for IWSLT'14 and WMT'14 and F1 score for BERT on SQuAD. All values are averaged over multiple runs for each experiment.
Experiment details are mentioned in the individual sections of the experiments.}

\label{tab:all_results_combined}

\begin{tabular}{ccc@{\hspace{.95\tabcolsep}}c@{\hspace{.75\tabcolsep}}cccc}
\toprule
%\multirow{3}{*}{Experiment} & Training & \multirow{3}{*}{\shortstack[l]{\textit{Knee} \\ \textit{Schedule}}} & \textit{Knee} & \multirow{3}{*}{Baseline} & \multirow{3}{*}{One-Cycle} & \multirow{3}{*}{\shortstack[l]{Cosine \\ Decay}} & \multirow{3}{*}{\shortstack[l]{Linear \\ Decay}} \\
%& Budget   &  & \textit{Schedule} & & & &  \\
%& (epochs) &                   & (Fixed 50\% explore) & & &  &  \\ 
           &          &               &   \textit{Knee}     &          &           &        &    \\
Experiment & Training & \textit{Knee} &  \textit{Schedule}  & Baseline & One-Cycle & Cosine & Linear \\
           & Budget   & \textit{Schedule} & (Fixed 50\%     &              & Decay  & Decay \\
           & (epochs) &                   &  explore)     &              &         &       \\
\midrule
%ImageNet top5    & 90 & \textbf{93.32}  & 93.33  & 92.90 & 92.56 & 93.28 & 93.21 \\
ImageNet    & 90 & \textbf{76.71}  & 76.58  & 75.87 & 75.39 & 76.41 & 76.54 \\
Cifar-10    & 200 & \textbf{95.26} & 95.26 & 95.10 & 94.09 & 95.23 & 95.18 \\
IWSLT       & 50 & \textbf{35.53}  & 35.23 & 34.97 & 34.77 & 35.21  & 34.97  \\
WMT'14       & 70 & \textbf{27.53} & 27.41 & 27.29 & 27.19 & 27.35  & 27.29  \\ 
BERT\textsubscript{LARGE}    & 31250 (iters) & \textbf{91.51} & 91.51 & 91.34 & - & -  & 91.34  \\ 
\bottomrule
%SQuAD       & 2 & \textbf{81.38}  & 80.89  & 79.9  & 81.31  & 80.89  \\  \hline
\end{tabular}}
\end{table*}

\begin{table*}[ht]
\small
\centering
{\setlength{\extrarowheight}{1pt}%

\caption{Shorter budget training: Test accuracy on all learning rate schedules tried in this paper, but trained with a shortened budget. We report same metrics as Table~\ref{tab:all_results_combined}. \lrschedule{} achieves the same accuracy as baseline schedules using much lower budget, saving precious GPU-hours.}

%e.g. 1002 V100 GPU-hours for BERT\textsubscript{LARGE} pre-training. All values are averaged over multiple runs.}

\label{tab:all_results_combined_short}

\begin{tabular}{ccccccc}
\toprule
           & Shortened Training & \textit{Knee} &    & Cosine  & Linear  & Saving\\
Experiment &    Budget          & \textit{Schedule} & One-Cycle & Decay & Decay    & ( V100 GPU\\
           &    (epochs)        &                   &  &       &        &    hours)\\
\midrule
% ImageNet top5 (30 explore)   &  50 & \textbf{92.89} & 92.53 & 92.81 & 92.84 & 27\\
%ImageNet top1 (30 explore)  &  50 & \textbf{75.77} & 75.36 & 75.71 & 75.82 & 27\\
ImageNet     &  50 & \textbf{75.92} & 75.36 & 75.71 & 75.82 & 27\\
Cifar-10     & 150 & \textbf{95.14} & 93.84 & 95.06 & 95.02 & 0.25\\
IWSLT        & 35 & \textbf{35.08} & 34.43 & 34.46  & 34.16 & 0.75\\
WMT'14       & 30 & \textbf{27.28} & 26.80 & 26.95  & 26.77 & 80\\
BERT\textsubscript{LARGE} & 20854 (iters) & \textbf{91.29}  & - & -  & 90.64  & 1002 \\
\bottomrule
\end{tabular}}
\end{table*}

\begin{table*}[!th]
\small
\centering
\caption{Epochs required by different LR schedules to reach the target accuracy. The target accuracy is chosen based on \lrschedule{}'s results with a reduced budget.}
{\setlength{\extrarowheight}{1pt}%
\begin{tabular}{ccccc}
\toprule
Experiment & Target BLEU Score & \lrschedule{} & Cosine Decay & Linear Decay \\ 
\midrule
IWSLT       & 35.08 & 35 & 45 & 60 \\
WMT'14      & 27.28 & 30 & 60 & 70  \\ 
\bottomrule
\end{tabular}}

\label{tab:target_accuracy_epochs}
\end{table*}

Table~\ref{tab:all_results_combined} shows the test accuracies of the various experiments, when trained with the original budget; while Table~\ref{tab:all_results_combined_short} shows the results when trained with a reduced budget. As shown, for the original budget runs, \lrschedule{} improves on the test accuracies in all experiments.  Note that in \lrschedule{}, the explore duration is a hyperparameter. To avoid tuning this hyperparameter, we experimented with a fixed 50\% explore duration for the full budget runs. Even the fixed 50\% explore \lrschedule{} outperforms all the other baselines. Also noteworthy is that \lrschedule{} is able to achieve the same test accuracies as the baseline's full budget runs with a much lower training budget, saving precious GPU cycles (Table~\ref{tab:all_results_combined_short}).

While the difference in accuracy values between the various schedules might appear deceptively small in absolute terms, achieving these gains require a large amount of compute. For example, 
the number of epochs needed by each scheme to reach the target BLEU score for IWSLT'14 DE-EN and WMT'14 EN-DE with the Transformer network is shown in Table~\ref{tab:target_accuracy_epochs}. One can see that \lrschedule{} is significantly more efficient as compared to say Cosine Decay, which takes 100\% more training time to achieve the same accuracy for WMT`14 EN-DE. Thus, the accuracy and/or compute gains achieved by \lrschedule{} is significant.

A summary of our main experimental results is as follows:
\begin{enumerate}%[nosep]
 \item Imagenet on Resnet-50: We show an absolute gain of 0.8\% in top-1 accuracy against the competitive step schedule baseline for this model. Also, \lrschedule{} can achieve the same accuracy as baseline in $\sim$45\% less training epochs.
 %\item Cifar-10 on Resnet-18: We beat the competitive baseline by 0.16\% and can achieve the same accuracy as baseline in 25\% less epochs.
 \item BERT\textsubscript{LARGE} pre-training on  Wikipedia+BooksCorpus dataset: Compared to the baseline of \cite{bert76lamb}, we improve the F1 score on SQuAD v1.1 fine-tuning task by 0.2\% (91.51 compared to 91.34). Also, we were able to achieve similar accuracy as baseline in 33\% less training steps (a saving of $\sim$1002 V100 GPU-hours!).
 \item WMT'14 and IWSLT machine translation on Transformers: Compared to competitive baselines, we were able to improve the BLEU scores by 0.24 and 0.56 points for the two tasks. Moreover, \lrschedule{} was able to achieve the same accuracy as baselines in 57\% and 30\% less training times.
 \item State of the Art (SOTA) Results: We also attain state of the art results on the IWSLT'14(DE-EN) machine translation dataset by simply replacing the learning rate schedule of the current SOTA model \citep{shen2020simple} with \lrscheduleshort{}. We were able to improve the BLEU score by 0.18, reaching a new SOTA score of 37.78. Moreover, \lrscheduleshort{} can achieve the current SOTA baseline value in 30\% less training time.
\end{enumerate}

\subsection{Detailed Results}

We now describe each of our main experimental results in detail.

\subsubsection{ImageNet Image Classification on Resnet-50}

We train ImageNet dataset \citep{imagenet-dataset} on Resnet-50 network \citep{resnet_he_2016} which has 25 million parameters, with a batch size of 256 and a seed learning rate of 0.1. Random cropping and random horizontal flipping augmentations were applied to the training dataset.  We use SGD optimizer with momentum of 0.9 and weight decay of $1e^{-4}$.
For baseline runs, we used the standard hand-tuned step learning rate schedule of 0.1, 0.01 and 0.001 for 30 epochs each. For \lrschedule{} we used a seed learning rate of 0.1 (same as baseline). We trained with the original budget of 90 epochs as well as with a reduced budget of 50 epochs. We used 30 explore epochs for the two experiments. \footnote{We used the opensource implementation at: https://github.com/cybertronai/imagenet18\_old}

Table~\ref{tab:imagenet_test_training_loss} shows the training loss and test accuracies for our experiments. \lrschedule{} comfortably beats the test accuracy of baseline in the full budget run
(with absolute gains of 0.8\% and 0.4\% in top-1 and top-5 accuracy, respectively), while meeting the baseline accuracy even with a much shorter budget. The fact that the baseline schedule takes almost $80\%$ more training time than \lrschedule{} for the same test accuracy, shows the effectiveness of our \textit{Explore-Exploit} scheme.
See Figure~\ref{fig:imagenet_momentum_result} in Appendix~\ref{sec:detailed_plots} for training curves.

\begin{table}[h]
\small
\centering
\caption{ImageNet on Resnet-50 results. We report mean (stddev) over 3 runs.}
%\vspace{4pt}
\label{tab:imagenet_test_training_loss}
\begin{tabular}{ccccc}
\toprule
\multirow{1}{*}{LR Schedule}  & Test Top 1 Acc. & Test Top 5 Acc. & Training Loss & Training Epochs\\
 % & Loss & Accuracy & Accuracy \\
\midrule
  Baseline     & \multirow{1}{*}{75.87 (0.035)} & \multirow{1}{*}{92.90 (0.015)} & \multirow{1}{*}{0.74 (1e-3)} & 90\\
  \lrscheduleshort{}     & \multirow{1}{*}{76.71 (0.097)} & \multirow{1}{*}{93.32 (0.031)} & \multirow{1}{*}{0.79 (1e-3)} & 90\\
%
%  \lrscheduleshort{} (short budget) (30 explore)    & \multirow{1}{*}{75.77 (0.202)} & \multirow{1}{*}{92.89 (0.049)} & \multirow{1}{*}{0.93 (9e-4)} & 50\\
%
  \lrscheduleshort{} (short budget)    & \multirow{1}{*}{75.92 (0.11)} & \multirow{1}{*}{92.90 (0.085)} & \multirow{1}{*}{0.90 (3e-3)} & 50\\
%
%  Linear Decay (full budget)    & \multirow{1}{*}{76.54 (0.155)} & \multirow{1}{*}{93.21 (0.051)} & \multirow{1}{*}{0.75 (0.001)} & 90\\
\bottomrule
\end{tabular}
\end{table}

\subsubsection{Cifar-10 Image Classification on Resnet-18}

We train Cifar-10 dataset~\citep{cifar-10-dataset} on Resnet-18 network~\citep{resnet_he_2016} which has around 11 million parameters. SGD optimizer is used with momentum of 0.9 and weight decay of $5e^{-4}$. Random cropping and random horizontal flipping augmentations were applied to the training dataset. \footnote{We used the open-source implementation at: https://github.com/kuangliu/pytorch-cifar}. 

For baseline, we used the hand-tuned step learning rate schedule of 0.1, 0.01 and 0.001 for 100, 50, 50 epochs, respectively. With \lrschedule{}, we train the network with the original budget of 200 epochs, as well as a reduced budget of 150 epochs. We used 100 explore epochs for both runs, and a seed learning rate of 0.1 (same as baseline).
Table~\ref{tab:cifar_results_train_loss_test_acc} shows the training loss and test accuracies for the experiments. \lrschedule{} beats the test accuracy of baseline in the full budget run, while meeting the baseline test accuracy in $25\%$ less budget. Refer to figure~\ref{fig:cifar_momentum_result} in Appendix~\ref{sec:detailed_plots} for detailed comparisons of training loss, test accuracy, and learning rate.

\begin{table}[h]
    \small
\centering
\caption{Training loss and Test accuracy for Cifar-10 on Resnet-18. We report mean (stddev) over 7 runs.} 
\label{tab:cifar_results_train_loss_test_acc}
\begin{tabular}{cccc}
\toprule
\multirow{1}{*}{LR Schedule} & Test Accuracy & Training Loss & Training Epochs \\
\midrule
  Baseline   & \multirow{1}{*}{95.10 (0.14)}  & \multirow{1}{*}{0.002 (1e-4)}  & 200 epochs \\ 

  \lrscheduleshort{}     & \multirow{1}{*}{95.26 (0.11)} &  \multirow{1}{*}{0.002 (1e-4)} & 200 epochs \\

  \lrscheduleshort{} (short budget)     & \multirow{1}{*}{95.14 (0.18)} &  \multirow{1}{*}{0.004 (3e-4)}  & 150 epochs \\  
\bottomrule
\end{tabular}
\end{table}

\subsubsection{BERT\textsubscript{LARGE} Pre-training}

We pretrain on BERT\textsubscript{LARGE} on  Wikipedia+BooksCorpus dataset with LAMB optimizer (\cite{bert76lamb}).
BERT\textsubscript{LARGE} has around 330 million parameters and the pre-training is divided into two phases with different sequence lengths. The first phase consists of 90\% steps with sequence length of 128 and the second phase consists of the remaining 10\% steps with sequence length of 512 (\cite{devlin2018bert}). We used a batch size of 16384 in both phases of training \footnote{We used the open-source implementation at: \\https://github.com/NVIDIA/DeepLearningExamples/tree/master/PyTorch/LanguageModeling/BERT}. We use the same training budget of 31250 steps  mentioned in (\cite{bert76lamb}). We also train the model on a shortened training budget of 2/3\textsuperscript{rd} the original steps (20854 steps).  

Since large batch training requires learning rate warmup (see \citet{goyal-imagenet-in-an-hour-2017}), we incorporate it into the \lrschedule{} by first doing a warmup of 10\% as suggested in \citep{bert76lamb} followed by the explore-exploit phases. We used an explore of 50\% of the total steps available for both phases of BERT training. For baseline, we use the warmup (10\%) + linear decay (90\%)  schedule~\citep{bert76lamb,devlin2018bert}. The pre-trained models are evaluated on the SQuAD v1.1~\citep{rajpurkar2016squad} dataset by fine-tuning on the dataset for 2 epochs.  See Table~\ref{tab:bert_pretraining} for the results. For the full budget run, \lrschedule{} improves the baseline by ~0.2\%, while for the reduced budget we achieved similar fine-tuning accuracy as baseline. The baseline schedule achieves a much lower accuracy with shorter budget training, showing the efficacy of \lrschedule{}. BERT pre-training is extremely compute expensive and takes around 47 hours on 64 V100 GPUs (3008 V100 GPU-hrs) on cloud VMs. The reduced budget amounts to a saving of approximately 1002 V100 GPU-hours!

\vspace{-4pt}
\begin{table}[h]
\small
\centering
\caption{BERT\textsubscript{LARGE} results. We report the pre-training train loss, and the test F1 accuracy on SQuAD v1.1 after fine-tuning. See figure~\ref{fig:bert_plots} in Appendix~\ref{sec:detailed_plots} for training curves.}
\label{tab:bert_pretraining}
%\vspace{4pt}
\begin{tabular}{cccc}
\toprule
LR Schedule  & F1 score on SQuAD v1.1 & Training loss & Total Training Steps \\ 
\midrule
Knee & 91.51  & 1.248 & 31250\\
Baseline \citep{bert76lamb} &  91.34 & - & 31250 \\
Baseline (short budget)   & 90.64  & 1.336 & 20854\\
\lrscheduleshort{} (short budget)  & 91.29 & 1.275 & 20854\\
\bottomrule
\end{tabular}
\end{table}

\subsubsection{Machine Translation on Transformer Network with WMT'14 and IWSLT}

In the second NLP task, we train the Transformer (base model)~\citep{vaswani2017attention} on the IWSLT'14 (De-En)~\citep{iwslt_dataset_cettolo2014} and  WMT'14 (En-De)~\citep{wmt14translation}   datasets with the RAdam \citep{liu2019variance_radam} optimizer.
%\footnote{We used the implementation at: https://github.com/pytorch/fairseq~\cite{ott2019fairseq}}. 

\paragraph{WMT'14 (EN-DE):}We use the default implementation provided by the fairseq package \citep{ott2019fairseq} \footnote{https://github.com/pytorch/fairseq}. We train WMT'14 (EN-DE) dataset on the Transformer\textsubscript{BASE}~\citep{vaswani2017attention} model which has around 86 million parameters and use the RAdam~\citep{liu2019variance_radam} optimizer with $\beta_{1}$ of 0.9 and $\beta_{2}$ of 0.999. Label smoothed cross entropy was used as the objective function with an uncertainty of 0.1. A dropout of 0.1, clipping norm of 25 and weight decay of $1e^{-4}$ is used. Each training batch contains approximately 30000 tokens.

The baseline schedule uses a linear decay for 70 epochs \citep{liu2019variance_radam}. With \lrschedule{}, we trained with the original budget of 70 epochs, as well as a reduced budget of 30 epochs. We used 50 and 25 explore epochs for the two runs, respectively and a seed learning rate of $3e^{-4}$ for both \lrschedule{} and baseline. In all cases we use the model checkpoint with least loss on the validation set for computing BLEU scores on the test set. Table~\ref{tab:wmt_results} shows the training loss and test accuracy averaged over 3 runs. \lrschedule{} improves the test BLEU score of baseline in the full budget run by 0.24 points. In the shorter budget run, \lrschedule{} matches the test accuracy of the baseline while taking $57\%$ less training time (a saving of 80 V100 GPU-hours!). See Figure~\ref{fig:wmt_radam_lrl} in Appendix~\ref{sec:detailed_plots} for training curves.

\paragraph{IWSLT'14 (DE-EN):} For IWSLT'14 (DE-EN) we use the same configuration as WMT'14 (EN-DE), except for a dropout of 0.3 following Fairseq's out-of-box implementation. Each training batch contains approximately 4000 tokens. For \lrschedule{}, we choose explore as 30 epochs for short budget runs and 40 epochs for full budget runs.

The baseline schedule uses a linear decay for 50 epochs \citep{liu2019variance_radam}. With \lrschedule{}, we trained with the original budget of 50 epochs, as well as a reduced budget of 35 epochs. We used 40 and 30 explore epochs for the two runs, respectively and a seed learning rate of $3e^{-4}$ for both \lrschedule{} and baseline. In all cases we use the model checkpoint with least loss on the validation set for computing BLEU scores on the test set. \lrschedule{} improves the baseline test BLEU score by 0.56 points in the full budget run. In the shorter budget run, \lrschedule{} matches the test accuracy of the baseline schedule while taking $30\%$ less training time. See Figure~\ref{fig:iwslt_adam_result} in Appendix~\ref{sec:detailed_plots} for training curves.

\vspace{-4pt}
\begin{table}[h]
\captionsetup{type=table} %% tell latex to change to table
\small
\centering
\caption{Results for WMT'14 (EN-DE) on Transformer networks. The test BLEU scores are computed on the checkpoint with the best validation perplexity. We report mean (stdev) over 3 runs.}
%\vspace{4pt}
\label{tab:wmt_results}
\begin{tabular}{ccccc}
\toprule
  \multirow{2}{*}{LR Schedule} & Test BLEU  & Train & Validation  & Training \\
  &  Score & Perplexity  & Perplexity & Epochs \\ 
 \midrule
  Baseline   &  27.29 (0.06) & 3.87 (0.017) & 4.89 (0.02) & 70 \\
  \lrscheduleshort{}  & 27.53 (0.12) & 3.89 (0.017) & 4.87 (0.006) & 70 \\ 
  \lrscheduleshort{} (short budget) & 27.28 (0.17) & 4.31 (0.02) & 4.92 (0.007)  & 30\\
\bottomrule
\end{tabular}

\end{table}
\vspace{10pt}
\begin{table}[th]
  \centering
  \small
\caption{Training, validation perplexity and test BLEU scores for IWSLT on Transformer networks. The test BLEU scores are computed on the checkpoint with the best validation perplexity. We report the mean and standard deviation over 3 runs.} 
\label{tab:iwslt_results}
\begin{tabular}{ccccc}
\toprule
  \multirow{2}{*}{LR Schedule} & Test BLEU  & Train & Validation  & Training \\
  &  Score & Perplexity  & Perplexity & Epochs \\ 
 \midrule
  Baseline   &  34.97 (0.035) & 3.36 (0.001) & 4.91 (0.035) & 50 \\
  \lrscheduleshort{}  & 35.53 (0.06) & 3.00 (0.044) & 4.86 (0.02) & 50 \\ 
  \lrscheduleshort{} (short budget) & 35.08 (0.12) & 3.58 (0.049) & 4.90 (0.063)  & 35\\

\bottomrule
\end{tabular}
\end{table}

\subsubsection{SQuAD-v1.1 fine-tuning on BERT\textsubscript{BASE}}

We also evaluate \lrschedule{} on the task of fine-tuning BERT\textsubscript{BASE} model~\cite{devlin2018bert} on SQuAD v1.1~\cite{rajpurkar2016squad} with the Adam~\cite{kingma2014adam} optimizer \footnote{We used the implementation at: https://github.com/huggingface/transformers}. BERT fine-tuning is prone to overfitting because of the huge model size compared to the small fine-tuning dataset, and is typically run for only a few epochs. For baseline we use the linear decay schedule mentioned in~\cite{devlin2018bert}. We use a seed learning rate of $3e^{-5}$ and train for 2 epochs. For \lrschedule{}, we train the network with 1 explore epoch with the same seed learning rate of $3e^{-5}$. Table~\ref{tab:squad_results} shows our results over 3 runs. We achieve a mean EM score of 81.4, compared to baseline's 80.9, a 0.5\% absolute improvement. We don't do a short budget run for this example, as the full budget is just 2 epochs. Please refer to Figure~\ref{fig:squad_bert_adam_result} in Appendix~\ref{sec:detailed_plots} for the training loss, test accuracy and learning rate curves.

\begin{table}[h]
\small
\centering
\caption{SQuAD fine-tuning on BERT\textsubscript{BASE}. We report the average training loss, and average test EM, F1 scores over 3 runs.}
\label{tab:squad_results}
{\setlength{\extrarowheight}{1pt}%
\begin{tabular}{cccccc}
  \toprule
  LR Schedule & EM  & F1 & Train Loss & Training Epochs \\
  \midrule
  Baseline   & 80.89 (0.15) & 88.38 (0.032) & 1.0003 (0.004) & 2\\
  \lrschedule{}{}   & 81.38 (0.02) & 88.66 (0.045) & 1.003 (0.002) & 2 \\
  \bottomrule
\end{tabular}}

\end{table}

\subsubsection{State of the Art Result}
To further demonstrate the effectiveness of \lrschedule{}, we took a recent high performing model, Cutoff~\citep{shen2020simple}\footnote{We used the code available at https://github.com/dinghanshen/Cutoff}, which had reported state-of-the-art accuracy on the IWSLT'14 (DE-EN) dataset. They reported a BLEU score of 37.6 when trained with an inverse square root learning rate schedule for 100 epochs, with the first 6000 steps allocated for warmup. We simply retrained the model with our \lrschedule{}, and achieved a new SOTA BLEU score of 37.78 (an absolute increase of 0.18). See Table~\ref{tab:iwslt_cutoff_results_train_loss_test_acc} for the BLEU scores, training and validation perplexities.

We also show that \lrschedule{} can train the model in 30\% less training time (70 epochs), while achieving slightly better accuracy of 37.66 BLUE score compared to the 100 epoch baseline. The baseline schedule when run for 70 epochs achieves a much worse accuracy of 37.31.

For both the full budget (100 epochs) and the short budget (70 epochs) \lrscheduleshort{} runs, we choose 50\% of the total training epochs as explore epochs. We also perform warmup for the same number of steps as baseline. For all runs (\lrscheduleshort{} and baseline), we report the BLEU score obtained by averaging the last 5 checkpoints and computing on the test set. See Figure~\ref{fig:Cutoff_adam_result} and~\ref{fig:Cutoff_short_adam_result} in Appendix~\ref{sec:detailed_plots} for training curves.
 
\begin{table}[h]
\small
\centering
\caption{Training, validation perplexity and test BLEU scores for IWSLT'14 DE-EN on Cutoff. The test BLEU scores are computed by averaging the last 5 checkpoints}
\label{tab:iwslt_cutoff_results_train_loss_test_acc}
{\setlength{\extrarowheight}{1pt}%
\begin{tabular}{cccccc}
\toprule
  \multirow{2}{*}{LR Schedule} & Test BLEU  & Train & Validation  & Training \\
  &  Score & Perplexity  & Perplexity & Epochs \\ 
 \midrule
  Inv. Sqrt  &  37.60   & 3.46  & 4.24  & 100  \\
  \lrscheduleshort{}  & 37.78   & 3.29   & 4.13  & 100  \\ 
 Inv. Sqrt (short budget)  & 37.31   & 3.76  & 4.29  & 70 \\ 
 \lrscheduleshort{} (short budget)  & 37.66   &  3.48  & 4.18  & 70 \\
 %
% \lrscheduleshort{} (short budget)  & 37.53   &    &   & 60 \\
 %
% Inv. Sqrt (short budget)  & 37.25   &   &   & 60 \\ 
\bottomrule 
\end{tabular}}
\end{table}

\subsection{Hypothesis Validation with \lrschedule{} on Language Tasks}
\label{sec:hypothesis_validation}

For validating our hypothesis on the density of wide minima vs narrow minima, we did multiple experiments on vision tasks, most of which were discussed in Section~\ref{sec:wide_minima_hypothesis}.
To summarize, in Figures~\ref{fig:keskar_hist_explores} and \ref{fig:accuracy_hist_explores}, we showed that for Cifar-10 on Resnet-18, as the number of explore steps increase, the distribution of minima width and test accuracy sharpens and shifts towards wider minima and better accuracy, respectively. 

% repeats earlier section??
%This behaviour is predictable from our hypothesis as increasing explore steps increases the probability of landing in a wide region. From the same argument, the average accuracy should increase as the number of explore steps increase, which is confirmed in Table~\ref{tab:warmup_accuracy_baseline}. Our hypothesis also predicts that even at low explore epochs, although the probability of landing in a wide region is low, it is non zero. Thus, out of many trials with low number of explore epochs, a few runs could still yield high test accuracy. This is what we observe in Figure~\ref{fig:accuracy_hist_explores}(b), where 1 out of 50 trials for the 30 explore runs obtains an accuracy more than $95.24$ even though the average accuracy for 30 explore is $94.81$!.

\begin{table}[h]
\small
\centering
\caption{IWSLT'14 (DE-EN) on the Transformer network trained with the \lrschedule{}. The \textit{explore} duration is varied, while keeping the total training budget fixed at 50 epochs. We report averages over 3 runs.}

\begin{tabular}{ccc}
  \toprule
  Explore Epochs & Test BLEU score & Training Perplexity \\
  \midrule
%  &  & \\ 
  5  & 34.93 & 3.29 \\ 
  10 & 35.02 & 3.22 \\ 
  15 & 35.08 & 3.11 \\ 
  20 & 35.10  & 3.08 \\ 
  25 & 35.23 & 3.02 \\ 
  30 & 35.28 & 2.99 \\
  40 & 35.53 & 3.00 \\
  \bottomrule
\end{tabular}
\label{tab:warmup_accuracy_iwslt}
\end{table}

We now perform similar experiments on the IWSLT'14 German to English dataset \citep{iwslt_dataset_cettolo2014} trained on Transformer networks \citep{vaswani2017attention} to demonstrate that our hypothesis holds even on a completely different NLP dataset and network architecture. We train with the \lrschedule{} for a total budget of 50 epochs with explore lr as $3e^{-4}$, but keep varying the number of explore epochs. As shown in Table~\ref{tab:warmup_accuracy_iwslt}, the test BLEU score increases as we increase the number of explore epochs. Further, we found that among multiple trials, a 20 epoch explore run had a high BLEU score of \textbf{35.29}, suggesting that the run got lucky. Thus, these results on the IWSLT'14 (DE-EN) dataset add more evidence to the wide-minima density hypothesis. 

%Also, see section~\ref{sec:seed_sensitivity} and Table~\ref{tab:seed_senstivity_cifar} for a sensitivity analysis of the \lrschedule{} on the starting learning rate. Interestingly, we found that the optimal explore duration varies inversely with the starting learning rate. Since a bigger learning rate has higher probability of escaping narrow minima compared to a lower learning rate, it would, on an average, require fewer steps to land in a wide minima. Thus, larger learning rates can explore faster. This observation is thus consistent with our hypothesis and further corroborates it.
\vspace{6pt}
\subsection{Learning Rate Sensitivity for \lrschedule{}}
\label{sec:seed_sensitivity}

We performed sensitivity analysis of the starting learning rate, referred to as the seed learning rate, for \lrschedule{}. We trained the Cifar-10 dataset on Resnet-18 with the \lrschedule{} for a shortened budget of 150 epochs, starting at different seed learning rates. For each experiment, we do a simple linear search to find the best explore duration. The test accuracies and optimal explore duration for the different seed learning rate choices is shown in Table~\ref{tab:seed_senstivity_cifar}. As shown, the seed learning rate can impact the final accuracy, but \lrschedule{} is not highly sensitive to it. In fact, we can achieve the target accuracy of 95.1 with multiple seed learning rates of 0.05, 0.075, 0.0875 and 0.115, as compared to the original seed learning rate of 0.1, by tuning the number of explore epochs.

Another interesting observation is that the optimal explore duration varies inversely with the seed learning rate. Since a bigger learning rate has higher probability of escaping narrow minima compared to a lower learning rate, it would, on an average, require fewer steps to land in a wide minima. Thus, larger learning rates can \textit{explore} faster, and spend more time in the \textit{exploit} phase to go deeper in the wide minimum. This observation is thus consistent with our hypothesis and further corroborates it.

We also note that by tuning both seed learning rate and explore duration, we can achieve the twin objectives of achieving a higher accuracy, as well as a shorter training time -- e.g. here we are able to achieve an accuracy of 95.34 in 150 epochs (seed learning rate 0.075), compared to 95.1 achieved by the baseline schedule in 200 epochs.

\begin{table}[h]
\small
\centering
\caption{Seed learning rate sensitivity analysis. Cifar-10 on Resnet-18 trained for 150 epochs with \lrschedule{}. We vary the seed learning rate and explore epochs to get the best test accuracy for the particular setting. We report averages over 3 runs.}
\label{tab:seed_senstivity_cifar}

\begin{tabular}{ccc}
  \toprule
  Seed LR        &  Test Accuracy & Optimal Explore Epochs   \\
%      &  &  \\ \hline
\midrule
  0.03   & 95.07 & 120\\ 
  0.05   & 95.12 & 120 \\ 
  0.0625    & 95.15 & 120\\ 
  0.075   & 95.34 & 100 \\ 
  0.0875   & 95.22 & 100\\ 
  0.1 & 95.14 & 100 \\
  0.115   & 95.20 & 60\\ 
  0.125   & 95.06 & 60 \\ 
  0.15   & 95.04 & 30 \\ 
 \bottomrule
\end{tabular}

\end{table}

\section{Conclusions} % and Future work}
\label{sec:conclusions}

%An \textit{explore} at high learning rate allows the optimizer to jump over narrow regions and finally get stuck in a wide region. Since the density of wide regions is much lower than that of narrow regions, we need to explore for a long enough duration to land in a wide region with high probability. 

In this paper, we make an observation that an initial \textit{explore} phase with a high learning rate is essential for good generalization of DNNs. Further, we find that a minimum \textit{explore} duration is required even if the training loss stops improving much earlier. We explain this observation via our hypothesis that in the DNN loss landscape, the density of wide minima is significantly lower than that of narrow minima. 
Motivated by this hypothesis, we present an \textit{Explore-Exploit} based learning rate schedule, called the \lrschedule{}. We do extensive evaluation of \lrschedule{} on multiple models and datasets. In all experiments, the \lrschedule{} outperforms prior hand-tuned baselines, including achieving SOTA test accuracies, when trained with the original training budget, and achieves the same test accuracy as the baseline when trained with a much shorter budget.

%Although we have done multiple experiments for validating this hypothesis, an exciting area of further study will be to explore this aspect theoretically. We are also interested in developing techniques to automatically ascertain that the optimizer has landed in a wide region, and switch to exploit part of the \lrschedule{}. This would save precious training cycles, unlike now where we have to choose a high explore duration to ensure landing in wide region with a high probability. Another area of future work would be to automate the \textit{exploit} part. Although simple linear decay works well, we believe that this can be improved further via an automated technique, especially given that our wide-minima density hypothesis seems to hold at multiple scales.
\section{Acknowledgement}
\label{sec:acknowledgement}

We would like to thank Sanjith Athlur for his help in setting up the VM cluster for large training runs and Harshay Shah for helpful discussions on minima width computation.

\bibliography{arxiv}

\begin{thebibliography}{44}
\providecommand{\natexlab}[1]{#1}
\providecommand{\url}[1]{\texttt{#1}}
\expandafter\ifx\csname urlstyle\endcsname\relax
  \providecommand{\doi}[1]{doi: #1}\else
  \providecommand{\doi}{doi: \begingroup \urlstyle{rm}\Url}\fi

\bibitem[Arora et~al.(2018)Arora, Ge, Neyshabur, and Zhang]{arora2018stronger}
Sanjeev Arora, Rong Ge, Behnam Neyshabur, and Yi~Zhang.
\newblock Stronger generalization bounds for deep nets via a compression
  approach.
\newblock \emph{arXiv preprint arXiv:1802.05296}, 2018.

\bibitem[Baldassi et~al.(2019)Baldassi, Pittorino, and
  Zecchina]{shapinglandscape2019baldassi}
Carlo Baldassi, Fabrizio Pittorino, and Riccardo Zecchina.
\newblock Shaping the learning landscape in neural networks around wide flat
  minima.
\newblock \emph{CoRR}, abs/1905.07833, 2019.
\newblock URL \url{http://arxiv.org/abs/1905.07833}.

\bibitem[Baldassi et~al.(2020)Baldassi, Pittorino, and
  Zecchina]{baldassi2020shaping}
Carlo Baldassi, Fabrizio Pittorino, and Riccardo Zecchina.
\newblock Shaping the learning landscape in neural networks around wide flat
  minima.
\newblock \emph{Proceedings of the National Academy of Sciences}, 117\penalty0
  (1):\penalty0 161--170, 2020.

\bibitem[Bojar et~al.(2014)Bojar, Buck, Federmann, Haddow, Koehn, Leveling,
  Monz, Pecina, Post, Saint-Amand, Soricut, Specia, and
  Tamchyna]{wmt14translation}
Ondrej Bojar, Christian Buck, Christian Federmann, Barry Haddow, Philipp Koehn,
  Johannes Leveling, Christof Monz, Pavel Pecina, Matt Post, Herve Saint-Amand,
  Radu Soricut, Lucia Specia, and Ale~{s} Tamchyna.
\newblock Findings of the 2014 workshop on statistical machine translation.
\newblock In \emph{Proceedings of the Ninth Workshop on Statistical Machine
  Translation}, pages 12--58, Baltimore, Maryland, USA, June 2014. Association
  for Computational Linguistics.
\newblock URL \url{http://www.aclweb.org/anthology/W/W14/W14-3302}.

\bibitem[Byrd et~al.(2003)Byrd, Lu, Nocedal, and Zhu]{byrd-lbgfs}
Richardh Byrd, Peihuang Lu, Jorge Nocedal, and Ciyou Zhu.
\newblock A limited memory algorithm for bound constrained optimization.
\newblock \emph{SIAM Journal on Scientific Computing}, 16, 02 2003.
\newblock \doi{10.1137/0916069}.

\bibitem[Cettolo et~al.(2014)Cettolo, Niehues, St{\"u}ker, Bentivogli, and
  Federico]{iwslt_dataset_cettolo2014}
Mauro Cettolo, Jan Niehues, Sebastian St{\"u}ker, Luisa Bentivogli, and
  Marcello Federico.
\newblock Report on the 11th iwslt evaluation campaign, iwslt 2014.
\newblock In \emph{Proceedings of the International Workshop on Spoken Language
  Translation, Hanoi, Vietnam}, page~57, 2014.

\bibitem[Chaudhari et~al.(2019)Chaudhari, Choromanska, Soatto, LeCun, Baldassi,
  Borgs, Chayes, Sagun, and Zecchina]{chaudhari2019entropy}
Pratik Chaudhari, Anna Choromanska, Stefano Soatto, Yann LeCun, Carlo Baldassi,
  Christian Borgs, Jennifer Chayes, Levent Sagun, and Riccardo Zecchina.
\newblock Entropy-sgd: Biasing gradient descent into wide valleys.
\newblock \emph{Journal of Statistical Mechanics: Theory and Experiment},
  2019\penalty0 (12):\penalty0 124018, 2019.

\bibitem[Devlin et~al.(2018)Devlin, Chang, Lee, and Toutanova]{devlin2018bert}
Jacob Devlin, Ming-Wei Chang, Kenton Lee, and Kristina Toutanova.
\newblock Bert: Pre-training of deep bidirectional transformers for language
  understanding.
\newblock \emph{arXiv preprint arXiv:1810.04805}, 2018.

\bibitem[Dinh et~al.(2017)Dinh, Pascanu, Bengio, and Bengio]{dinh2017sharp}
Laurent Dinh, Razvan Pascanu, Samy Bengio, and Yoshua Bengio.
\newblock Sharp minima can generalize for deep nets.
\newblock In \emph{Proceedings of the 34th International Conference on Machine
  Learning-Volume 70}, pages 1019--1028. JMLR. org, 2017.

\bibitem[Draxler et~al.(2018)Draxler, Veschgini, Salmhofer, and
  Hamprecht]{draxler2018essentially}
Felix Draxler, Kambis Veschgini, Manfred Salmhofer, and Fred Hamprecht.
\newblock Essentially no barriers in neural network energy landscape.
\newblock In \emph{International conference on machine learning}, pages
  1309--1318. PMLR, 2018.

\bibitem[Freeman and Bruna(2016)]{freeman2016topology}
C~Daniel Freeman and Joan Bruna.
\newblock Topology and geometry of half-rectified network optimization.
\newblock \emph{arXiv preprint arXiv:1611.01540}, 2016.

\bibitem[Garipov et~al.(2018)Garipov, Izmailov, Podoprikhin, Vetrov, and
  Wilson]{garipov2018loss}
Timur Garipov, Pavel Izmailov, Dmitrii Podoprikhin, Dmitry Vetrov, and
  Andrew~Gordon Wilson.
\newblock Loss surfaces, mode connectivity, and fast ensembling of dnns.
\newblock \emph{arXiv preprint arXiv:1802.10026}, 2018.

\bibitem[Golatkar et~al.(2019)Golatkar, Achille, and Soatto]{golatkar2019time}
Aditya Golatkar, Alessandro Achille, and Stefano Soatto.
\newblock Time matters in regularizing deep networks: Weight decay and data
  augmentation affect early learning dynamics, matter little near convergence.
\newblock \emph{arXiv preprint arXiv:1905.13277}, 2019.

\bibitem[Goyal et~al.(2017)Goyal, Doll{\'a}r, Girshick, Noordhuis, Wesolowski,
  Kyrola, Tulloch, Jia, and He]{goyal-imagenet-in-an-hour-2017}
Priya Goyal, Piotr Doll{\'a}r, Ross Girshick, Pieter Noordhuis, Lukasz
  Wesolowski, Aapo Kyrola, Andrew Tulloch, Yangqing Jia, and Kaiming He.
\newblock Accurate, large minibatch sgd: Training imagenet in 1 hour.
\newblock \emph{arXiv preprint arXiv:1706.02677}, 2017.

\bibitem[Guiroy et~al.(2019)Guiroy, Verma, and Pal]{guiroy2019towards}
Simon Guiroy, Vikas Verma, and Christopher Pal.
\newblock Towards understanding generalization in gradient-based meta-learning.
\newblock \emph{arXiv preprint arXiv:1907.07287}, 2019.

\bibitem[He et~al.(2016)He, Zhang, Ren, and Sun]{resnet_he_2016}
Kaiming He, Xiangyu Zhang, Shaoqing Ren, and Jian Sun.
\newblock Deep residual learning for image recognition.
\newblock In \emph{Proceedings of the IEEE conference on computer vision and
  pattern recognition}, pages 770--778, 2016.

\bibitem[Hochreiter and Schmidhuber(1997)]{hochreiter1997flat}
Sepp Hochreiter and J{\"u}rgen Schmidhuber.
\newblock Flat minima.
\newblock \emph{Neural Computation}, 9\penalty0 (1):\penalty0 1--42, 1997.

\bibitem[Jastrzebski et~al.(2017)Jastrzebski, Kenton, Arpit, Ballas, Fischer,
  Bengio, and Storkey]{jastrzkebski2017three}
Stanis{\l}aw Jastrzebski, Zachary Kenton, Devansh Arpit, Nicolas Ballas, Asja
  Fischer, Yoshua Bengio, and Amos Storkey.
\newblock Three factors influencing minima in sgd.
\newblock \emph{arXiv preprint arXiv:1711.04623}, 2017.

\bibitem[Jastrzebski et~al.(2019)Jastrzebski, Kenton, Ballas, Fischer, Bengio,
  and Storkey]{jastrzebski_iclr_2019}
Stanis{\l}aw Jastrzebski, Zachary Kenton, Nicolas Ballas, Asja Fischer, Yoshua
  Bengio, and Amost Storkey.
\newblock On the relation between the sharpest directions of {DNN} loss and the
  {SGD} step length.
\newblock In \emph{International Conference on Learning Representations}, 2019.
\newblock URL \url{https://openreview.net/forum?id=SkgEaj05t7}.

\bibitem[Kawaguchi(2016)]{kawaguchi2016deep}
Kenji Kawaguchi.
\newblock Deep learning without poor local minima.
\newblock In \emph{Advances in neural information processing systems}, pages
  586--594, 2016.

\bibitem[Keskar et~al.(2016)Keskar, Mudigere, Nocedal, Smelyanskiy, and
  Tang]{keskar2016large}
Nitish~Shirish Keskar, Dheevatsa Mudigere, Jorge Nocedal, Mikhail Smelyanskiy,
  and Ping Tak~Peter Tang.
\newblock On large-batch training for deep learning: Generalization gap and
  sharp minima.
\newblock \emph{arXiv preprint arXiv:1609.04836}, 2016.

\bibitem[Kingma and Ba(2014{\natexlab{a}})]{adam_kingma2014adam}
Diederik~P Kingma and Jimmy Ba.
\newblock Adam: A method for stochastic optimization.
\newblock \emph{arXiv preprint arXiv:1412.6980}, 2014{\natexlab{a}}.

\bibitem[Kingma and Ba(2014{\natexlab{b}})]{kingma2014adam}
Diederik~P Kingma and Jimmy Ba.
\newblock Adam: A method for stochastic optimization.
\newblock \emph{arXiv preprint arXiv:1412.6980}, 2014{\natexlab{b}}.

\bibitem[Krizhevsky et~al.(2009)Krizhevsky, Hinton, et~al.]{cifar-10-dataset}
Alex Krizhevsky, Geoffrey Hinton, et~al.
\newblock Learning multiple layers of features from tiny images.
\newblock Technical report, Citeseer, 2009.

\bibitem[Kunstner et~al.(2019)Kunstner, Balles, and
  Hennig]{empiricalfisher2019limitations}
Frederik Kunstner, Lukas Balles, and Philipp Hennig.
\newblock Limitations of the empirical fisher approximation.
\newblock \emph{arXiv preprint arXiv:1905.12558}, 2019.

\bibitem[Li et~al.(2019)Li, Wei, and Ma]{li2019towards}
Yuanzhi Li, Colin Wei, and Tengyu Ma.
\newblock Towards explaining the regularization effect of initial large
  learning rate in training neural networks.
\newblock \emph{arXiv preprint arXiv:1907.04595}, 2019.

\bibitem[Liu et~al.(2019)Liu, Jiang, He, Chen, Liu, Gao, and
  Han]{liu2019variance_radam}
Liyuan Liu, Haoming Jiang, Pengcheng He, Weizhu Chen, Xiaodong Liu, Jianfeng
  Gao, and Jiawei Han.
\newblock On the variance of the adaptive learning rate and beyond.
\newblock \emph{arXiv preprint arXiv:1908.03265}, 2019.

\bibitem[Loshchilov and Hutter(2016)]{loshchilov2016sgdr}
Ilya Loshchilov and Frank Hutter.
\newblock Sgdr: Stochastic gradient descent with warm restarts.
\newblock \emph{arXiv preprint arXiv:1608.03983}, 2016.

\bibitem[McCandlish et~al.(2018)McCandlish, Kaplan, Amodei, and
  Team]{large-batch-training-openai-2018}
Sam McCandlish, Jared Kaplan, Dario Amodei, and OpenAI~Dota Team.
\newblock An empirical model of large-batch training.
\newblock \emph{arXiv preprint arXiv:1812.06162}, 2018.

\bibitem[Ott et~al.(2019)Ott, Edunov, Baevski, Fan, Gross, Ng, Grangier, and
  Auli]{ott2019fairseq}
Myle Ott, Sergey Edunov, Alexei Baevski, Angela Fan, Sam Gross, Nathan Ng,
  David Grangier, and Michael Auli.
\newblock fairseq: A fast, extensible toolkit for sequence modeling.
\newblock In \emph{Proceedings of NAACL-HLT 2019: Demonstrations}, 2019.

\bibitem[Rajpurkar et~al.(2016)Rajpurkar, Zhang, Lopyrev, and
  Liang]{rajpurkar2016squad}
Pranav Rajpurkar, Jian Zhang, Konstantin Lopyrev, and Percy Liang.
\newblock Squad: 100,000+ questions for machine comprehension of text.
\newblock \emph{arXiv preprint arXiv:1606.05250}, 2016.

\bibitem[Russakovsky et~al.(2015)Russakovsky, Deng, Su, Krause, Satheesh, Ma,
  Huang, Karpathy, Khosla, Bernstein, et~al.]{imagenet-dataset}
Olga Russakovsky, Jia Deng, Hao Su, Jonathan Krause, Sanjeev Satheesh, Sean Ma,
  Zhiheng Huang, Andrej Karpathy, Aditya Khosla, Michael Bernstein, et~al.
\newblock Imagenet large scale visual recognition challenge.
\newblock \emph{International journal of computer vision}, 115\penalty0
  (3):\penalty0 211--252, 2015.

\bibitem[Sagun et~al.(2017)Sagun, Evci, G{\"u}ney, Dauphin, and
  Bottou]{sagun2017empirical}
Levent Sagun, Utku Evci, V~Ugur G{\"u}ney, Yann Dauphin, and L{\'e}on Bottou.
\newblock Empirical analysis of the hessian of over-parametrized neural
  networks. iclr 2018 workshop contribution.
\newblock \emph{arXiv preprint arXiv:1706.04454}, 2017.

\bibitem[Shallue et~al.(2018)Shallue, Lee, Antognini, Sohl-Dickstein, Frostig,
  and Dahl]{large-batch-training-google-2018}
Christopher~J Shallue, Jaehoon Lee, Joe Antognini, Jascha Sohl-Dickstein, Roy
  Frostig, and George~E Dahl.
\newblock Measuring the effects of data parallelism on neural network training.
\newblock \emph{arXiv preprint arXiv:1811.03600}, 2018.

\bibitem[Shen et~al.(2020)Shen, Zheng, Shen, Qu, and Chen]{shen2020simple}
Dinghan Shen, Mingzhi Zheng, Yelong Shen, Yanru Qu, and Weizhu Chen.
\newblock A simple but tough-to-beat data augmentation approach for natural
  language understanding and generation.
\newblock \emph{arXiv preprint arXiv:2009.13818}, 2020.

\bibitem[Smith(2017)]{smith2017cyclical}
Leslie~N Smith.
\newblock Cyclical learning rates for training neural networks.
\newblock In \emph{2017 IEEE Winter Conference on Applications of Computer
  Vision (WACV)}, pages 464--472. IEEE, 2017.

\bibitem[Smith(2018)]{smith2018disciplined_onecycle}
Leslie~N Smith.
\newblock A disciplined approach to neural network hyper-parameters: Part
  1--learning rate, batch size, momentum, and weight decay.
\newblock \emph{arXiv preprint arXiv:1803.09820}, 2018.

\bibitem[Sokol and Park(2018)]{fim2018information}
Piotr~A Sokol and Il~Memming Park.
\newblock Information geometry of orthogonal initializations and training.
\newblock \emph{arXiv preprint arXiv:1810.03785}, 2018.

\bibitem[Vaswani et~al.(2017)Vaswani, Shazeer, Parmar, Uszkoreit, Jones, Gomez,
  Kaiser, and Polosukhin]{vaswani2017attention}
Ashish Vaswani, Noam Shazeer, Niki Parmar, Jakob Uszkoreit, Llion Jones,
  Aidan~N Gomez, {\L}ukasz Kaiser, and Illia Polosukhin.
\newblock Attention is all you need.
\newblock In \emph{Advances in neural information processing systems}, pages
  5998--6008, 2017.

\bibitem[Wang et~al.(2018)Wang, Keskar, Xiong, and Socher]{wang2018identifying}
Huan Wang, Nitish~Shirish Keskar, Caiming Xiong, and Richard Socher.
\newblock Identifying generalization properties in neural networks.
\newblock \emph{arXiv preprint arXiv:1809.07402}, 2018.

\bibitem[Wu et~al.(2018)Wu, Ma, and Weinan]{wu2018sgd}
Lei Wu, Chao Ma, and E~Weinan.
\newblock How sgd selects the global minima in over-parameterized learning: A
  dynamical stability perspective.
\newblock In \emph{Advances in Neural Information Processing Systems}, pages
  8279--8288, 2018.

\bibitem[Xie et~al.(2020)Xie, Sato, and Sugiyama]{xie2020diffusion}
Zeke Xie, Issei Sato, and Masashi Sugiyama.
\newblock A diffusion theory for deep learning dynamics: Stochastic gradient
  descent exponentially favors flat minima.
\newblock \emph{arXiv e-prints}, pages arXiv--2002, 2020.

\bibitem[Yoshida and Miyato(2017)]{yoshida2017spectral}
Yuichi Yoshida and Takeru Miyato.
\newblock Spectral norm regularization for improving the generalizability of
  deep learning.
\newblock \emph{arXiv preprint arXiv:1705.10941}, 2017.

\bibitem[You et~al.(2019)You, Li, Reddi, Hseu, Kumar, Bhojanapalli, Song,
  Demmel, Keutzer, and Hsieh]{bert76lamb}
Yang You, Jing Li, Sashank Reddi, Jonathan Hseu, Sanjiv Kumar, Srinadh
  Bhojanapalli, Xiaodan Song, James Demmel, Kurt Keutzer, and Cho-Jui Hsieh.
\newblock Large batch optimization for deep learning: Training bert in 76
  minutes.
\newblock In \emph{International Conference on Learning Representations}, 2019.

\end{thebibliography}

\clearpage

\appendix

\clearpage
\section{Comparisons with Other Baseline Learning Rate Schedules}
\label{sec:extra_baselines}

In this section we compare \lrschedule{} against several other learning rate schedules -- one-cycle, linear decay and cosine decay.

\begin{figure}[th]
    \centering
    \begin{subfigure}[t]{0.49\textwidth}
        \centering
        \includegraphics[width=0.99\linewidth]{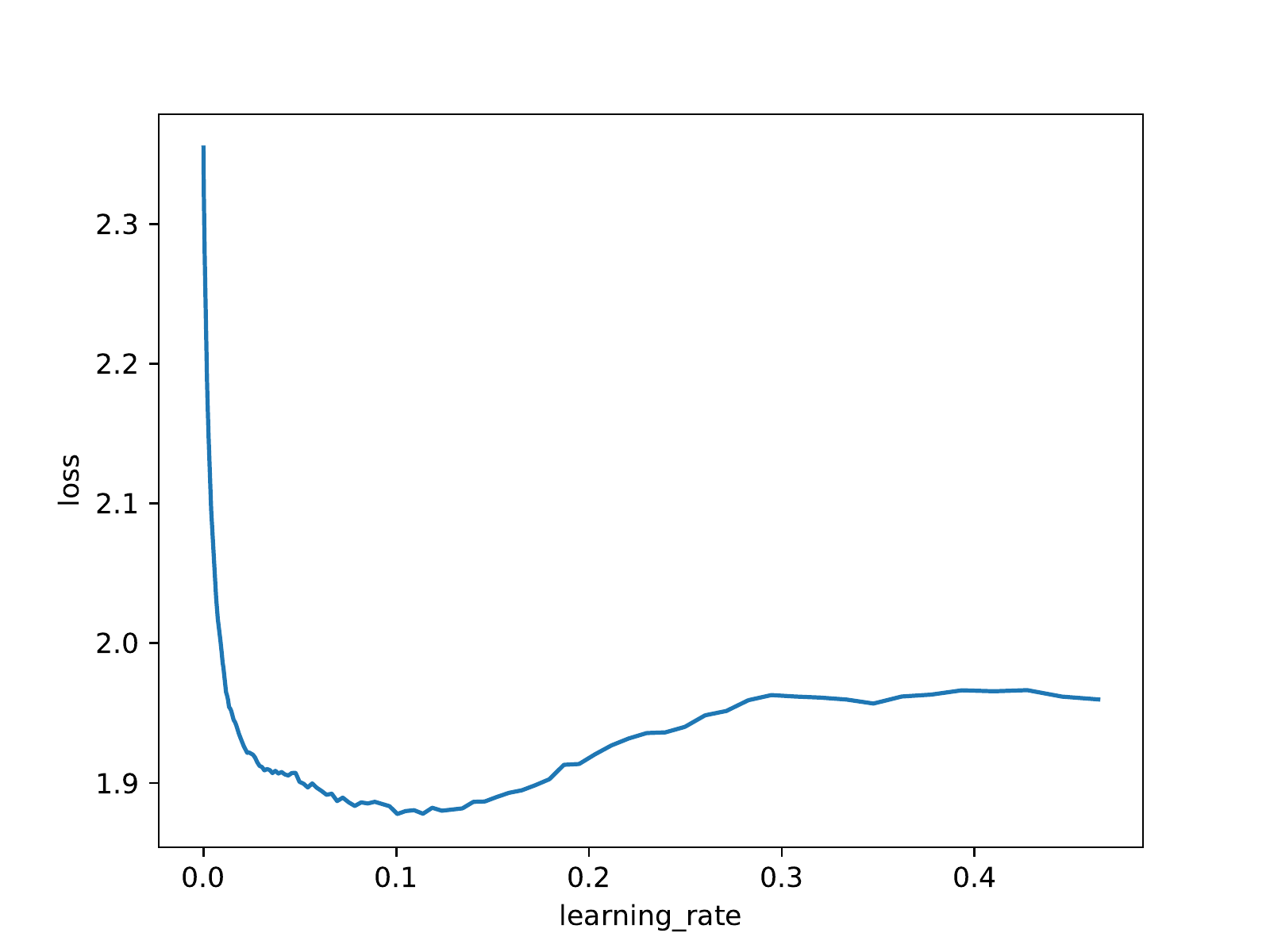}
        \caption{LR range test for CIFAR-10}
        \label{fig:lr_range_test_cifar}
    \end{subfigure} \hfill
    \begin{subfigure}[t]{0.49\textwidth}
        \centering
        \includegraphics[width=0.99\linewidth]{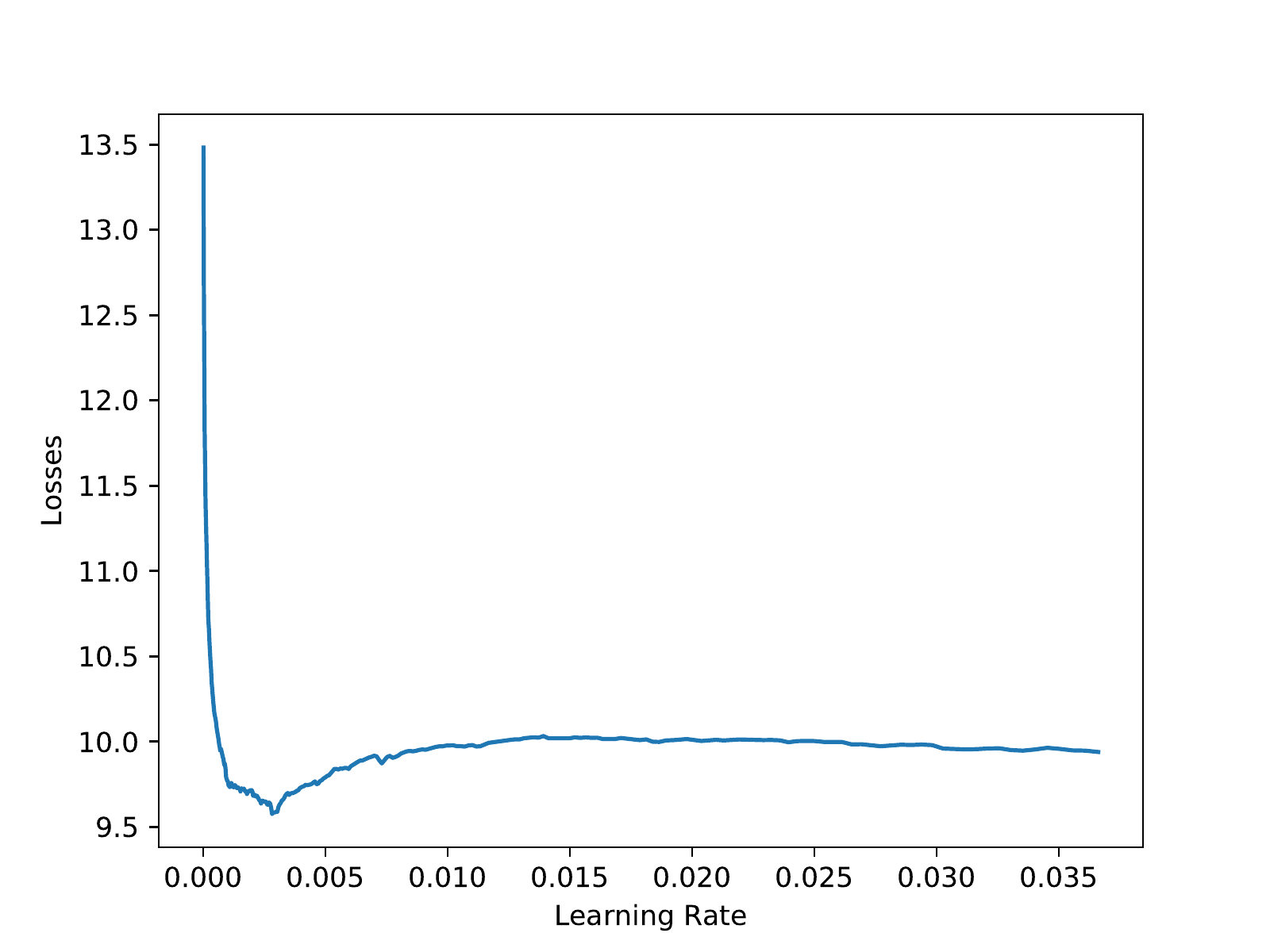}
        \caption{LR range test for IWSLT'14 DE-EN}
        \label{fig:lr_range_test_iwslt}
    \end{subfigure}
    \begin{subfigure}[t]{0.49\textwidth}
        \centering
        \includegraphics[width=0.99\linewidth]{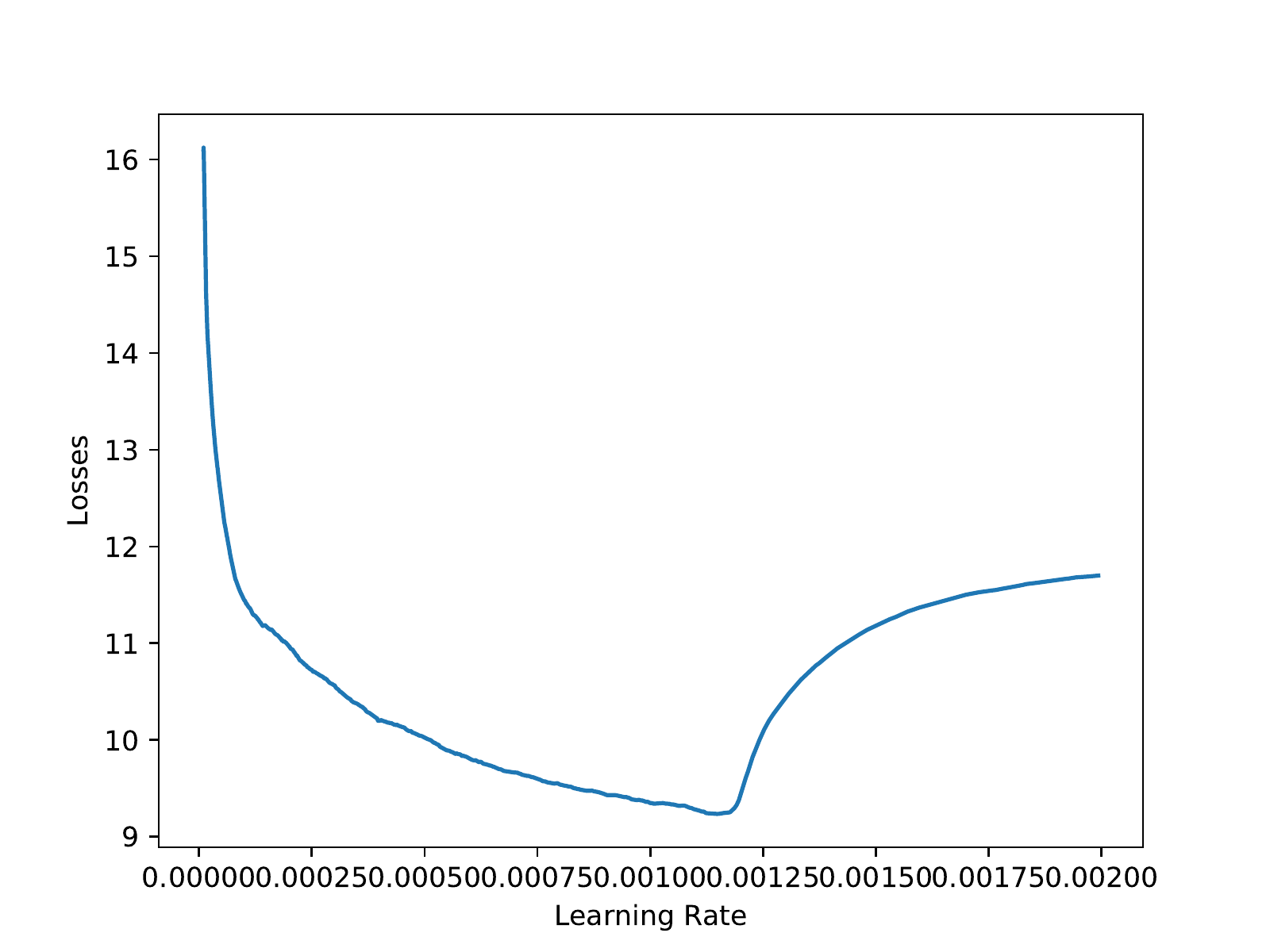}
        \caption{LR range test for WMT'14 EN-DE}
        \label{fig:lr_range_test_WMT}
    \end{subfigure} \hfill
    \begin{subfigure}[t]{0.49\textwidth}
        \centering
        \includegraphics[width=0.99\linewidth]{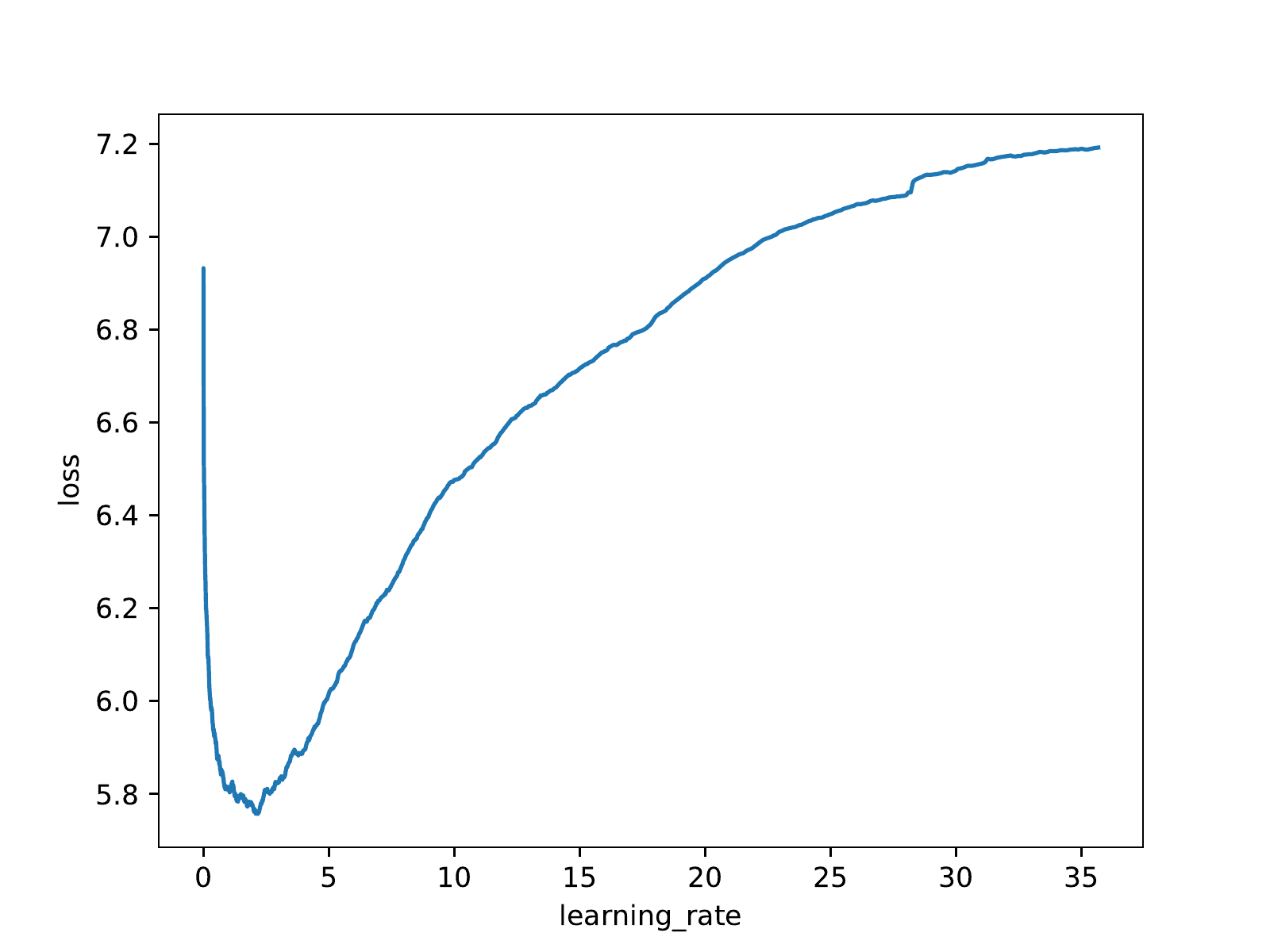}
        \caption{LR range test for ImageNet}
        \label{fig:lr_range_test_imagenet}
    \end{subfigure}
\caption{learning rate range test for selecting the maximum learning rate. A good choice is the learning rate is a bit before the minima in a region where the loss is still decreasing.}
\label{fig:lr_range_tests}
\end{figure}

\textbf{One-Cycle}: The one-cycle learning rate schedule was proposed in \cite{smith2018disciplined_onecycle} (also see \cite{smith2017cyclical}). This schedule first chooses a maximum learning rate based on an learning rate range test. The learning rate range test starts from a small learning rate and keeps increasing the learning rate until the loss starts exploding (see figure~\ref{fig:lr_range_tests}). \cite{smith2018disciplined_onecycle} suggests that the maximum learning rate should be chosen to be bit before the minima, in a region where the loss is still decreasing. There is some subjectivity in making this choice, although some blogs and libraries\footnote{See e.g. \url{https://towardsdatascience.com/finding-good-learning-rate-and-the-one-cycle-policy-7159fe1db5d6} and \url{https://sgugger.github.io/how-do-you-find-a-good-learning-rate.html}. Also see \url{https://docs.fast.ai/callbacks.lr_finder.html} and \url{https://docs.fast.ai/callbacks.one_cycle.html}} suggest using a learning rate one order lower than the one at minima. We go with this choice for all our runs.

Once the maximum learning rate is chosen, the one-cycle schedule proceeds as follows. The learning rate starts at a specified fraction\footnote{See div\_factor in \url{https://docs.fast.ai/callbacks.one_cycle.html}. We chose the fraction to be 0.1 in our experiments.} of the maximum learning rate and is increased linearly to the maximum learning rate for 45 percent of the training budget and then decreased linearly for the remaining 45. For the final 10 percent, the learning rate is reduced by a large factor (we chose a factor of 10).  We used an opensource implementation \footnote{\url{https://github.com/nachiket273/One_Cycle_Policy} } for our experiments.

\textbf{Linear Decay}: The linear decay learning rate schedule simply decays the learning rate linearly to zero starting from a seed learning rate.

\textbf{Cosine Decay}: The cosine decay learning rate schedule decays the learning rate to zero following a cosine curve, starting from a seed learning rate.

\subsection{Cifar-10}
Figure~\ref{fig:lr_range_test_cifar}  shows the learning rate range test for Cifar-10 with the Resnet-18 network. The minima occurs around learning rate of 0.09, and we choose $9e^{-3}$ as the maximum learning rate for the One-Cycle runs. For linear, cosine decay schedules we start with a seed learning rate of 0.1 as used in the standard baselines. The training loss and test accuracy for the various schedules are shown in Table~\ref{tab:cifar_results_extra_baselines_full_budget} for the full budget runs (200 epochs), and in Table~\ref{tab:cifar_results_extra_baselines_short_budget} for the short budget runs (150 epochs).

\begin{table}[h!]
\small
\centering
\caption{Cifar-10 on Resnet-18 full budget training (200 epochs): Training loss and Test accuracy for more learning rate schedules. We report the mean and standard deviation over 7 runs.}
\label{tab:cifar_results_extra_baselines_full_budget}

\begin{tabular}{ccc}
  \toprule
  LR Schedule     & Test Accuracy  & Train Loss \\ 
  \midrule
  One-Cycle       & 94.08 (0.07)  & 0.0041 (6e-5) \\
  Cosine Decay    & 95.23 (0.11) & 0.0023 (9e-5)  \\
  Linear Decay    & 95.18 (0.15) & 0.0018 (7e-5)  \\
  \lrschedule{}   & \textbf{95.26} (0.11) & 0.0023 (1e-4)  \\ 
\bottomrule
\end{tabular}

\end{table}

\begin{table}[h!]
\small
\centering
\caption{Cifar-10 on Resnet-18 short budget training (150 epochs): Training loss and Test accuracy for more learning rate schedules. We report the mean and standard deviation over 7 runs.}
\label{tab:cifar_results_extra_baselines_short_budget}

\begin{tabular}{ccc}
  \toprule
  LR Schedule   & Test Accuracy   & Train Loss  \\ 
  \midrule
  One-Cycle     & 93.84 (0.082)    & 0.0052 (7e-5)   \\
  Cosine Decay   & 95.06 (0.16)    & 0.0030 (2e-4)\\
  Linear Decay    & 95.02 (0.10)  & 0.0021 (1e-4) \\
  \lrschedule{}   & \textbf{95.14} (0.18) & 0.0044 (3e-4)  \\ 
  \bottomrule
\end{tabular}

\end{table}

\subsection{ImageNet}
Figure~\ref{fig:lr_range_test_imagenet} shows the learning rate range test for ImageNet with the Resnet-50 network. The minima occurs around learning rate of 2.16, and we choose $0.216$ as the maximum learning rate for One-Cycle runs. For linear, cosine decay schedules we start with a seed learning rate of 0.1 as used in the standard baselines. The training loss and test accuracy for the various schedules are shown in Table~\ref{tab:ImageNet_full_budget_runs} for the full budget runs (90 epochs), and in Table~\ref{tab:ImageNet_short_budget_runs} for the short budget runs (50 epochs).

\begin{table}[h]
\small
\centering
\caption{ImageNet with ResNet-50 full budget training (90 epochs): Training loss, Test Top-1 and Test Top-5 for more learning rate schedules. We report the mean and standard deviation over 3 runs.}
\label{tab:ImageNet_full_budget_runs}
\begin{tabular}{cccc}
  \toprule
  LR Schedule    & Test Top-1  & Test Top-5  & Train Loss (av) \\ 
  \midrule
  One Cycle        & 75.39 (0.137) & 92.56 (0.040) & 0.96 (0.003) \\
  Cosine Decay      & 76.41 (0.212) & 93.28 (0.066) & 0.80 (0.002) \\
  Linear decay     &  76.54 (0.155) & 93.21 (0.051)  & 0.75 (0.001)\\
  \lrschedule{}     & \textbf{76.71} (0.097)  & \textbf{93.32} (0.031) & 0.79 (0.001) \\ \bottomrule
\end{tabular}

\end{table}
\begin{table}[h!]
\small
\centering
\caption{ImageNet with ResNet-50 short budget training (50 epochs): Training loss, Test Top-1 and Test Top-5 for more learning rate schedules. We report the mean and standard deviation over 3 runs.}
\label{tab:ImageNet_short_budget_runs}
\begin{tabular}{cccc}
  \toprule
  LR Schedule    & Test Top-1  & Test Top-5  & Train Loss (av) \\ 
  \midrule
  One Cycle        & 75.36 (0.096) & 92.53 (0.079) & 1.033 (0.004)\\
  Cosine Decay     & 75.71 (0.116) & 92.81 (0.033) & 0.96 (0.002) \\
  Linear decay     & 75.82 (0.080) & 92.84 (0.036) & 0.91 (0.002) \\
  \lrschedule{}    & \textbf{75.92} (0.11) & \textbf{92.90} (0.085) & 0.90 (0.003) \\
%  \lrschedule{} (30 explore)     & 75.77 (0.202)  & \textbf{92.89} (0.049) & 0.93 (9e-4) \\ 
  \bottomrule
\end{tabular}

\end{table}

\subsection{WMT'14 EN-DE}
Figure~\ref{fig:lr_range_test_WMT} shows the learning rate range test for WMT'14 EN-DE on the transformer networks. The minima occurs near $1.25e^{-3}$. For the maximum learning rate, we choose $2.5e^{-4}$ for the default one-cycle policy. For linear, cosine decay schedules we start with a seed learning rate of $3e^{-4}$ as used in the standard baselines The training, validation perplexity and BLEU scores for the various schedules are shown in Table~\ref{tab:wmt_results_extra_baselines_full_budget} for the full budget runs (70 epochs), and in Table~\ref{tab:wmt_results_extra_baselines_short_budget} for the short budget runs (30 epochs).

\begin{table}[h]
\small
\centering
\caption{WMT'14 (EN-DE) on Transformer networks full budget training (70 epochs): Training, validation perplexity and test BLEU scores for more learning rate schedules. The test BLEU scores are computed on the checkpoint with the best validation perplexity. We report the mean and standard deviation over 3 runs.}
\label{tab:wmt_results_extra_baselines_full_budget}
\begin{tabular}{cccc}
  \toprule
  LR Schedule   & Test BLEU Score  & Train ppl & Validation ppl   \\ 
  \midrule 
  One-Cycle     &  27.19 (0.081)    & 3.96 (0.014)  & 4.95 (0.013)     \\
  Cosine Decay  &  27.35 (0.09)     & 3.87 (0.011)   & 4.91 (0.008)   \\
  Linear Decay  &  27.29 (0.06)     & 3.87 (0.017)   & 4.89 (0.02)   \\
  \lrschedule{} & \textbf{27.53} (0.12)      & 3.89 (0.017)   & 4.87 (0.006)    \\ 
  \bottomrule
\end{tabular}

\end{table}

\begin{table}[!th]
\small
\centering
\caption{WMT'14 (EN-DE) on Transformer networks short budget training (30 epochs): Training, validation perplexity and test BLEU scores for more learning rate schedules. The test BLEU scores are computed on the checkpoint with the best validation perplexity. We report the mean and standard deviation over 3 runs.}
\label{tab:wmt_results_extra_baselines_short_budget}
\begin{tabular}{cccc}
  \toprule
  LR Schedule    & Test BLEU Score  & Train ppl & Validation ppl  \\ 
  \midrule
  One-Cycle      &  26.80 (0.2) & 4.38 (0.017)  & 5.02 (0.007)    \\
  Cosine Decay  &  26.95 (0.23) & 4.32 (0.013)  & 4.99 (0.011)   \\
  Linear Decay  & 26.77  (0.12) & 4.36 (0.092)  & 5.02 (0.01)  \\
  \lrschedule{} & \textbf{27.28} (0.17)  & 4.31 (0.02)   & 4.92 (0.007)    \\ 
  \bottomrule
\end{tabular}

\end{table}

\vspace{2in}
\subsection{IWSLT'14 DE-EN}
Figure~\ref{fig:lr_range_test_iwslt} shows the learning rate range test for IWSLT on the transformer networks. The minima occurs near $2.5e^{-3}$. For the maximum learning rate, we choose $2.5e^{-4}$ for the default one-cycle policy. For linear, cosine decay schedules we start with a seed learning rate of $3e^{-4}$ as used in the standard baselines The training, validation perplexity and BLEU scores for the various schedules are shown in Table~\ref{tab:iwslt_results_extra_baselines_full_budget} for the full budget runs (50 epochs), and in Table~\ref{tab:iwslt_results_extra_baselines_short_budget} for the short budget runs (35 epochs).

\begin{table}[h]
\small
\centering
\caption{IWSLT'14 (DE-EN) on Transformer networks full budget training (50 epochs): Training, validation perplexity and test BLEU scores for more learning rate schedules. The test BLEU scores are computed on the checkpoint with the best validation perplexity. We report the mean and standard deviation over 3 runs.}
\label{tab:iwslt_results_extra_baselines_full_budget}
\begin{tabular}{cccc}
  \toprule
  LR Schedule   & Test BLEU Score  & Train ppl & Validation ppl  \\ 
  \midrule
  One-Cycle     &  34.77 (0.064) & 3.68 (0.009)  & 4.97 (0.010)   \\
  Cosine Decay  &  35.21 (0.063) & 3.08 (0.004)   & 4.88 (0.014)  \\
  Linear Decay  &  34.97 (0.035) & 3.36 (0.001)   & 4.92 (0.035)   \\
  \lrschedule{} &  \textbf{35.53} (0.06)  & 3.00 (0.044)   & 4.86 (0.02)    \\ 
  \bottomrule
\end{tabular}

\end{table}

\begin{table}[h]
\small
\centering
\caption{IWSLT'14 (DE-EN) on Transformer networks short budget training (35 epochs): Training, validation perplexity and test BLEU scores for more learning rate schedules. The test BLEU scores are computed on the checkpoint with the best validation perplexity. We report the mean and standard deviation over 3 runs.}
\label{tab:iwslt_results_extra_baselines_short_budget}
\begin{tabular}{cccc}
  \toprule
  LR Schedule   & Test BLEU Score & Train ppl & Validation ppl    \\
  \midrule
  One-Cycle     &  34.43 (0.26) & 3.98 (0.028)  & 5.09 (0.017)     \\
  Cosine Decay  &  34.46 (0.33) & 3.86 (0.131)   & 5.06 (0.106)   \\
  Linear Decay  & 34.16  (0.28) & 4.11 (0.092)   & 5.14 (0.066)   \\
  \lrschedule{} & \textbf{35.08} (0.12) & 3.58 (0.063)   & 4.90 (0.049)    \\ 
  \bottomrule
\end{tabular}

\end{table}

\subsection{SQuAD-v1.1 finetuning with BERT\textsubscript{BASE}}

We choose $1e^{-5}$ as the maximum learning rate for One-Cycle runs as the minima occurs close to $1e^{-4}$ . For linear, cosine decays we start with a seed learning rate of $3e^{-5}$ as used in standard baselines.  Table~\ref{tab:squad_results_extra_baselines} show the average training loss, average test EM and F1 scores for the various schedules. We did not do a short budget training for this dataset, as the full budget is just 2 epochs.

\begin{table}[h!]
\small
\centering
\caption{SQuAD-v1.1 fine-tuning on BERT\textsubscript{BASE} for more learning rate schedules. We report the average training loss, average test EM, F1 scores over 3 runs.}
\label{tab:squad_results_extra_baselines}
\begin{tabular}{cccc}
  \toprule
  LR Schedule    & EM (av)  & F1 (av) & Train Loss (av)  \\ 
  \midrule
  One Cycle        & 79.9 (0.17) & 87.8 (0.091) & 1.062 (0.003) \\
  Cosine Decay      & 81.31 (0.07) & 88.61 (0.040) & 0.999 (0.003) \\
  Linear decay      & 80.89 (0.15) & 88.38 (0.042) & 1.0003 (0.004)\\
  \lrschedule{}     & \textbf{81.38} (0.02)  & \textbf{88.66} (0.045) & 1.003 (0.002)\\
  \bottomrule
\end{tabular}
\end{table}

%We also ran all our experiments with multiple other learning rate schedules -- one-cycle \citep{smith2018disciplined_onecycle}, cosine decay~\citep{loshchilov2016sgdr} and linear decay. See section~\ref{sec:extra_baselines} in supplementary material for details of these learning rate schedules, and a detailed performance comparison.
%For one-cycle, the maximum learning rate was chosen to be 1/10th of the learning rate found via the learning rate range test of \cite{smith2018disciplined_onecycle,smith2017cyclical}. %\footnote{See  \url{https://docs.fast.ai/callbacks.lr_finder.html} and \url{https://docs.fast.ai/callbacks.one_cycle.html}}.

%the IWSLT Bleu score of 35.08 (knee) vs 34.46 (cosine). We note that the ‘small’ deltas in accuracy take non-trivial amount of extra training to be achieved -- Cosine schedule takes ~30\% more training time than Knee to reach 35.01. 

%Also note that in some experiments of the full budget runs, the accuracy achieved by cosine decay and linear decay schedules is similar to \lrschedule{}; however, this is not observed in the reduced budget runs. This observation is consistent with our hypothesis as the cosine and linear decay schedules have an implicit explore portion in the initial part of the run (where the learning rate stays high enough). The duration of this ``implicit'' explore reduces as we reduce the training budget, hurting test accuracy. The \lrschedule{} on the other hand prioritizes the essential explore phase even in the reduced budget runs, thus, achieving higher test accuracies.

\clearpage
\section{Detailed Plots}
\label{sec:detailed_plots}
%%%%% ImageNet  %%%%%

\begin{figure}[h]
    \centering
    \begin{subfigure}[t]{0.88\textwidth}
        \centering
        \includegraphics[width=0.99\linewidth]{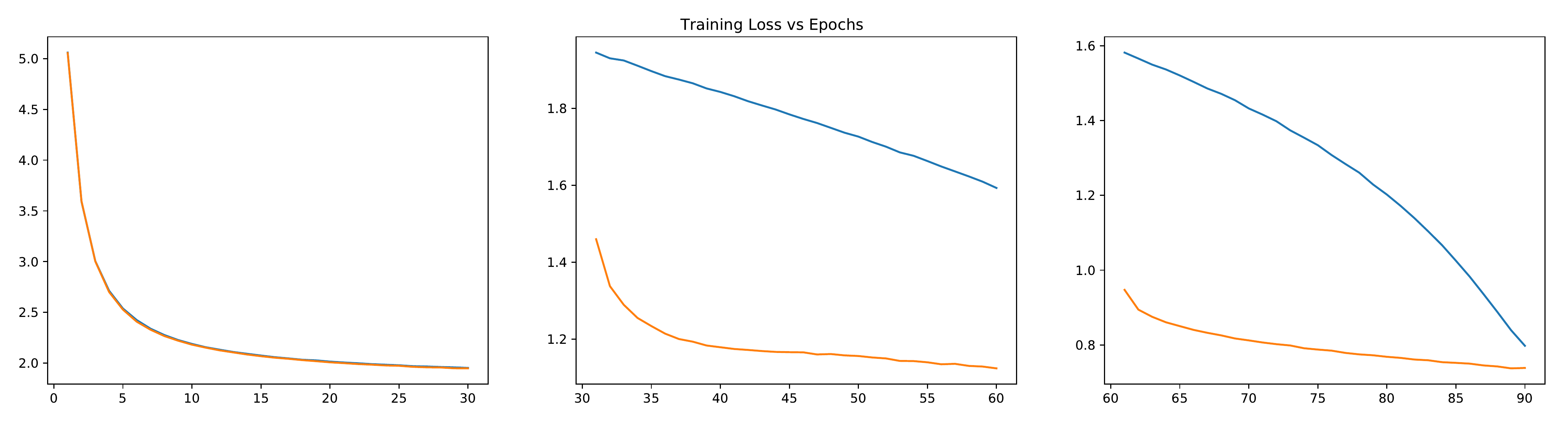}
        %\caption{Training Loss}
        \label{fig:imagenet_momentum_tr_loss}
    \end{subfigure}
    \begin{subfigure}[t]{0.88\textwidth}
        \centering
        \includegraphics[width=0.99\linewidth]{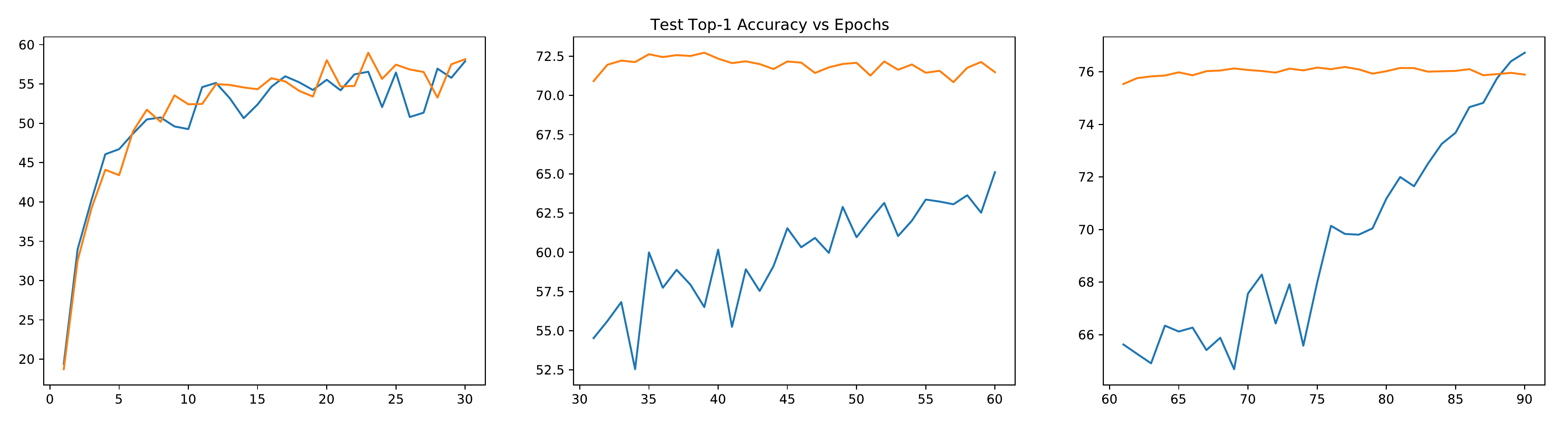}
        %\caption{Test Accuracy}
        \label{fig:imagenet_momentum_test_top1_acc}
    \end{subfigure}
    \begin{subfigure}[t]{0.88\textwidth}
        \centering
        \includegraphics[width=0.99\linewidth]{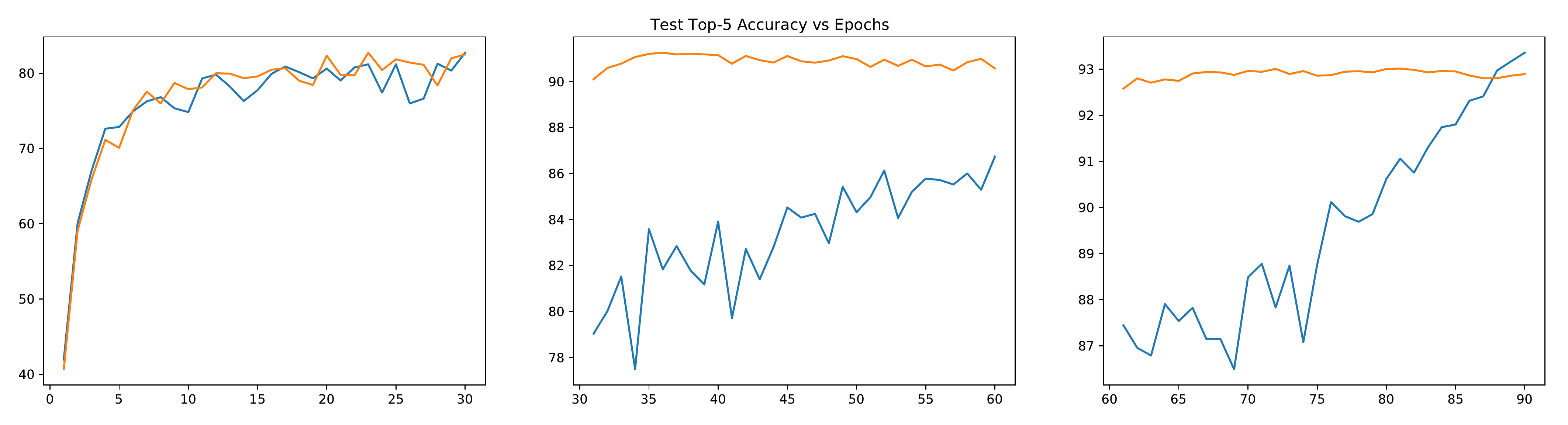}
        %\caption{Test Accuracy}
        \label{fig:imagenet_momentum_test_top5_acc}
    \end{subfigure}
    \begin{subfigure}[t]{0.88\textwidth}
        \centering
        \includegraphics[width=0.99\linewidth]{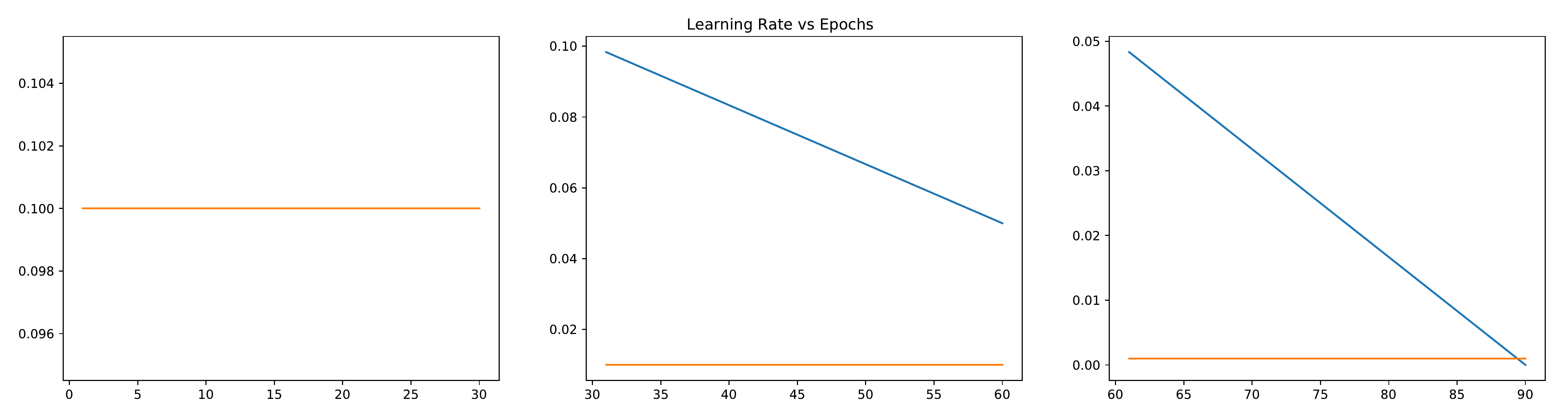}
        %\caption{Learning Rate}
        \label{fig:imagenet_momentum_lr}
    \end{subfigure}
\caption{ImageNet on Resnet-50 trained with Momentum. Shown are the training loss, top-1/top-5 test accuracy and learning rate as a function of epochs, for the baseline scheme (orange) vs the \lrschedule{} scheme (blue). The plot is split into 3 parts to permit higher fidelity in the y-axis range.}
\label{fig:imagenet_momentum_result}
\end{figure}

%%%%% Cifar-10  %%%%%

\begin{figure}[h]
    \begin{subfigure}[t]{\textwidth}
        \centering
        \includegraphics[width=0.99\linewidth]{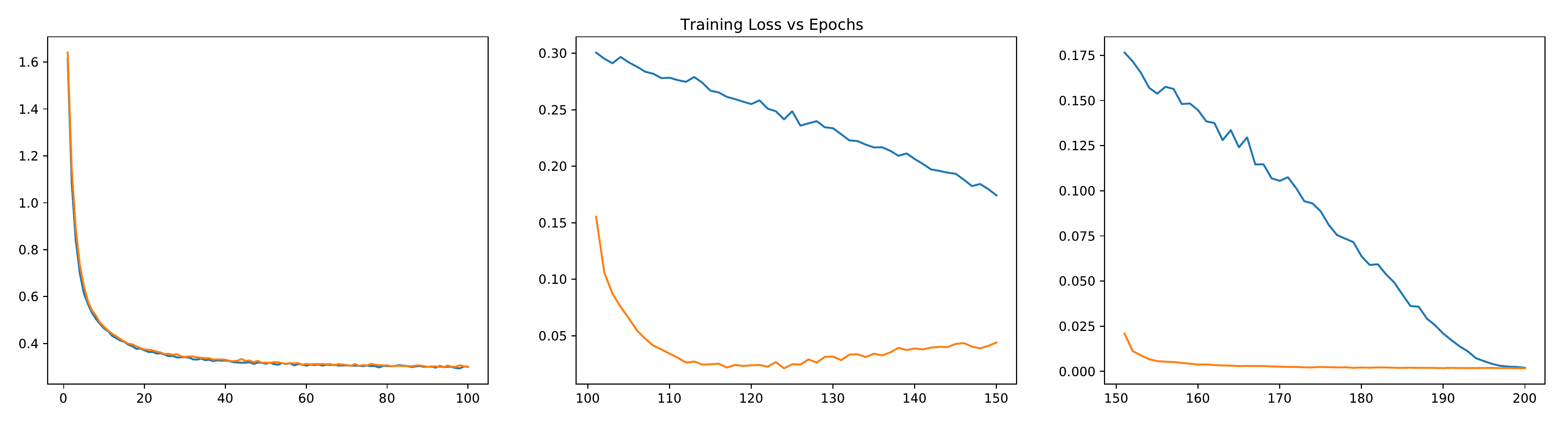}
        %\caption{Training Loss}
        \label{fig:cifar_momentum_tr_loss}
    \end{subfigure}
    \begin{subfigure}[t]{\textwidth}
        \centering
        \includegraphics[width=0.99\linewidth]{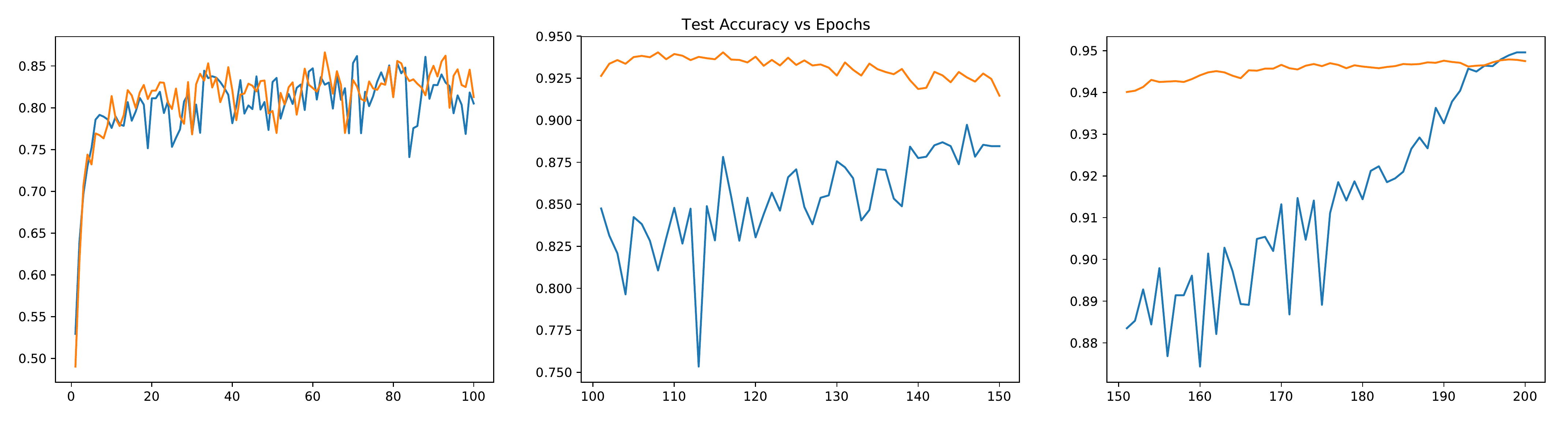}
        %\caption{Test Accuracy}
        \label{fig:cifar_momentum_test_acc}
    \end{subfigure}
    \begin{subfigure}[t]{\textwidth}
        \centering
        \includegraphics[width=0.99\linewidth]{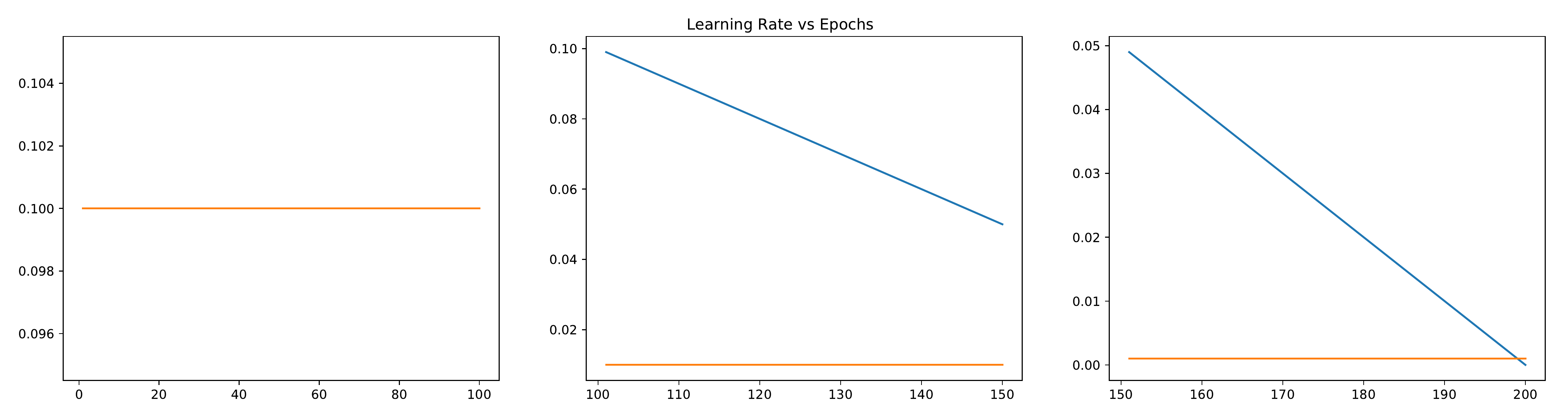}
        %\caption{Learning Rate}
        \label{fig:cifar_momentum_lr}
    \end{subfigure}
\caption{Cifar-10 on Resnet-18 trained with Momentum. Shown are the training loss, test accuracy and learning rate as a function of epochs, for the baseline scheme (orange) vs the \lrschedule{} scheme (blue). The plot is split into 3 parts to permit higher fidelity in the y-axis range.}
\label{fig:cifar_momentum_result}
\end{figure}

%%%%%% BERT pretraining  %%%%%

\begin{figure}[ht]
\begin{minipage}{\textwidth}
  \begin{subfigure}[t]{\textwidth}
        \centering
        \includegraphics[width=0.99\linewidth]{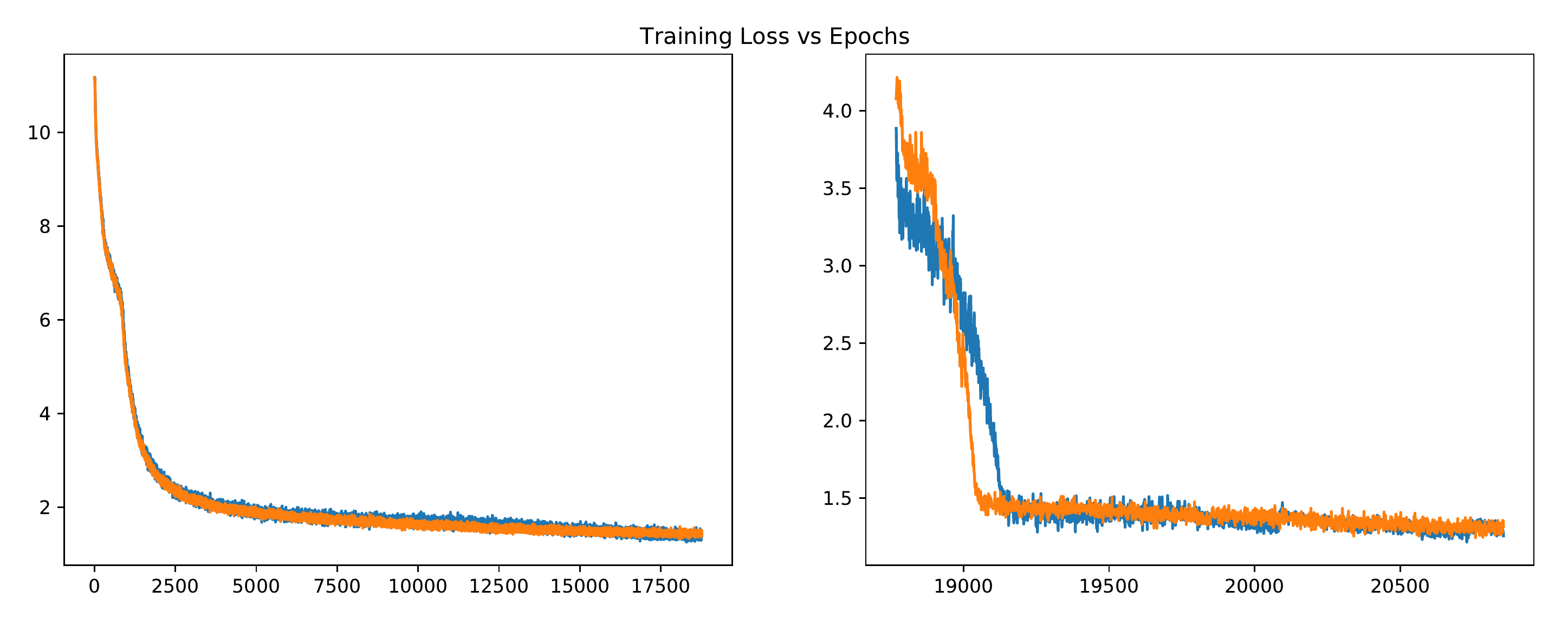}
        %\caption{Training Loss}
        \label{fig:bert_tr_loss}
    \end{subfigure}
    \begin{subfigure}[t]{\textwidth}
        \centering
        \includegraphics[width=0.99\linewidth]{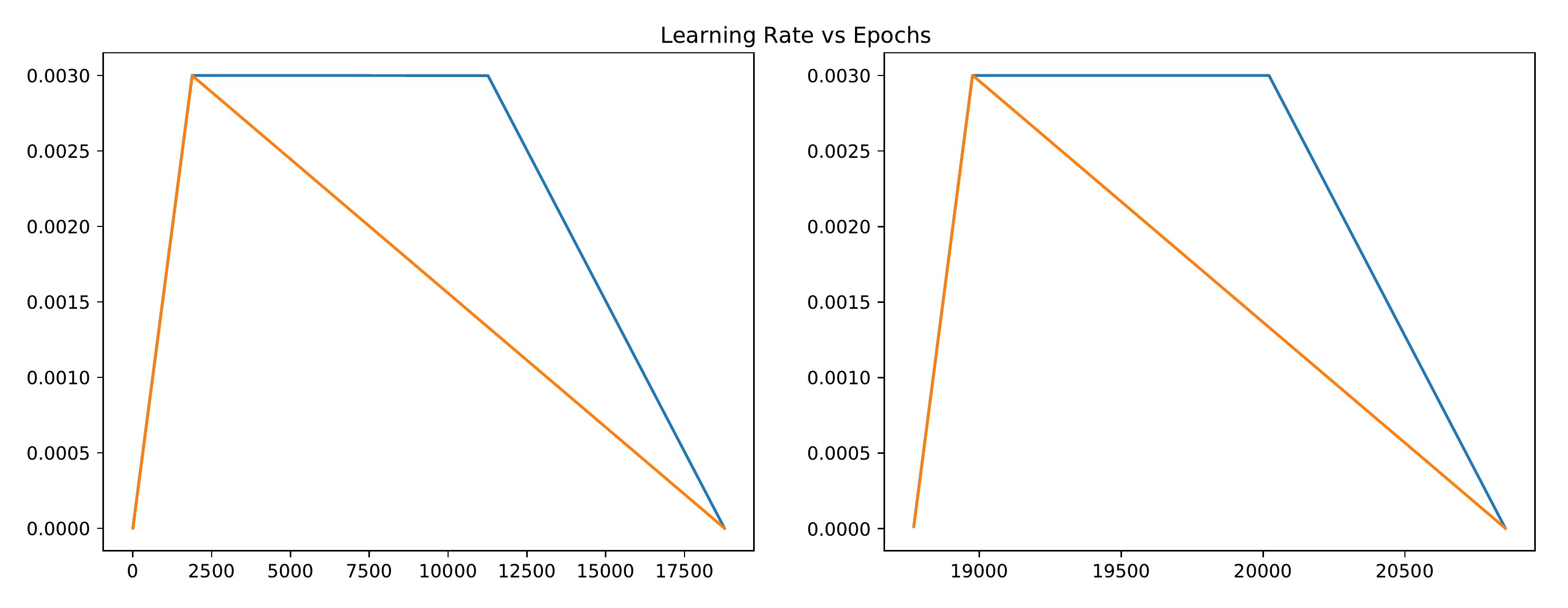}
        %\caption{Test Accuracy}
        \label{fig:bert_lr}
    \end{subfigure}
\caption{BERT\textsubscript{LARGE} pretraining for batch size of 16k with LAMB optimizer for the short budget runs. Shown are the training loss and learning rate as a function of steps, for the baseline scheme short budget (orange) vs the \lrschedule{} scheme short budget (blue). The plot is split into 2 parts to give a clear picture of the two phases of training~\cite{devlin2018bert}. Note that even though the training loss curves look similar for the two runs, we see a significant gap in F1 score obtained when we fine-tune the model checkpoints on SQuAD-v1.1 \cite{rajpurkar2016squad}. See Table~\ref{tab:bert-large_finetuning} for details.
}
\label{fig:bert_plots}
\end{minipage}
%\end{figure}
\begin{minipage}{\textwidth}
\small
\centering
\vspace{25pt}
\begin{tabular}{cccccc}
  \toprule
  LR Schedule & F1 - Trial 1   & F1 - Trial 2 & F1 - Trial 3 & F1 avg. & F1 max \\
  \midrule
    Baseline (short budget)  & 90.39 & 90.64 & 90.53 & 90.52 & 90.64\\
  \lrschedule{}{} ( short budget ) & 91.22  & 91.29 & 91.18 & 91.23 & 91.29\\
  \lrschedule{}{} ( full budget ) & 91.45  & 91.41 & 91.51 & 91.46 & 91.51\\
  
  \bottomrule
\end{tabular}
\captionsetup{type=table}
\caption{SQuAD fine-tuning on BERT\textsubscript{LARGE}. We report F1 scores for 3 different trials as well as the maximum and average values.}
\label{tab:bert-large_finetuning}
\end{minipage}
\end{figure}

%%%%% WMT  %%%%%
\begin{figure}[h]
    \begin{subfigure}[t]{\textwidth}
        \centering
        \includegraphics[width=0.99\linewidth]{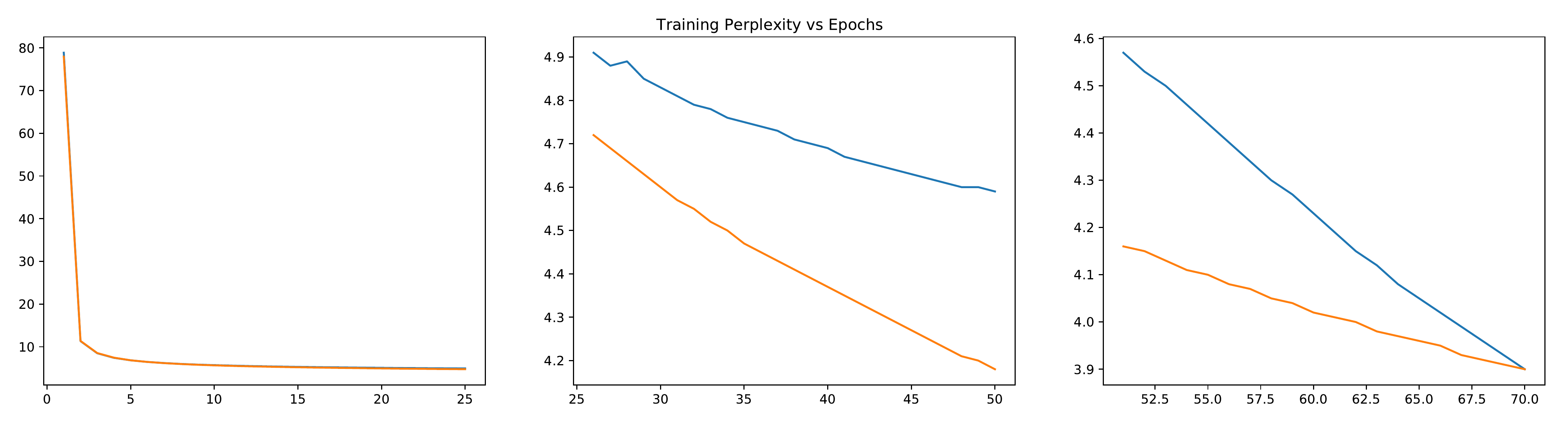}
        %\caption{Training Loss}
        \label{fig:wmt_tr_loss}
    \end{subfigure}
    \begin{subfigure}[t]{\textwidth}
        \centering
        \includegraphics[width=0.99\linewidth]{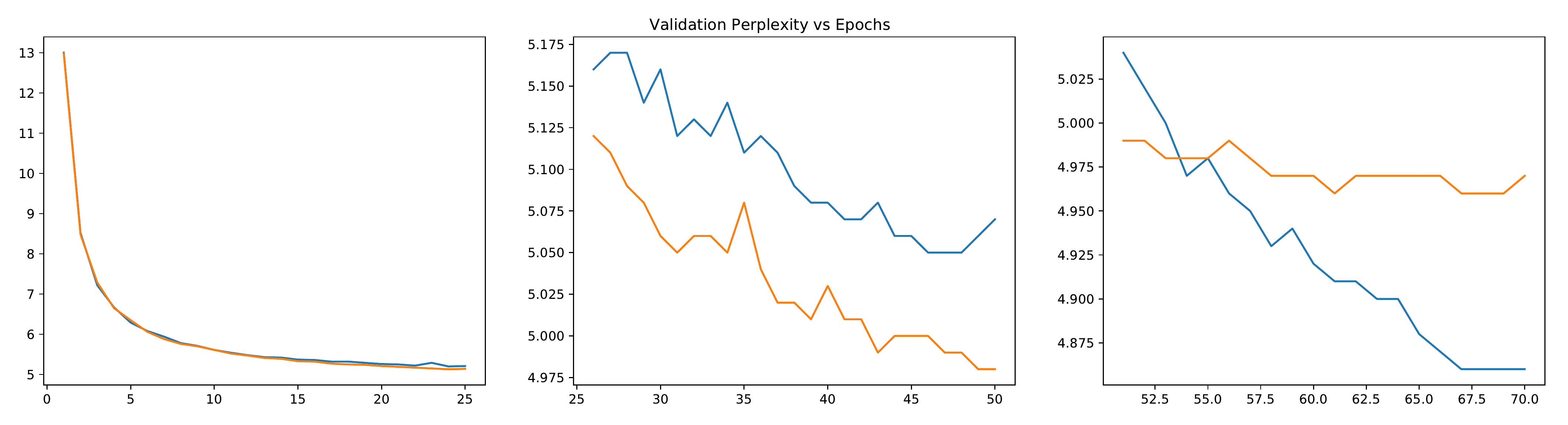}
        %\caption{Test Accuracy}
        \label{fig:wmt_test_acc}
    \end{subfigure}
    \begin{subfigure}[t]{\textwidth}
        \centering
        \includegraphics[width=0.99\linewidth]{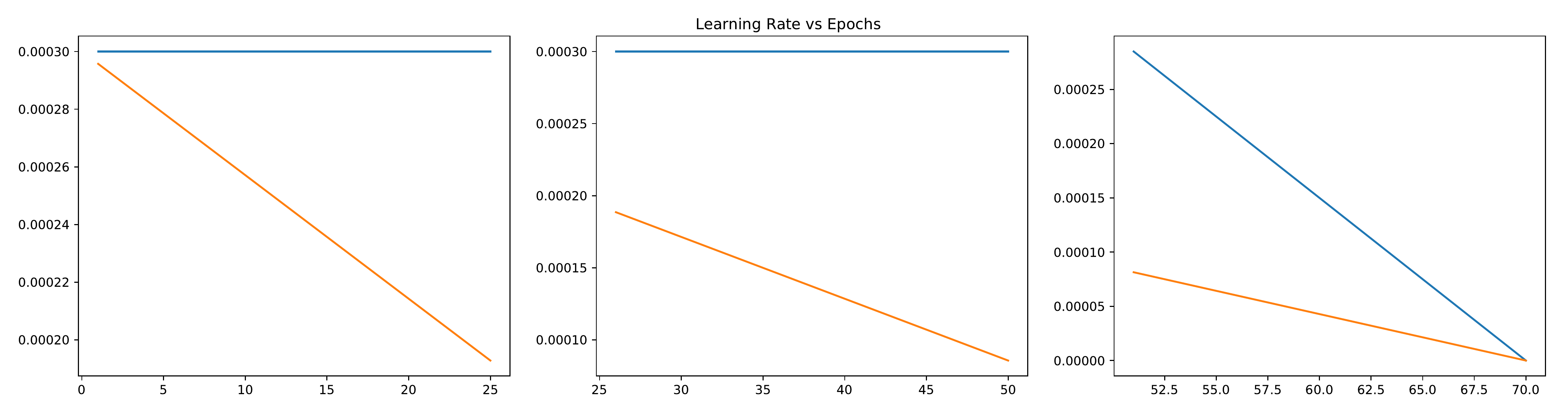}
        %\caption{Learning Rate}
        \label{fig:wmt_lr}
    \end{subfigure}
\caption{WMT'14 (EN-DE) on Transformer\textsubscript{BASE} network trained with RAdam. Shown are the training perplexity, validation perplexity and learning rate as a function of epochs, for the baseline scheme (orange) vs the \lrschedule{} scheme (blue). The plot is split into 3 parts to permit higher fidelity in the y-axis range.}
\label{fig:wmt_radam_lrl}
\end{figure}

%%%%% IWSLT  %%%%%

\begin{figure}[h]
    \begin{subfigure}[t]{\textwidth}
        \centering
        \includegraphics[width=0.99\linewidth]{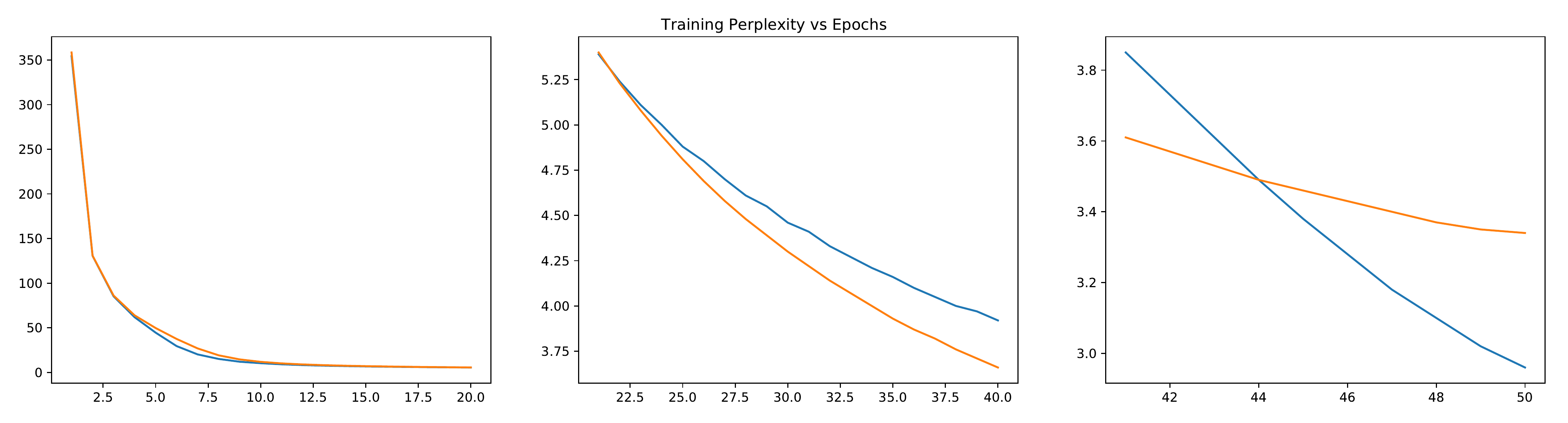}
        %\caption{Training Loss}
        \label{fig:iwslt_adam_tr_loss}
    \end{subfigure}
    \begin{subfigure}[t]{\textwidth}
        \centering
        \includegraphics[width=0.99\linewidth]{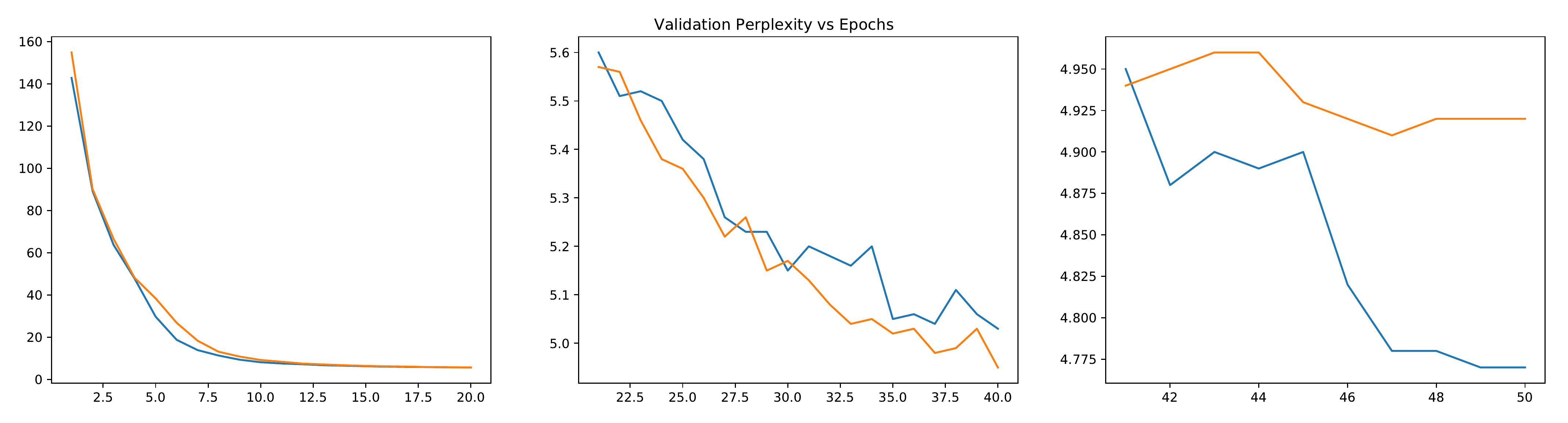}
        %\caption{Test Accuracy}
        \label{fig:iwslt_adam_test_acc}
    \end{subfigure}
    \begin{subfigure}[t]{\textwidth}
        \centering
        \includegraphics[width=0.99\linewidth]{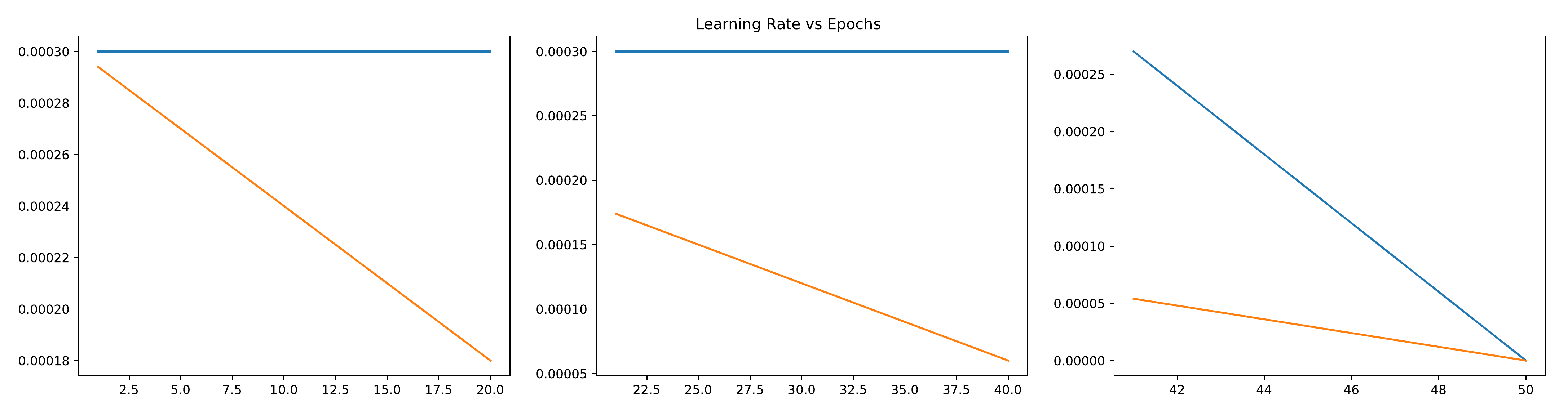}
        %\caption{Learning Rate}
        \label{fig:iwslt_adam_lr}
    \end{subfigure}
\caption{IWSLT'14 (DE-EN) on Transformer\textsubscript{BASE} network trained with RAdam. Shown are the training perplexity, validation perplexity and learning rate as a function of epochs, for the baseline scheme (orange) vs the \lrschedule{} scheme (blue). The plot is split into 3 parts to permit higher fidelity in the y-axis range.}
\label{fig:iwslt_adam_result}
\end{figure}

\begin{figure}[h]
    \begin{subfigure}[t]{\textwidth}
        \centering
        \includegraphics[width=0.99\linewidth]{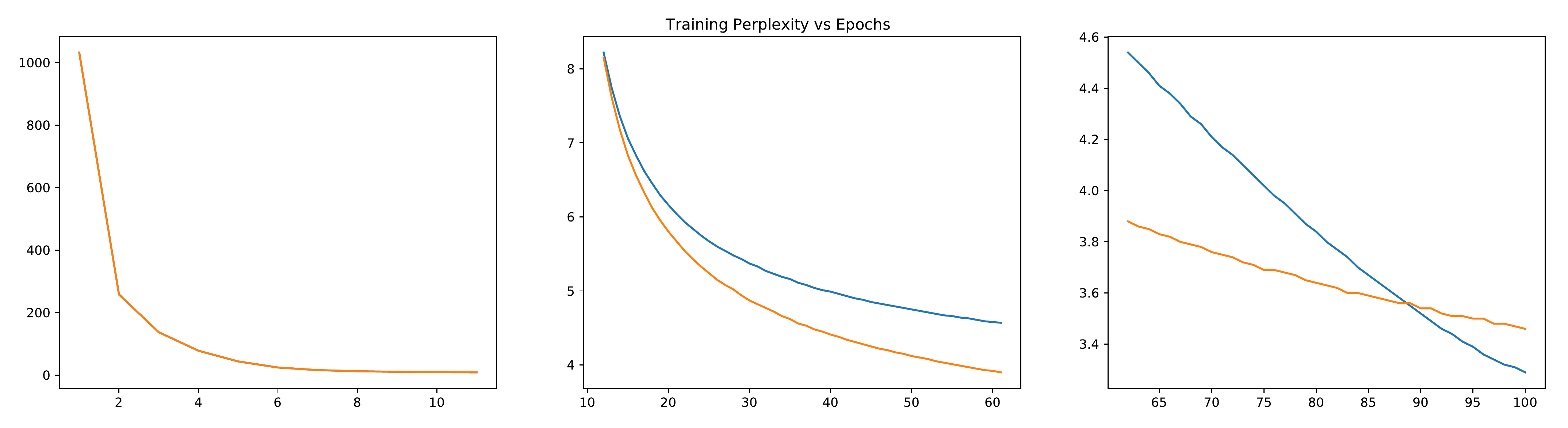}
        %\caption{Training Loss}
        \label{fig:cutoff_adam_tr_ppl}
    \end{subfigure}
    \begin{subfigure}[t]{\textwidth}
        \centering
        \includegraphics[width=0.99\linewidth]{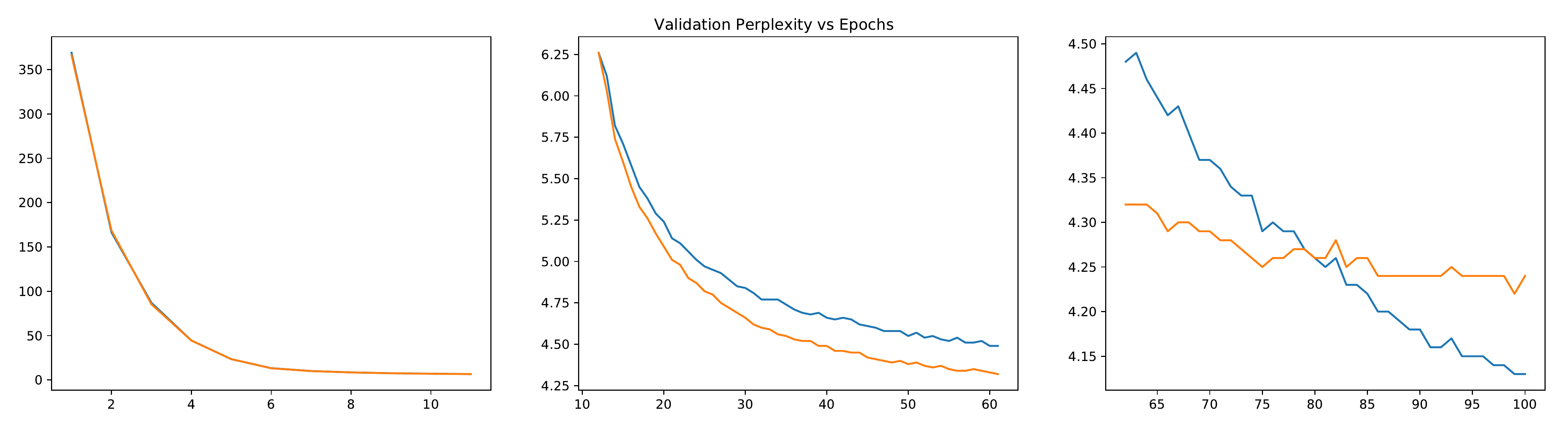}
        %\caption{Training Loss}
        \label{fig:cutoff_adam_val_ppl}
    \end{subfigure}
    \begin{subfigure}[t]{\textwidth}
        \centering
        \includegraphics[width=0.99\linewidth]{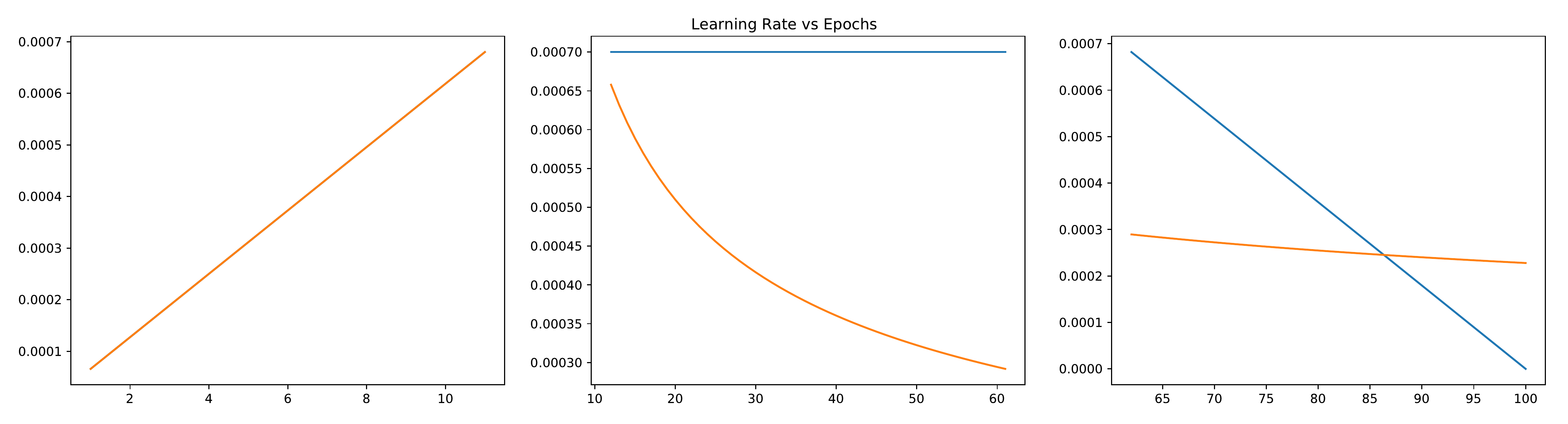}
        %\caption{Learning Rate}
        \label{fig:cutoff_adam_lr}
    \end{subfigure}
\caption{IWSLT'14 (DE-EN) on the SOTA model   Cutoff\citep{shen2020simple}, trained with Adam. Shown are the training perplexity, validation perplexity and learning rate as a function of epochs, for the baseline scheme (orange) vs the \lrschedule{} scheme (blue).}
\label{fig:Cutoff_adam_result}
\end{figure}

\begin{figure}[h]
    \begin{subfigure}[t]{\textwidth}
        \centering
        \includegraphics[width=0.99\linewidth]{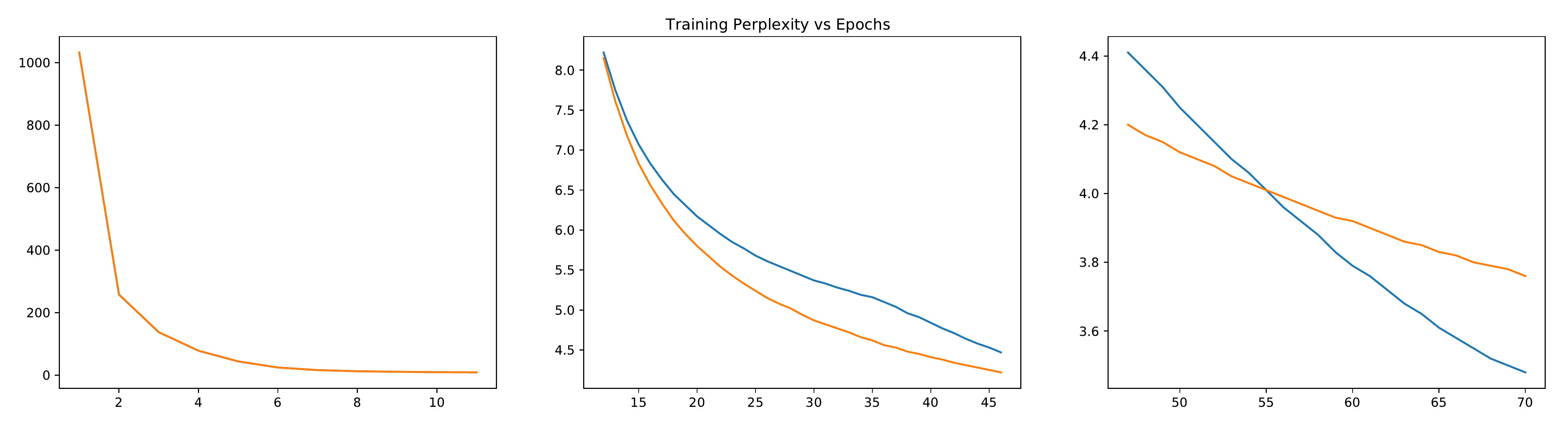}
        %\caption{Training Loss}
        \label{fig:cutoff_short_adam_tr_ppl}
    \end{subfigure}
    \begin{subfigure}[t]{\textwidth}
        \centering
        \includegraphics[width=0.99\linewidth]{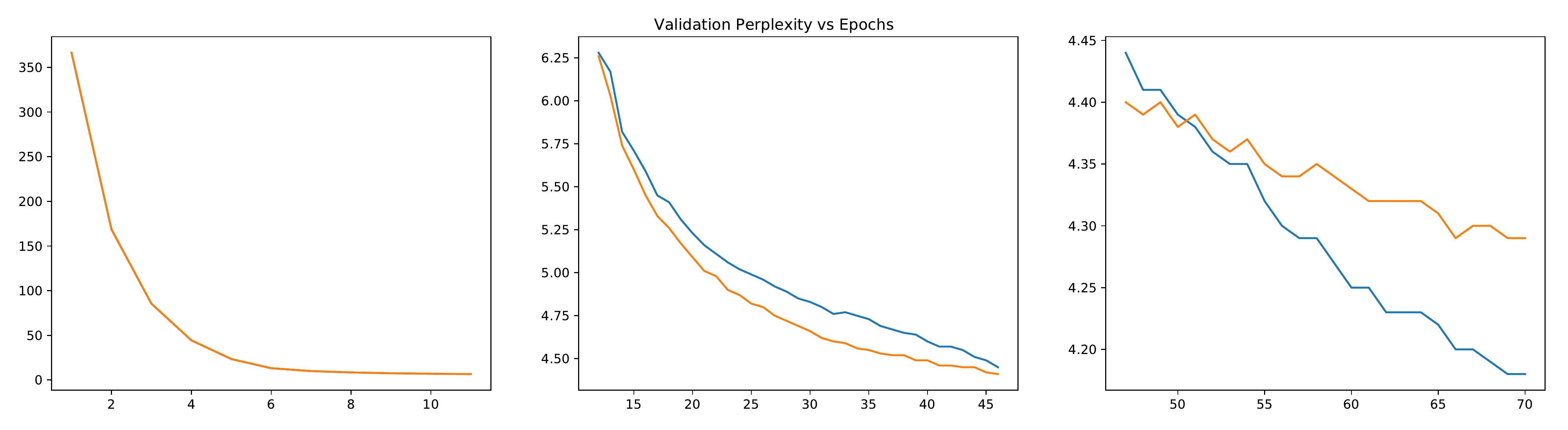}
        %\caption{Training Loss}
        \label{fig:cutoff_short_adam_val_ppl}
    \end{subfigure}
    \begin{subfigure}[t]{\textwidth}
        \centering
        \includegraphics[width=0.99\linewidth]{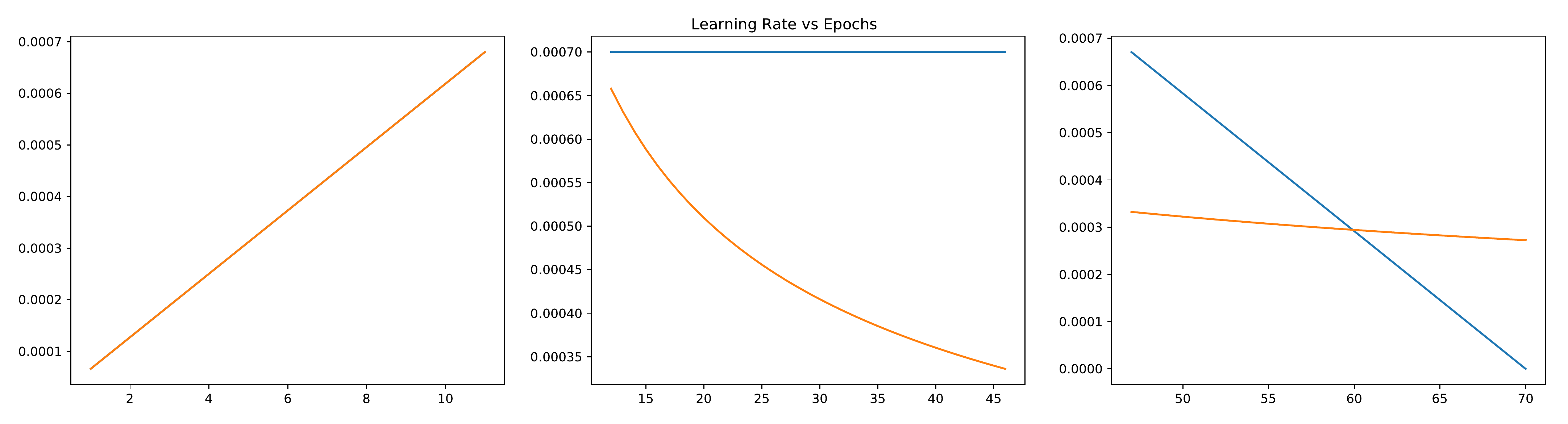}
        %\caption{Learning Rate}
        \label{fig:cutoff_short_adam_lr}
    \end{subfigure}
\caption{IWSLT'14 (DE-EN) on the SOTA model Cutoff\citep{shen2020simple}, trained with Adam with a reduced training budget of 70 epochs. Shown are the training perplexity, validation perplexity and learning rate as a function of epochs, for the baseline scheme (orange) vs the \lrschedule{} scheme (blue).}
\label{fig:Cutoff_short_adam_result}
\end{figure}

\iffalse
%%%%% IWSLT MAT %%%%%

\begin{figure}[h]
    \begin{subfigure}[t]{\textwidth}
        \centering
        \includegraphics[width=0.99\linewidth]{figures/MAT_IWSLT/tr_ppl.pdf}
        %\caption{Training Loss}
        \label{fig:MAT_adam_tr_ppl}
    \end{subfigure}
    \begin{subfigure}[t]{\textwidth}
        \centering
        \includegraphics[width=0.99\linewidth]{figures/MAT_IWSLT/val_ppl.pdf}
        %\caption{Training Loss}
        \label{fig:MAT_adam_val_ppl}
    \end{subfigure}
    \begin{subfigure}[t]{\textwidth}
        \centering
        \includegraphics[width=0.99\linewidth]{figures/MAT_IWSLT/lr.pdf}
        %\caption{Learning Rate}
        \label{fig:MAT_adam_lr}
    \end{subfigure}
\caption{IWSLT'14 (DE-EN) on MAT network trained with Adam. Shown are the training perplexity, validation perplexity and learning rate as a function of epochs, for the baseline scheme (orange) vs the \lrschedule{} scheme (blue). MAT training involves two training phases with 200 epochs each, shown in separate columns above.}
\label{fig:MAT_adam_result}
\end{figure}
\fi
%%%%% Bert finetuning Squad %%%%%

\begin{figure}[h]
    \begin{subfigure}[t]{\textwidth}
        \centering
        \includegraphics[width=0.99\linewidth]{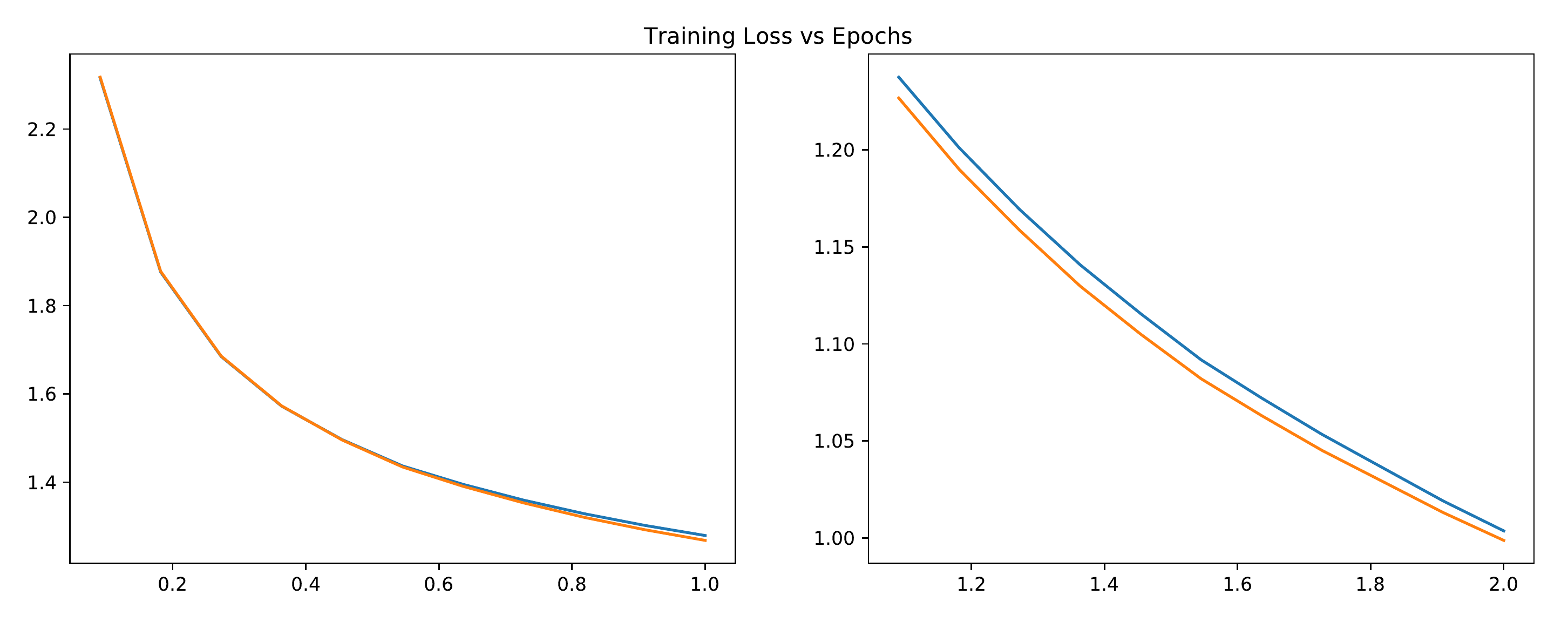}
        %\caption{Training Loss}
        \label{fig:squad_bert_adam_tr_loss}
    \end{subfigure}
    \begin{subfigure}[t]{\textwidth}
        \centering
        \includegraphics[width=0.99\linewidth]{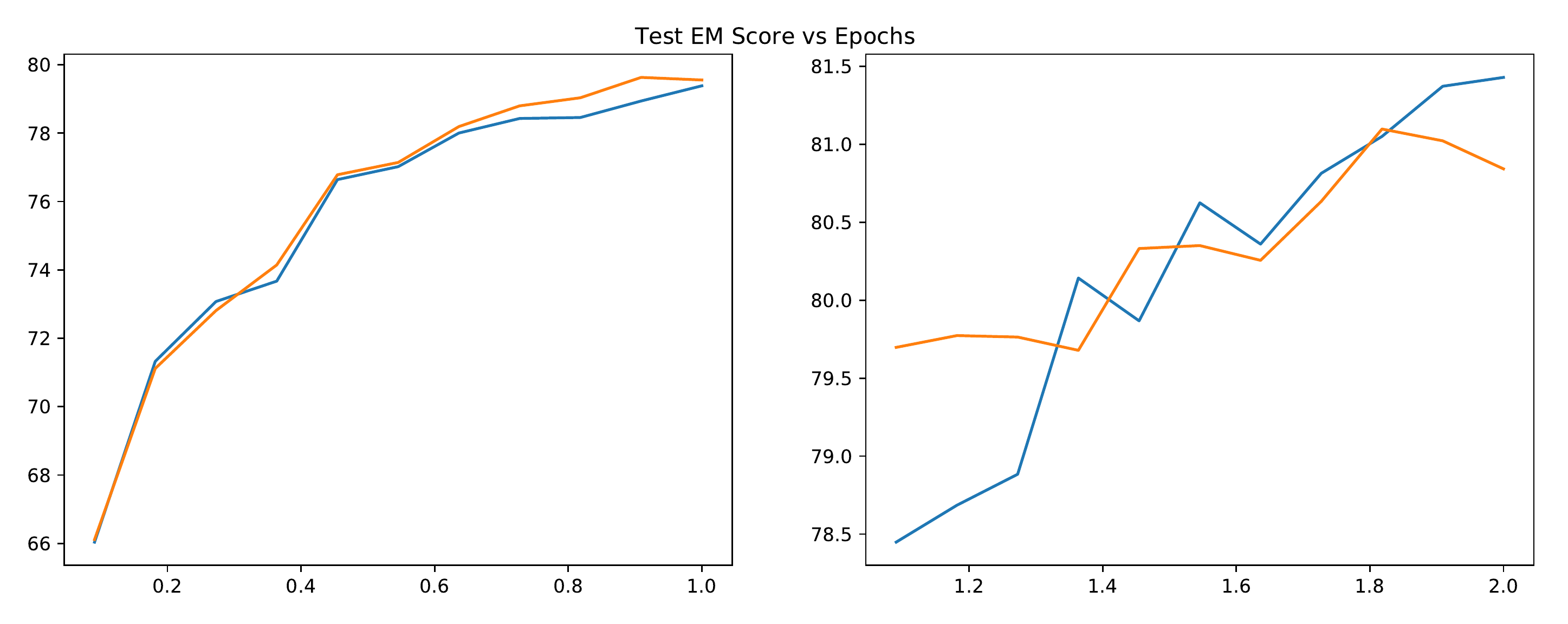}
        %\caption{Test Accuracy}
        \label{fig:squad_bert_adam_test_acc}
    \end{subfigure}
    \begin{subfigure}[t]{\textwidth}
        \centering
        \includegraphics[width=0.99\linewidth]{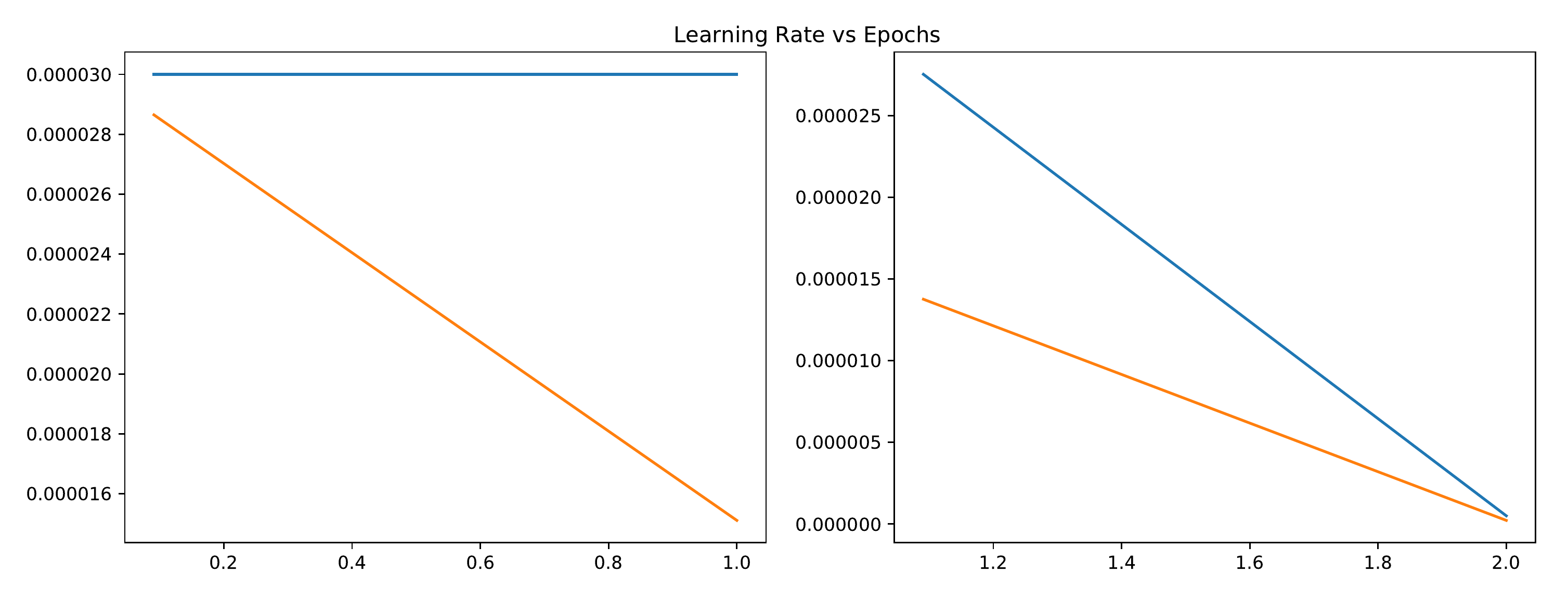}
        %\caption{Learning Rate}
        \label{fig:squad_bert_adam_lr}
    \end{subfigure}
\caption{SQuAD-v1.1 fine-tuning on BERT\textsubscript{BASE} trained with Adam. Shown are the training loss, test EM score, and learning rate as a function of epochs, for the baseline scheme (orange) vs the \lrschedule{} scheme (blue). The plot is split into 2 parts to permit higher fidelity in the y-axis range. It is clear that with \lrschedule{} the network starts to overfit after the 2nd epoch, where the testing loss continues to go down, but generalization suffers. We saw similar behavior with different seeds, and thus need to train with \lrschedule{} for only 2 epochs.}

\label{fig:squad_bert_adam_result}
\end{figure}

\end{document}